%% file: loopcd.tex
\documentclass[letterpaper]{article}
\usepackage{latexml}
\iflatexml
  \usepackage{amsmath,amssymb,booktabs,textcomp,graphicx,xcolor}
\else
\def\sheetscale{1}
\usepackage[paperwidth=612bp,paperheight=792bp,margin=0pt]{geometry}
\usepackage{pdfpages}
\usepackage{natbib}
\usepackage{atbegshi}
\fi
\usepackage{hyperref}
\iflatexml\else
\usepackage{bookmark}
\fi
\hypersetup{
  pdftitle={Decoding Looped Transformers Better for (Almost) Free},
  pdfauthor={Anonymous Authors},
  pdfsubject={LoopCD: contrastive decoding from the recurrent states a looped Transformer already
    computes, at one extra output pass or none},
  pdfkeywords={looped Transformers, recurrent depth, contrastive decoding, guidance, inference},
  pdfpagemode=UseOutlines,
  pdfstartview={FitH},
  bookmarksnumbered=false,
  hidelinks,
}

\newif\ifprintbib
\printbibfalse

\newif\ifpagelinks
\pagelinkstrue

\newif\ifdropshipout
\dropshipoutfalse
\newcounter{sheet}
\iflatexml\else
\makeatletter
\AtBeginShipout{%
  \ifdropshipout
    \AtBeginShipoutDiscard
  \else
    \stepcounter{sheet}%
    \ifpagelinks
      \@ifundefined{lcd@sheet@\the\value{sheet}}{}{%
        \AtBeginShipoutUpperLeft{\@nameuse{lcd@sheet@\the\value{sheet}}}%
      }%
    \fi
  \fi
}
\makeatother
\input{links}
\fi

\begin{document}

\iflatexml
\input{html-body}

\else

\includepdf[pages=-,noautoscale=true,scale=\sheetscale,offset=0 0]{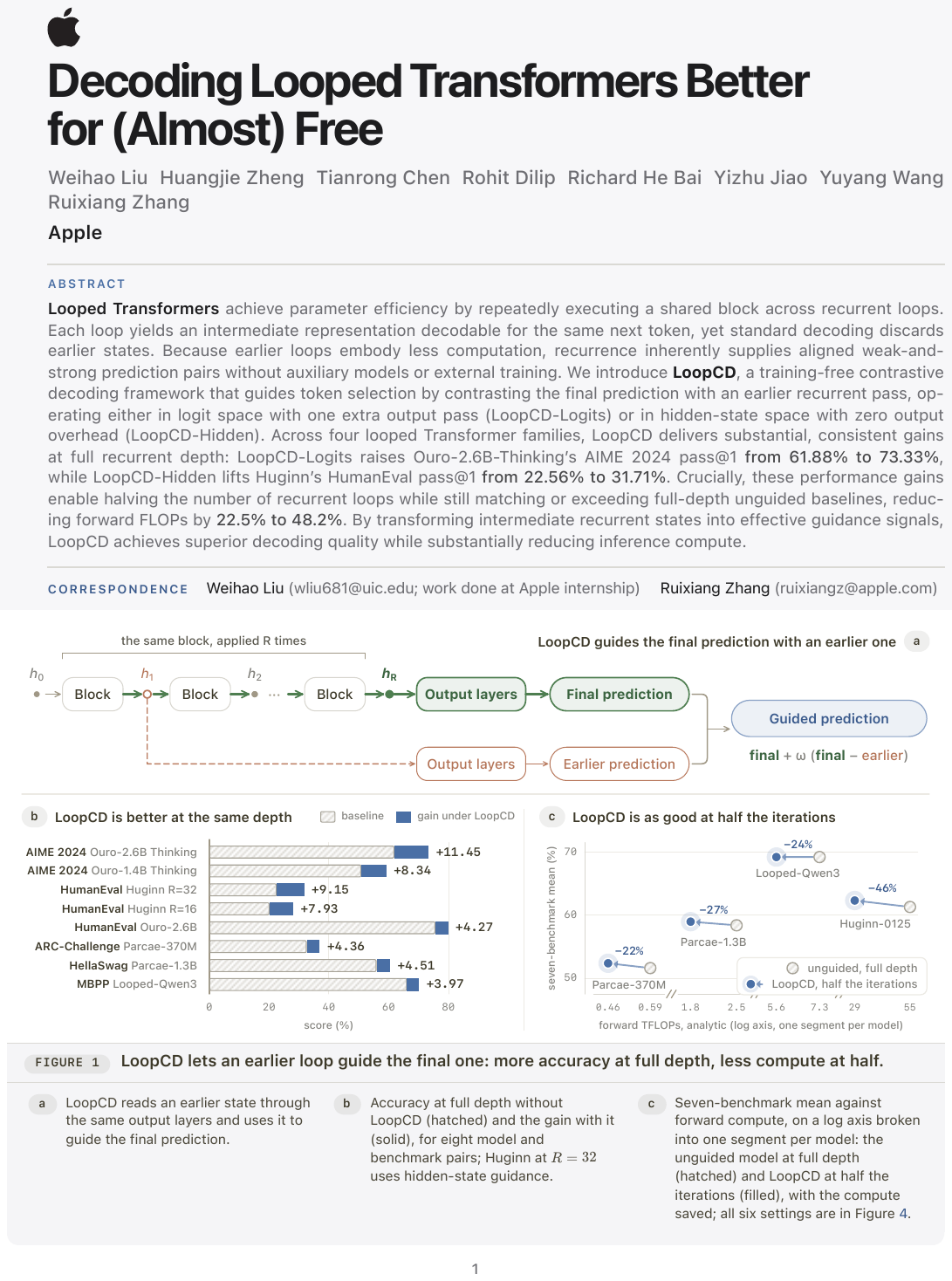}

\input{bookmarks}

\clearpage
\ifprintbib
  \newgeometry{margin=20mm}
\else
  \dropshipouttrue
\fi
\nocite{*}
\bibliographystyle{plainnat}
\bibliography{loopcd}
\fi

\end{document}

%% file: links.tex
\makeatletter
\@namedef{lcd@sheet@1}{%
  \setlength{\unitlength}{1.0000000000bp}%
  \put(36.75,-390.75){\pdfdest name{fig-hero} xyz}%
  \put(195,-380.25){\pdfannot width 66.0000bp height 10.5000bp depth 0bp {/Subtype/Link/Border[0 0 0]/F 4/A<</S/URI/URI(mailto:wliu681@uic.edu)>>}}%
  \put(459,-380.25){\pdfannot width 90.0000bp height 10.5000bp depth 0bp {/Subtype/Link/Border[0 0 0]/F 4/A<</S/URI/URI(mailto:ruixiangz@apple.com)>>}}%
  \put(528,-723){\pdfannot width 4.5000bp height 9.0000bp depth 0bp {/Subtype/Link/Border[0 0 0]/F 4/A<</S/GoTo/D(fig-halfloop)>>}}%
}
\@namedef{lcd@sheet@2}{%
  \setlength{\unitlength}{1.0000000000bp}%
  \put(36.75,-660){\pdfdest name{sec-preliminaries} xyz}%
  \put(372.75,-128.25){\pdfannot width 95.2500bp height 11.2500bp depth 0bp {/Subtype/Link/Border[0 0 0]/F 4/A<</S/GoTo/D(ref-dehghani2019universal)>>}}%
  \put(473.25,-128.25){\pdfannot width 76.5000bp height 11.2500bp depth 0bp {/Subtype/Link/Border[0 0 0]/F 4/A<</S/GoTo/D(ref-ouro2025)>>}}%
  \put(59.25,-140.25){\pdfannot width 90.0000bp height 11.2500bp depth 0bp {/Subtype/Link/Border[0 0 0]/F 4/A<</S/GoTo/D(ref-geiping2025huginn)>>}}%
  \put(155.25,-140.25){\pdfannot width 83.2500bp height 11.2500bp depth 0bp {/Subtype/Link/Border[0 0 0]/F 4/A<</S/GoTo/D(ref-parcae2026)>>}}%
  \put(478.5,-194.25){\pdfannot width 67.5000bp height 11.2500bp depth 0bp {/Subtype/Link/Border[0 0 0]/F 4/A<</S/GoTo/D(ref-li2022contrastive)>>}}%
  \put(282.75,-219){\pdfannot width 83.2500bp height 11.2500bp depth 0bp {/Subtype/Link/Border[0 0 0]/F 4/A<</S/GoTo/D(ref-karras2024autoguidance)>>}}%
  \put(104.25,-243.75){\pdfannot width 62.2500bp height 11.2500bp depth 0bp {/Subtype/Link/Border[0 0 0]/F 4/A<</S/GoTo/D(ref-li2022contrastive)>>}}%
  \put(172.5,-243.75){\pdfannot width 68.2500bp height 11.2500bp depth 0bp {/Subtype/Link/Border[0 0 0]/F 4/A<</S/GoTo/D(ref-shi2023trusting)>>}}%
  \put(246.75,-243.75){\pdfannot width 89.2500bp height 11.2500bp depth 0bp {/Subtype/Link/Border[0 0 0]/F 4/A<</S/GoTo/D(ref-chuang2023dola)>>}}%
  \put(111.75,-450){\pdfannot width 76.5000bp height 11.2500bp depth 0bp {/Subtype/Link/Border[0 0 0]/F 4/A<</S/GoTo/D(ref-ouro2025)>>}}%
  \put(236.25,-450){\pdfannot width 88.5000bp height 11.2500bp depth 0bp {/Subtype/Link/Border[0 0 0]/F 4/A<</S/GoTo/D(ref-geiping2025huginn)>>}}%
  \put(371.25,-450){\pdfannot width 83.2500bp height 11.2500bp depth 0bp {/Subtype/Link/Border[0 0 0]/F 4/A<</S/GoTo/D(ref-parcae2026)>>}}%
  \put(63,-462.75){\pdfannot width 78.7500bp height 11.2500bp depth 0bp {/Subtype/Link/Border[0 0 0]/F 4/A<</S/GoTo/D(ref-tflt2026)>>}}%
  \put(506.25,-474.75){\pdfannot width 46.5000bp height 11.2500bp depth 0bp {/Subtype/Link/Border[0 0 0]/F 4/A<</S/GoTo/D(ref-ouro2025)>>}}%
  \put(59.25,-487.5){\pdfannot width 28.5000bp height 11.2500bp depth 0bp {/Subtype/Link/Border[0 0 0]/F 4/A<</S/GoTo/D(ref-ouro2025)>>}}%
  \put(351,-487.5){\pdfannot width 3.7500bp height 11.2500bp depth 0bp {/Subtype/Link/Border[0 0 0]/F 4/A<</S/GoTo/D(fig-hero)>>}}%
  \put(84,-708.75){\pdfannot width 96.0000bp height 11.2500bp depth 0bp {/Subtype/Link/Border[0 0 0]/F 4/A<</S/GoTo/D(ref-dehghani2019universal)>>}}%
}
\@namedef{lcd@sheet@3}{%
  \setlength{\unitlength}{1.0000000000bp}%
  \put(59.25,-144){\pdfdest name{fig-looped-background} xyz}%
  \put(59.25,-440.25){\pdfdest name{eq-recurrence} xyz}%
  \put(36.75,-524.25){\pdfdest name{sec-method} xyz}%
  \put(59.25,-653.25){\pdfdest name{sec-guidance} xyz}%
  \put(77.25,-72){\pdfannot width 5.2500bp height 11.2500bp depth 0bp {/Subtype/Link/Border[0 0 0]/F 4/A<</S/GoTo/D(fig-looped-background)>>}}%
  \put(255.75,-72){\pdfannot width 77.2500bp height 11.2500bp depth 0bp {/Subtype/Link/Border[0 0 0]/F 4/A<</S/GoTo/D(ref-ouro2025)>>}}%
  \put(234,-96.75){\pdfannot width 89.2500bp height 11.2500bp depth 0bp {/Subtype/Link/Border[0 0 0]/F 4/A<</S/GoTo/D(ref-geiping2025huginn)>>}}%
  \put(465,-121.5){\pdfannot width 84.7500bp height 11.2500bp depth 0bp {/Subtype/Link/Border[0 0 0]/F 4/A<</S/GoTo/D(ref-parcae2026)>>}}%
  \put(59.25,-133.5){\pdfannot width 77.2500bp height 11.2500bp depth 0bp {/Subtype/Link/Border[0 0 0]/F 4/A<</S/GoTo/D(ref-tflt2026)>>}}%
  \put(190.5,-133.5){\pdfannot width 15.0000bp height 11.2500bp depth 0bp {/Subtype/Link/Border[0 0 0]/F 4/A<</S/GoTo/D(app-architecture-models)>>}}%
  \put(498.75,-378){\pdfannot width 9.0000bp height 9.7500bp depth 0bp {/Subtype/Link/Border[0 0 0]/F 4/A<</S/GoTo/D(tab-arch)>>}}%
  \put(277.5,-561){\pdfannot width 11.2500bp height 11.2500bp depth 0bp {/Subtype/Link/Border[0 0 0]/F 4/A<</S/GoTo/D(eq-recurrence)>>}}%
  \put(434.25,-597.75){\pdfannot width 63.7500bp height 11.2500bp depth 0bp {/Subtype/Link/Border[0 0 0]/F 4/A<</S/GoTo/D(ref-li2022contrastive)>>}}%
  \put(228.75,-622.5){\pdfannot width 93.7500bp height 11.2500bp depth 0bp {/Subtype/Link/Border[0 0 0]/F 4/A<</S/GoTo/D(ref-ho2022cfg)>>}}%
  \put(327.75,-622.5){\pdfannot width 82.5000bp height 11.2500bp depth 0bp {/Subtype/Link/Border[0 0 0]/F 4/A<</S/GoTo/D(ref-karras2024autoguidance)>>}}%
  \put(338.25,-639){\pdfannot width 12.7500bp height 11.2500bp depth 0bp {/Subtype/Link/Border[0 0 0]/F 4/A<</S/GoTo/D(sec-guidance)>>}}%
  \put(528,-639){\pdfannot width 14.2500bp height 11.2500bp depth 0bp {/Subtype/Link/Border[0 0 0]/F 4/A<</S/GoTo/D(sec-guidance-strength)>>}}%
  \put(491.25,-683.25){\pdfannot width 6.0000bp height 11.2500bp depth 0bp {/Subtype/Link/Border[0 0 0]/F 4/A<</S/GoTo/D(fig-looped-architecture)>>}}%
}
\@namedef{lcd@sheet@4}{%
  \setlength{\unitlength}{1.0000000000bp}%
  \put(59.25,-57){\pdfdest name{eq-hidden-guidance} xyz}%
  \put(59.25,-192){\pdfdest name{eq-loopcd} xyz}%
  \put(59.25,-307.5){\pdfdest name{fig-looped-architecture} xyz}%
  \put(59.25,-505.5){\pdfdest name{sec-guidance-strength} xyz}%
  \put(59.25,-588){\pdfdest name{eq-tokenwise} xyz}%
  \put(129,-153){\pdfannot width 14.2500bp height 11.2500bp depth 0bp {/Subtype/Link/Border[0 0 0]/F 4/A<</S/GoTo/D(sec-reduced-loop)>>}}%
  \put(106.5,-284.25){\pdfannot width 15.0000bp height 11.2500bp depth 0bp {/Subtype/Link/Border[0 0 0]/F 4/A<</S/GoTo/D(app-probability-guidance)>>}}%
  \put(236.25,-462.75){\pdfannot width 11.2500bp height 9.7500bp depth 0bp {/Subtype/Link/Border[0 0 0]/F 4/A<</S/GoTo/D(eq-hidden-guidance)>>}}%
  \put(264,-474){\pdfannot width 11.2500bp height 9.7500bp depth 0bp {/Subtype/Link/Border[0 0 0]/F 4/A<</S/GoTo/D(eq-loopcd)>>}}%
  \put(494.25,-552.75){\pdfannot width 15.0000bp height 11.2500bp depth 0bp {/Subtype/Link/Border[0 0 0]/F 4/A<</S/GoTo/D(app-selection)>>}}%
}
\@namedef{lcd@sheet@5}{%
  \setlength{\unitlength}{1.0000000000bp}%
  \put(59.25,-57.75){\pdfdest name{tab-reason} xyz}%
  \put(59.25,-645.75){\pdfdest name{sec-main} xyz}%
  \put(522,-516.75){\pdfannot width 12.7500bp height 11.2500bp depth 0bp {/Subtype/Link/Border[0 0 0]/F 4/A<</S/GoTo/D(sec-main)>>}}%
  \put(430.5,-528.75){\pdfannot width 14.2500bp height 11.2500bp depth 0bp {/Subtype/Link/Border[0 0 0]/F 4/A<</S/GoTo/D(sec-generation)>>}}%
  \put(321.75,-566.25){\pdfannot width 14.2500bp height 11.2500bp depth 0bp {/Subtype/Link/Border[0 0 0]/F 4/A<</S/GoTo/D(sec-reduced-loop)>>}}%
  \put(282,-582.75){\pdfannot width 83.2500bp height 11.2500bp depth 0bp {/Subtype/Link/Border[0 0 0]/F 4/A<</S/GoTo/D(ref-ouro2025)>>}}%
  \put(416.25,-582.75){\pdfannot width 94.5000bp height 11.2500bp depth 0bp {/Subtype/Link/Border[0 0 0]/F 4/A<</S/GoTo/D(ref-geiping2025huginn)>>}}%
  \put(63,-595.5){\pdfannot width 88.5000bp height 11.2500bp depth 0bp {/Subtype/Link/Border[0 0 0]/F 4/A<</S/GoTo/D(ref-parcae2026)>>}}%
  \put(261,-595.5){\pdfannot width 83.2500bp height 11.2500bp depth 0bp {/Subtype/Link/Border[0 0 0]/F 4/A<</S/GoTo/D(ref-tflt2026)>>}}%
  \put(461.25,-595.5){\pdfannot width 84.7500bp height 11.2500bp depth 0bp {/Subtype/Link/Border[0 0 0]/F 4/A<</S/GoTo/D(ref-qwen3)>>}}%
  \put(138.75,-620.25){\pdfannot width 13.5000bp height 11.2500bp depth 0bp {/Subtype/Link/Border[0 0 0]/F 4/A<</S/GoTo/D(app-model-settings)>>}}%
  \put(474.75,-620.25){\pdfannot width 15.0000bp height 11.2500bp depth 0bp {/Subtype/Link/Border[0 0 0]/F 4/A<</S/GoTo/D(app-benchmarks)>>}}%
  \put(247.5,-689.25){\pdfannot width 103.5000bp height 11.2500bp depth 0bp {/Subtype/Link/Border[0 0 0]/F 4/A<</S/GoTo/D(ref-h42025aime2024)>>}}%
  \put(425.25,-689.25){\pdfannot width 98.2500bp height 11.2500bp depth 0bp {/Subtype/Link/Border[0 0 0]/F 4/A<</S/GoTo/D(ref-opencompass2025aime2025)>>}}%
  \put(137.25,-701.25){\pdfannot width 69.7500bp height 11.2500bp depth 0bp {/Subtype/Link/Border[0 0 0]/F 4/A<</S/GoTo/D(ref-he2024olympiadbench)>>}}%
  \put(510.75,-701.25){\pdfannot width 4.5000bp height 11.2500bp depth 0bp {/Subtype/Link/Border[0 0 0]/F 4/A<</S/GoTo/D(tab-reason)>>}}%
}
\@namedef{lcd@sheet@6}{%
  \setlength{\unitlength}{1.0000000000bp}%
  \put(59.25,-57.75){\pdfdest name{tab-gen} xyz}%
  \put(59.25,-582.75){\pdfdest name{sec-generation} xyz}%
  \put(414.75,-396.75){\pdfannot width 12.0000bp height 9.7500bp depth 0bp {/Subtype/Link/Border[0 0 0]/F 4/A<</S/GoTo/D(app-benchmarks)>>}}%
  \put(213,-507.75){\pdfannot width 78.7500bp height 11.2500bp depth 0bp {/Subtype/Link/Border[0 0 0]/F 4/A<</S/GoTo/D(ref-chen2021humaneval)>>}}%
  \put(352.5,-507.75){\pdfannot width 83.2500bp height 11.2500bp depth 0bp {/Subtype/Link/Border[0 0 0]/F 4/A<</S/GoTo/D(ref-austin2021mbpp)>>}}%
  \put(153.75,-519.75){\pdfannot width 69.7500bp height 11.2500bp depth 0bp {/Subtype/Link/Border[0 0 0]/F 4/A<</S/GoTo/D(ref-liu2023evalplus)>>}}%
  \put(261.75,-519.75){\pdfannot width 3.7500bp height 11.2500bp depth 0bp {/Subtype/Link/Border[0 0 0]/F 4/A<</S/GoTo/D(tab-reason)>>}}%
  \put(174.75,-569.25){\pdfannot width 5.2500bp height 11.2500bp depth 0bp {/Subtype/Link/Border[0 0 0]/F 4/A<</S/GoTo/D(tab-hidden)>>}}%
  \put(403.5,-626.25){\pdfannot width 75.7500bp height 11.2500bp depth 0bp {/Subtype/Link/Border[0 0 0]/F 4/A<</S/GoTo/D(ref-clark2018arc)>>}}%
  \put(516,-626.25){\pdfannot width 36.7500bp height 11.2500bp depth 0bp {/Subtype/Link/Border[0 0 0]/F 4/A<</S/GoTo/D(ref-welbl2017sciq)>>}}%
  \put(59.25,-638.25){\pdfannot width 39.0000bp height 11.2500bp depth 0bp {/Subtype/Link/Border[0 0 0]/F 4/A<</S/GoTo/D(ref-welbl2017sciq)>>}}%
  \put(145.5,-638.25){\pdfannot width 104.2500bp height 11.2500bp depth 0bp {/Subtype/Link/Border[0 0 0]/F 4/A<</S/GoTo/D(ref-hendrycks2021mmlu)>>}}%
  \put(315,-638.25){\pdfannot width 87.0000bp height 11.2500bp depth 0bp {/Subtype/Link/Border[0 0 0]/F 4/A<</S/GoTo/D(ref-zellers2019hellaswag)>>}}%
  \put(475.5,-638.25){\pdfannot width 77.2500bp height 11.2500bp depth 0bp {/Subtype/Link/Border[0 0 0]/F 4/A<</S/GoTo/D(ref-sakaguchi2020winogrande)>>}}%
  \put(59.25,-651){\pdfannot width 24.0000bp height 11.2500bp depth 0bp {/Subtype/Link/Border[0 0 0]/F 4/A<</S/GoTo/D(ref-sakaguchi2020winogrande)>>}}%
  \put(144,-651){\pdfannot width 76.5000bp height 11.2500bp depth 0bp {/Subtype/Link/Border[0 0 0]/F 4/A<</S/GoTo/D(ref-bisk2020piqa)>>}}%
  \put(456,-663){\pdfannot width 6.0000bp height 11.2500bp depth 0bp {/Subtype/Link/Border[0 0 0]/F 4/A<</S/GoTo/D(tab-gen)>>}}%
}
\@namedef{lcd@sheet@7}{%
  \setlength{\unitlength}{1.0000000000bp}%
  \put(59.25,-57.75){\pdfdest name{tab-hidden} xyz}%
  \put(59.25,-437.25){\pdfdest name{sec-reduced-loop} xyz}%
  \put(59.25,-731.25){\pdfdest name{sec-weak} xyz}%
  \put(363,-307.5){\pdfannot width 78.7500bp height 11.2500bp depth 0bp {/Subtype/Link/Border[0 0 0]/F 4/A<</S/GoTo/D(ref-wang2024mmlupro)>>}}%
  \put(478.5,-307.5){\pdfannot width 12.0000bp height 11.2500bp depth 0bp {/Subtype/Link/Border[0 0 0]/F 4/A<</S/GoTo/D(tab-mmlupro)>>}}%
  \put(355.5,-324.75){\pdfannot width 6.0000bp height 11.2500bp depth 0bp {/Subtype/Link/Border[0 0 0]/F 4/A<</S/GoTo/D(tab-hidden)>>}}%
  \put(140.25,-348.75){\pdfannot width 12.0000bp height 11.2500bp depth 0bp {/Subtype/Link/Border[0 0 0]/F 4/A<</S/GoTo/D(sec-guidance)>>}}%
  \put(312,-348.75){\pdfannot width 6.0000bp height 11.2500bp depth 0bp {/Subtype/Link/Border[0 0 0]/F 4/A<</S/GoTo/D(tab-hidden)>>}}%
  \put(87.75,-373.5){\pdfannot width 6.0000bp height 11.2500bp depth 0bp {/Subtype/Link/Border[0 0 0]/F 4/A<</S/GoTo/D(tab-gen)>>}}%
  \put(401.25,-373.5){\pdfannot width 6.0000bp height 11.2500bp depth 0bp {/Subtype/Link/Border[0 0 0]/F 4/A<</S/GoTo/D(tab-hidden)>>}}%
  \put(451.5,-386.25){\pdfannot width 4.5000bp height 11.2500bp depth 0bp {/Subtype/Link/Border[0 0 0]/F 4/A<</S/GoTo/D(tab-reason)>>}}%
  \put(373.5,-398.25){\pdfannot width 15.0000bp height 11.2500bp depth 0bp {/Subtype/Link/Border[0 0 0]/F 4/A<</S/GoTo/D(app-form-comparison)>>}}%
  \put(223.5,-529.5){\pdfannot width 6.0000bp height 11.2500bp depth 0bp {/Subtype/Link/Border[0 0 0]/F 4/A<</S/GoTo/D(fig-halfloop)>>}}%
  \put(129,-595.5){\pdfannot width 6.0000bp height 11.2500bp depth 0bp {/Subtype/Link/Border[0 0 0]/F 4/A<</S/GoTo/D(fig-halfloop)>>}}%
}
\@namedef{lcd@sheet@8}{%
  \setlength{\unitlength}{1.0000000000bp}%
  \put(59.25,-57.75){\pdfdest name{fig-halfloop} xyz}%
  \put(59.25,-345){\pdfdest name{fig-weak} xyz}%
  \put(378.75,-702.75){\pdfannot width 6.0000bp height 11.2500bp depth 0bp {/Subtype/Link/Border[0 0 0]/F 4/A<</S/GoTo/D(fig-weak)>>}}%
  \put(172.5,-714.75){\pdfannot width 6.0000bp height 11.2500bp depth 0bp {/Subtype/Link/Border[0 0 0]/F 4/A<</S/GoTo/D(fig-weak)>>}}%
  \put(255,-714.75){\pdfannot width 12.7500bp height 11.2500bp depth 0bp {/Subtype/Link/Border[0 0 0]/F 4/A<</S/GoTo/D(app-reference-sweep)>>}}%
}
\@namedef{lcd@sheet@9}{%
  \setlength{\unitlength}{1.0000000000bp}%
  \put(59.25,-57.75){\pdfdest name{fig-contrast-gain} xyz}%
  \put(59.25,-343.5){\pdfdest name{fig-reference} xyz}%
  \put(296.25,-623.25){\pdfannot width 6.0000bp height 11.2500bp depth 0bp {/Subtype/Link/Border[0 0 0]/F 4/A<</S/GoTo/D(fig-contrast-gain)>>}}%
  \put(527.25,-623.25){\pdfannot width 6.0000bp height 11.2500bp depth 0bp {/Subtype/Link/Border[0 0 0]/F 4/A<</S/GoTo/D(fig-contrast-gain)>>}}%
  \put(465,-726){\pdfannot width 6.0000bp height 11.2500bp depth 0bp {/Subtype/Link/Border[0 0 0]/F 4/A<</S/GoTo/D(fig-reference)>>}}%
}
\@namedef{lcd@sheet@10}{%
  \setlength{\unitlength}{1.0000000000bp}%
  \put(59.25,-57.75){\pdfdest name{fig-decomp} xyz}%
  \put(59.25,-393.75){\pdfdest name{sec-pair-analysis} xyz}%
  \put(59.25,-640.5){\pdfdest name{sec-strength-analysis} xyz}%
  \put(201,-368.25){\pdfannot width 5.2500bp height 11.2500bp depth 0bp {/Subtype/Link/Border[0 0 0]/F 4/A<</S/GoTo/D(fig-reference)>>}}%
  \put(356.25,-380.25){\pdfannot width 15.0000bp height 11.2500bp depth 0bp {/Subtype/Link/Border[0 0 0]/F 4/A<</S/GoTo/D(app-selection)>>}}%
  \put(232.5,-449.25){\pdfannot width 6.0000bp height 11.2500bp depth 0bp {/Subtype/Link/Border[0 0 0]/F 4/A<</S/GoTo/D(fig-decomp)>>}}%
  \put(174.75,-510.75){\pdfannot width 6.0000bp height 11.2500bp depth 0bp {/Subtype/Link/Border[0 0 0]/F 4/A<</S/GoTo/D(fig-decomp)>>}}%
  \put(137.25,-523.5){\pdfannot width 12.7500bp height 11.2500bp depth 0bp {/Subtype/Link/Border[0 0 0]/F 4/A<</S/GoTo/D(app-rerank-identity)>>}}%
  \put(427.5,-589.5){\pdfannot width 6.7500bp height 11.2500bp depth 0bp {/Subtype/Link/Border[0 0 0]/F 4/A<</S/GoTo/D(fig-decomp)>>}}%
  \put(497.25,-589.5){\pdfannot width 11.2500bp height 11.2500bp depth 0bp {/Subtype/Link/Border[0 0 0]/F 4/A<</S/GoTo/D(app-close-decisions)>>}}%
  \put(452.25,-614.25){\pdfannot width 14.2500bp height 11.2500bp depth 0bp {/Subtype/Link/Border[0 0 0]/F 4/A<</S/GoTo/D(sec-guidance-strength)>>}}%
  \put(365.25,-696){\pdfannot width 6.0000bp height 11.2500bp depth 0bp {/Subtype/Link/Border[0 0 0]/F 4/A<</S/GoTo/D(fig-strength-sweep)>>}}%
}
\@namedef{lcd@sheet@11}{%
  \setlength{\unitlength}{1.0000000000bp}%
  \put(59.25,-57.75){\pdfdest name{fig-strength-sweep} xyz}%
  \put(59.25,-321.75){\pdfdest name{fig-strength-adaptive} xyz}%
  \put(320.25,-620.25){\pdfannot width 14.2500bp height 11.2500bp depth 0bp {/Subtype/Link/Border[0 0 0]/F 4/A<</S/GoTo/D(app-strength-sweeps)>>}}%
  \put(408,-648.75){\pdfannot width 13.5000bp height 11.2500bp depth 0bp {/Subtype/Link/Border[0 0 0]/F 4/A<</S/GoTo/D(eq-tokenwise)>>}}%
  \put(352.5,-686.25){\pdfannot width 9.7500bp height 11.2500bp depth 0bp {/Subtype/Link/Border[0 0 0]/F 4/A<</S/GoTo/D(fig-strength-adaptive)>>}}%
  \put(413.25,-686.25){\pdfannot width 15.0000bp height 11.2500bp depth 0bp {/Subtype/Link/Border[0 0 0]/F 4/A<</S/GoTo/D(app-quintile-reach)>>}}%
}
\@namedef{lcd@sheet@12}{%
  \setlength{\unitlength}{1.0000000000bp}%
  \put(59.25,-258.75){\pdfdest name{ref-austin2021mbpp} xyz}%
  \put(313.5,-258.75){\pdfdest name{ref-dehghani2019universal} xyz}%
  \put(313.5,-306.75){\pdfdest name{ref-ICLR2026_a4b1a25e} xyz}%
  \put(59.25,-318){\pdfdest name{ref-bisk2020piqa} xyz}%
  \put(59.25,-366){\pdfdest name{ref-tflt2026} xyz}%
  \put(313.5,-398.25){\pdfdest name{ref-fan2025length} xyz}%
  \put(59.25,-402.75){\pdfdest name{ref-chen2021humaneval} xyz}%
  \put(59.25,-462){\pdfdest name{ref-chen2022analog} xyz}%
  \put(313.5,-479.25){\pdfdest name{ref-fan2026lotus} xyz}%
  \put(59.25,-498.75){\pdfdest name{ref-chen2025reasoning} xyz}%
  \put(313.5,-527.25){\pdfdest name{ref-gao2023lmeval} xyz}%
  \put(59.25,-558){\pdfdest name{ref-chuang2023dola} xyz}%
  \put(59.25,-616.5){\pdfdest name{ref-clark2018arc} xyz}%
  \put(313.5,-618.75){\pdfdest name{ref-geiping2025huginn} xyz}%
  \put(59.25,-664.5){\pdfdest name{ref-cobbe2021gsm8k} xyz}%
  \put(488.25,-371.25){\pdfannot width 64.5000bp height 9.7500bp depth 0bp {/Subtype/Link/Border[0 0 0]/F 4/A<</S/URI/URI(https://proceedings.iclr.cc/paper_files/paper/2026/file/a4b1a25e1b37a17733c1fea50d0b7446-Paper-Conference.pdf)>>}}%
  \put(324,-382.5){\pdfannot width 226.5000bp height 9.7500bp depth 0bp {/Subtype/Link/Border[0 0 0]/F 4/A<</S/URI/URI(https://proceedings.iclr.cc/paper_files/paper/2026/file/a4b1a25e1b37a17733c1fea50d0b7446-Paper-Conference.pdf)>>}}%
  \put(324,-393){\pdfannot width 147.0000bp height 9.7500bp depth 0bp {/Subtype/Link/Border[0 0 0]/F 4/A<</S/URI/URI(https://proceedings.iclr.cc/paper_files/paper/2026/file/a4b1a25e1b37a17733c1fea50d0b7446-Paper-Conference.pdf)>>}}%
  \put(423,-452.25){\pdfannot width 129.7500bp height 9.7500bp depth 0bp {/Subtype/Link/Border[0 0 0]/F 4/A<</S/URI/URI(https://proceedings.iclr.cc/paper_files/paper/2025/file/25cc3adf8c85f7c70989cb8a97a691a7-Paper-Conference.pdf)>>}}%
  \put(324,-462.75){\pdfannot width 225.7500bp height 9.7500bp depth 0bp {/Subtype/Link/Border[0 0 0]/F 4/A<</S/URI/URI(https://proceedings.iclr.cc/paper_files/paper/2025/file/25cc3adf8c85f7c70989cb8a97a691a7-Paper-Conference.pdf)>>}}%
  \put(324,-474){\pdfannot width 83.2500bp height 9.7500bp depth 0bp {/Subtype/Link/Border[0 0 0]/F 4/A<</S/URI/URI(https://proceedings.iclr.cc/paper_files/paper/2025/file/25cc3adf8c85f7c70989cb8a97a691a7-Paper-Conference.pdf)>>}}%
  \put(512.25,-510.75){\pdfannot width 40.5000bp height 9.7500bp depth 0bp {/Subtype/Link/Border[0 0 0]/F 4/A<</S/URI/URI(https://arxiv.org/abs/2606.31779)>>}}%
  \put(324,-522){\pdfannot width 81.7500bp height 9.7500bp depth 0bp {/Subtype/Link/Border[0 0 0]/F 4/A<</S/URI/URI(https://arxiv.org/abs/2606.31779)>>}}%
}
\@namedef{lcd@sheet@13}{%
  \setlength{\unitlength}{1.0000000000bp}%
  \put(59.25,-48){\pdfdest name{ref-giannou2023programmable} xyz}%
  \put(313.5,-48){\pdfdest name{ref-liu2026latent} xyz}%
  \put(313.5,-106.5){\pdfdest name{ref-obrien2023reasoning} xyz}%
  \put(313.5,-154.5){\pdfdest name{ref-opencompass2025aime2025} xyz}%
  \put(59.25,-161.25){\pdfdest name{ref-hao2025training} xyz}%
  \put(313.5,-202.5){\pdfdest name{ref-parcae2026} xyz}%
  \put(59.25,-219.75){\pdfdest name{ref-he2024olympiadbench} xyz}%
  \put(313.5,-240){\pdfdest name{ref-qwen3} xyz}%
  \put(313.5,-255){\pdfdest name{ref-sakaguchi2020winogrande} xyz}%
  \put(313.5,-303){\pdfdest name{ref-sanchez2023stay} xyz}%
  \put(59.25,-311.25){\pdfdest name{ref-hendrycks2021mmlu} xyz}%
  \put(313.5,-351){\pdfdest name{ref-saunshi2025latentthoughts} xyz}%
  \put(59.25,-359.25){\pdfdest name{ref-ho2022cfg} xyz}%
  \put(59.25,-385.5){\pdfdest name{ref-h42025aime2024} xyz}%
  \put(59.25,-433.5){\pdfdest name{ref-ICLR2026_843e1d8a} xyz}%
  \put(313.5,-442.5){\pdfdest name{ref-shi2023trusting} xyz}%
  \put(313.5,-490.5){\pdfdest name{ref-tang2026subtraction} xyz}%
  \put(59.25,-525){\pdfdest name{ref-karras2024autoguidance} xyz}%
  \put(313.5,-549.75){\pdfdest name{ref-wang2024mmlupro} xyz}%
  \put(59.25,-573){\pdfdest name{ref-li2022contrastive} xyz}%
  \put(313.5,-619.5){\pdfdest name{ref-welbl2017sciq} xyz}%
  \put(59.25,-642.75){\pdfdest name{ref-liu2023evalplus} xyz}%
  \put(313.5,-667.5){\pdfdest name{ref-williams2026rltt} xyz}%
  \put(112.5,-144.75){\pdfannot width 185.2500bp height 9.7500bp depth 0bp {/Subtype/Link/Border[0 0 0]/F 4/A<</S/URI/URI(https://proceedings.mlr.press/v202/giannou23a.html)>>}}%
  \put(69.75,-156){\pdfannot width 12.0000bp height 9.7500bp depth 0bp {/Subtype/Link/Border[0 0 0]/F 4/A<</S/URI/URI(https://proceedings.mlr.press/v202/giannou23a.html)>>}}%
  \put(282.75,-204){\pdfannot width 15.0000bp height 9.7500bp depth 0bp {/Subtype/Link/Border[0 0 0]/F 4/A<</S/URI/URI(https://openreview.net/forum?id=Itxz7S4Ip3)>>}}%
  \put(69.75,-214.5){\pdfannot width 149.2500bp height 9.7500bp depth 0bp {/Subtype/Link/Border[0 0 0]/F 4/A<</S/URI/URI(https://openreview.net/forum?id=Itxz7S4Ip3)>>}}%
  \put(69.75,-406.5){\pdfannot width 222.0000bp height 9.7500bp depth 0bp {/Subtype/Link/Border[0 0 0]/F 4/A<</S/URI/URI(https://huggingface.co/datasets/HuggingFaceH4/aime_2024)>>}}%
  \put(69.75,-417.75){\pdfannot width 5.2500bp height 9.7500bp depth 0bp {/Subtype/Link/Border[0 0 0]/F 4/A<</S/URI/URI(https://huggingface.co/datasets/HuggingFaceH4/aime_2024)>>}}%
  \put(228.75,-498){\pdfannot width 69.0000bp height 9.7500bp depth 0bp {/Subtype/Link/Border[0 0 0]/F 4/A<</S/URI/URI(https://proceedings.iclr.cc/paper_files/paper/2026/file/843e1d8ab4819c8c8a4221fce2c8d455-Paper-Conference.pdf)>>}}%
  \put(69.75,-509.25){\pdfannot width 225.7500bp height 9.7500bp depth 0bp {/Subtype/Link/Border[0 0 0]/F 4/A<</S/URI/URI(https://proceedings.iclr.cc/paper_files/paper/2026/file/843e1d8ab4819c8c8a4221fce2c8d455-Paper-Conference.pdf)>>}}%
  \put(69.75,-519.75){\pdfannot width 146.2500bp height 9.7500bp depth 0bp {/Subtype/Link/Border[0 0 0]/F 4/A<</S/URI/URI(https://proceedings.iclr.cc/paper_files/paper/2026/file/843e1d8ab4819c8c8a4221fce2c8d455-Paper-Conference.pdf)>>}}%
  \put(438,-90.75){\pdfannot width 114.7500bp height 9.7500bp depth 0bp {/Subtype/Link/Border[0 0 0]/F 4/A<</S/URI/URI(https://openreview.net/forum?id=SFTFsixO2K)>>}}%
  \put(324,-101.25){\pdfannot width 55.5000bp height 9.7500bp depth 0bp {/Subtype/Link/Border[0 0 0]/F 4/A<</S/URI/URI(https://openreview.net/forum?id=SFTFsixO2K)>>}}%
  \put(439.5,-138.75){\pdfannot width 113.2500bp height 9.7500bp depth 0bp {/Subtype/Link/Border[0 0 0]/F 4/A<</S/URI/URI(https://arxiv.org/abs/2309.09117)>>}}%
  \put(324,-149.25){\pdfannot width 7.5000bp height 9.7500bp depth 0bp {/Subtype/Link/Border[0 0 0]/F 4/A<</S/URI/URI(https://arxiv.org/abs/2309.09117)>>}}%
  \put(324,-175.5){\pdfannot width 216.0000bp height 9.7500bp depth 0bp {/Subtype/Link/Border[0 0 0]/F 4/A<</S/URI/URI(https://huggingface.co/datasets/opencompass/AIME2025)>>}}%
  \put(369,-415.5){\pdfannot width 183.7500bp height 9.7500bp depth 0bp {/Subtype/Link/Border[0 0 0]/F 4/A<</S/URI/URI(https://proceedings.iclr.cc/paper_files/paper/2025/file/2676109d49d1eb26d6bc584a8f556305-Paper-Conference.pdf)>>}}%
  \put(324,-426.75){\pdfannot width 225.7500bp height 9.7500bp depth 0bp {/Subtype/Link/Border[0 0 0]/F 4/A<</S/URI/URI(https://proceedings.iclr.cc/paper_files/paper/2025/file/2676109d49d1eb26d6bc584a8f556305-Paper-Conference.pdf)>>}}%
  \put(324,-437.25){\pdfannot width 36.0000bp height 9.7500bp depth 0bp {/Subtype/Link/Border[0 0 0]/F 4/A<</S/URI/URI(https://proceedings.iclr.cc/paper_files/paper/2025/file/2676109d49d1eb26d6bc584a8f556305-Paper-Conference.pdf)>>}}%
  \put(498,-533.25){\pdfannot width 54.7500bp height 9.7500bp depth 0bp {/Subtype/Link/Border[0 0 0]/F 4/A<</S/URI/URI(https://arxiv.org/abs/2602.18232)>>}}%
  \put(324,-544.5){\pdfannot width 68.2500bp height 9.7500bp depth 0bp {/Subtype/Link/Border[0 0 0]/F 4/A<</S/URI/URI(https://arxiv.org/abs/2602.18232)>>}}%
  \put(495,-720.75){\pdfannot width 57.7500bp height 9.7500bp depth 0bp {/Subtype/Link/Border[0 0 0]/F 4/A<</S/URI/URI(https://openreview.net/forum?id=cPAgJwcn3h)>>}}%
  \put(324,-732){\pdfannot width 115.5000bp height 9.7500bp depth 0bp {/Subtype/Link/Border[0 0 0]/F 4/A<</S/URI/URI(https://openreview.net/forum?id=cPAgJwcn3h)>>}}%
}
\@namedef{lcd@sheet@14}{%
  \setlength{\unitlength}{1.0000000000bp}%
  \put(59.25,-48){\pdfdest name{ref-yang2026loopcoderv2} xyz}%
  \put(313.5,-48){\pdfdest name{ref-zhou2025alw} xyz}%
  \put(59.25,-128.25){\pdfdest name{ref-yang2024learningalgorithms} xyz}%
  \put(313.5,-161.25){\pdfdest name{ref-ouro2025} xyz}%
  \put(59.25,-219.75){\pdfdest name{ref-zellers2019hellaswag} xyz}%
  \put(313.5,-263.25){\pdfdest name{ref-zhu2025virtualdepth} xyz}%
  \put(111.75,-123){\pdfannot width 122.2500bp height 9.7500bp depth 0bp {/Subtype/Link/Border[0 0 0]/F 4/A<</S/URI/URI(https://arxiv.org/abs/2606.18023)>>}}%
  \put(168,-192.75){\pdfannot width 129.7500bp height 9.7500bp depth 0bp {/Subtype/Link/Border[0 0 0]/F 4/A<</S/URI/URI(https://proceedings.iclr.cc/paper_files/paper/2024/file/b8402301e7f06bdc97a31bfaa653dc32-Paper-Conference.pdf)>>}}%
  \put(69.75,-204){\pdfannot width 226.5000bp height 9.7500bp depth 0bp {/Subtype/Link/Border[0 0 0]/F 4/A<</S/URI/URI(https://proceedings.iclr.cc/paper_files/paper/2024/file/b8402301e7f06bdc97a31bfaa653dc32-Paper-Conference.pdf)>>}}%
  \put(69.75,-214.5){\pdfannot width 83.2500bp height 9.7500bp depth 0bp {/Subtype/Link/Border[0 0 0]/F 4/A<</S/URI/URI(https://proceedings.iclr.cc/paper_files/paper/2024/file/b8402301e7f06bdc97a31bfaa653dc32-Paper-Conference.pdf)>>}}%
  \put(481.5,-144.75){\pdfannot width 71.2500bp height 9.7500bp depth 0bp {/Subtype/Link/Border[0 0 0]/F 4/A<</S/URI/URI(https://aclanthology.org/2025.findings-acl.447/)>>}}%
  \put(324,-156){\pdfannot width 105.7500bp height 9.7500bp depth 0bp {/Subtype/Link/Border[0 0 0]/F 4/A<</S/URI/URI(https://aclanthology.org/2025.findings-acl.447/)>>}}%
  \put(324,-317.25){\pdfannot width 122.2500bp height 9.7500bp depth 0bp {/Subtype/Link/Border[0 0 0]/F 4/A<</S/URI/URI(https://arxiv.org/abs/2506.18233)>>}}%
}
\@namedef{lcd@sheet@15}{%
  \setlength{\unitlength}{1.0000000000bp}%
  \put(59.25,-250.5){\pdfdest name{tab-arch} xyz}%
  \put(59.25,-483.75){\pdfdest name{app-architecture-models} xyz}%
  \put(194.25,-100.5){\pdfannot width 5.2500bp height 10.5000bp depth 0bp {/Subtype/Link/Border[0 0 0]/F 4/A<</S/GoTo/D(sec-preliminaries)>>}}%
  \put(219,-100.5){\pdfannot width 6.0000bp height 10.5000bp depth 0bp {/Subtype/Link/Border[0 0 0]/F 4/A<</S/GoTo/D(sec-method)>>}}%
  \put(437.25,-100.5){\pdfannot width 12.0000bp height 10.5000bp depth 0bp {/Subtype/Link/Border[0 0 0]/F 4/A<</S/GoTo/D(eq-loopcd)>>}}%
  \put(522.75,-217.5){\pdfannot width 9.7500bp height 10.5000bp depth 0bp {/Subtype/Link/Border[0 0 0]/F 4/A<</S/GoTo/D(tab-arch)>>}}%
  \put(311.25,-462.75){\pdfannot width 9.7500bp height 9.7500bp depth 0bp {/Subtype/Link/Border[0 0 0]/F 4/A<</S/GoTo/D(tab-flops)>>}}%
  \put(168.75,-609){\pdfannot width 69.7500bp height 10.5000bp depth 0bp {/Subtype/Link/Border[0 0 0]/F 4/A<</S/GoTo/D(ref-ouro2025)>>}}%
  \put(350.25,-632.25){\pdfannot width 82.5000bp height 10.5000bp depth 0bp {/Subtype/Link/Border[0 0 0]/F 4/A<</S/GoTo/D(ref-geiping2025huginn)>>}}%
}
\@namedef{lcd@sheet@16}{%
  \setlength{\unitlength}{1.0000000000bp}%
  \put(59.25,-303){\pdfdest name{app-probability-guidance} xyz}%
  \put(59.25,-467.25){\pdfdest name{app-model-settings} xyz}%
  \put(59.25,-595.5){\pdfdest name{app-benchmarks} xyz}%
  \put(254.25,-173.25){\pdfannot width 36.7500bp height 10.5000bp depth 0bp {/Subtype/Link/Border[0 0 0]/F 4/A<</S/GoTo/D(ref-parcae2026)>>}}%
  \put(70.5,-185.25){\pdfannot width 35.2500bp height 10.5000bp depth 0bp {/Subtype/Link/Border[0 0 0]/F 4/A<</S/GoTo/D(ref-parcae2026)>>}}%
  \put(492.75,-185.25){\pdfannot width 48.7500bp height 10.5000bp depth 0bp {/Subtype/Link/Border[0 0 0]/F 4/A<</S/GoTo/D(ref-tflt2026)>>}}%
  \put(321,-196.5){\pdfannot width 21.0000bp height 10.5000bp depth 0bp {/Subtype/Link/Border[0 0 0]/F 4/A<</S/GoTo/D(ref-tflt2026)>>}}%
  \put(462.75,-231.75){\pdfannot width 5.2500bp height 10.5000bp depth 0bp {/Subtype/Link/Border[0 0 0]/F 4/A<</S/GoTo/D(fig-halfloop)>>}}%
  \put(520.5,-344.25){\pdfannot width 12.0000bp height 10.5000bp depth 0bp {/Subtype/Link/Border[0 0 0]/F 4/A<</S/GoTo/D(eq-loopcd)>>}}%
  \put(93,-366.75){\pdfannot width 11.2500bp height 10.5000bp depth 0bp {/Subtype/Link/Border[0 0 0]/F 4/A<</S/GoTo/D(sec-guidance)>>}}%
  \put(380.25,-519.75){\pdfannot width 13.5000bp height 10.5000bp depth 0bp {/Subtype/Link/Border[0 0 0]/F 4/A<</S/GoTo/D(app-architecture-models)>>}}%
  \put(438,-519.75){\pdfannot width 9.7500bp height 10.5000bp depth 0bp {/Subtype/Link/Border[0 0 0]/F 4/A<</S/GoTo/D(tab-arch)>>}}%
  \put(365.25,-665.25){\pdfannot width 72.7500bp height 10.5000bp depth 0bp {/Subtype/Link/Border[0 0 0]/F 4/A<</S/GoTo/D(ref-clark2018arc)>>}}%
  \put(497.25,-665.25){\pdfannot width 55.5000bp height 10.5000bp depth 0bp {/Subtype/Link/Border[0 0 0]/F 4/A<</S/GoTo/D(ref-zellers2019hellaswag)>>}}%
  \put(59.25,-676.5){\pdfannot width 20.2500bp height 10.5000bp depth 0bp {/Subtype/Link/Border[0 0 0]/F 4/A<</S/GoTo/D(ref-zellers2019hellaswag)>>}}%
  \put(144.75,-676.5){\pdfannot width 91.5000bp height 10.5000bp depth 0bp {/Subtype/Link/Border[0 0 0]/F 4/A<</S/GoTo/D(ref-sakaguchi2020winogrande)>>}}%
  \put(270,-676.5){\pdfannot width 66.7500bp height 10.5000bp depth 0bp {/Subtype/Link/Border[0 0 0]/F 4/A<</S/GoTo/D(ref-bisk2020piqa)>>}}%
  \put(369.75,-676.5){\pdfannot width 70.5000bp height 10.5000bp depth 0bp {/Subtype/Link/Border[0 0 0]/F 4/A<</S/GoTo/D(ref-welbl2017sciq)>>}}%
  \put(498,-676.5){\pdfannot width 54.7500bp height 10.5000bp depth 0bp {/Subtype/Link/Border[0 0 0]/F 4/A<</S/GoTo/D(ref-hendrycks2021mmlu)>>}}%
  \put(59.25,-687.75){\pdfannot width 34.5000bp height 10.5000bp depth 0bp {/Subtype/Link/Border[0 0 0]/F 4/A<</S/GoTo/D(ref-hendrycks2021mmlu)>>}}%
  \put(440.25,-687.75){\pdfannot width 65.2500bp height 10.5000bp depth 0bp {/Subtype/Link/Border[0 0 0]/F 4/A<</S/GoTo/D(ref-gao2023lmeval)>>}}%
  \put(258.75,-705){\pdfannot width 73.5000bp height 10.5000bp depth 0bp {/Subtype/Link/Border[0 0 0]/F 4/A<</S/GoTo/D(ref-chen2021humaneval)>>}}%
  \put(372.75,-705){\pdfannot width 77.2500bp height 10.5000bp depth 0bp {/Subtype/Link/Border[0 0 0]/F 4/A<</S/GoTo/D(ref-austin2021mbpp)>>}}%
  \put(498,-705){\pdfannot width 54.7500bp height 10.5000bp depth 0bp {/Subtype/Link/Border[0 0 0]/F 4/A<</S/GoTo/D(ref-cobbe2021gsm8k)>>}}%
  \put(59.25,-716.25){\pdfannot width 20.2500bp height 10.5000bp depth 0bp {/Subtype/Link/Border[0 0 0]/F 4/A<</S/GoTo/D(ref-cobbe2021gsm8k)>>}}%
  \put(157.5,-716.25){\pdfannot width 75.0000bp height 10.5000bp depth 0bp {/Subtype/Link/Border[0 0 0]/F 4/A<</S/GoTo/D(ref-wang2024mmlupro)>>}}%
  \put(513.75,-716.25){\pdfannot width 39.0000bp height 10.5000bp depth 0bp {/Subtype/Link/Border[0 0 0]/F 4/A<</S/GoTo/D(ref-liu2023evalplus)>>}}%
  \put(59.25,-727.5){\pdfannot width 21.7500bp height 10.5000bp depth 0bp {/Subtype/Link/Border[0 0 0]/F 4/A<</S/GoTo/D(ref-liu2023evalplus)>>}}%
}
\@namedef{lcd@sheet@17}{%
  \setlength{\unitlength}{1.0000000000bp}%
  \put(59.25,-191.25){\pdfdest name{tab-fewshot} xyz}%
  \put(59.25,-423){\pdfdest name{tab-fewshot-gen} xyz}%
  \put(59.25,-611.25){\pdfdest name{app-selection} xyz}%
  \put(285.75,-58.5){\pdfannot width 93.0000bp height 10.5000bp depth 0bp {/Subtype/Link/Border[0 0 0]/F 4/A<</S/GoTo/D(ref-h42025aime2024)>>}}%
  \put(442.5,-58.5){\pdfannot width 88.5000bp height 10.5000bp depth 0bp {/Subtype/Link/Border[0 0 0]/F 4/A<</S/GoTo/D(ref-opencompass2025aime2025)>>}}%
  \put(512.25,-69.75){\pdfannot width 40.5000bp height 10.5000bp depth 0bp {/Subtype/Link/Border[0 0 0]/F 4/A<</S/GoTo/D(ref-he2024olympiadbench)>>}}%
  \put(59.25,-81){\pdfannot width 21.7500bp height 10.5000bp depth 0bp {/Subtype/Link/Border[0 0 0]/F 4/A<</S/GoTo/D(ref-he2024olympiadbench)>>}}%
  \put(503.25,-92.25){\pdfannot width 49.5000bp height 10.5000bp depth 0bp {/Subtype/Link/Border[0 0 0]/F 4/A<</S/GoTo/D(ref-chen2021humaneval)>>}}%
  \put(59.25,-103.5){\pdfannot width 20.2500bp height 10.5000bp depth 0bp {/Subtype/Link/Border[0 0 0]/F 4/A<</S/GoTo/D(ref-chen2021humaneval)>>}}%
  \put(84.75,-103.5){\pdfannot width 69.0000bp height 10.5000bp depth 0bp {/Subtype/Link/Border[0 0 0]/F 4/A<</S/GoTo/D(ref-ouro2025)>>}}%
  \put(88.5,-147){\pdfannot width 11.2500bp height 10.5000bp depth 0bp {/Subtype/Link/Border[0 0 0]/F 4/A<</S/GoTo/D(tab-fewshot)>>}}%
  \put(120.75,-147){\pdfannot width 11.2500bp height 10.5000bp depth 0bp {/Subtype/Link/Border[0 0 0]/F 4/A<</S/GoTo/D(tab-fewshot-gen)>>}}%
  \put(408.75,-391.5){\pdfannot width 5.2500bp height 9.7500bp depth 0bp {/Subtype/Link/Border[0 0 0]/F 4/A<</S/GoTo/D(tab-gen)>>}}%
  \put(432.75,-391.5){\pdfannot width 4.5000bp height 9.7500bp depth 0bp {/Subtype/Link/Border[0 0 0]/F 4/A<</S/GoTo/D(tab-hidden)>>}}%
  \put(490.5,-641.25){\pdfannot width 12.7500bp height 10.5000bp depth 0bp {/Subtype/Link/Border[0 0 0]/F 4/A<</S/GoTo/D(app-strength-sweeps)>>}}%
  \put(544.5,-641.25){\pdfannot width 5.2500bp height 10.5000bp depth 0bp {/Subtype/Link/Border[0 0 0]/F 4/A<</S/GoTo/D(fig-strength-sweep)>>}}%
  \put(77.25,-652.5){\pdfannot width 9.7500bp height 10.5000bp depth 0bp {/Subtype/Link/Border[0 0 0]/F 4/A<</S/GoTo/D(fig-strength-adaptive)>>}}%
  \put(119.25,-652.5){\pdfannot width 11.2500bp height 10.5000bp depth 0bp {/Subtype/Link/Border[0 0 0]/F 4/A<</S/GoTo/D(tab-strengths)>>}}%
  \put(528,-732){\pdfannot width 11.2500bp height 10.5000bp depth 0bp {/Subtype/Link/Border[0 0 0]/F 4/A<</S/GoTo/D(tab-strengths)>>}}%
}
\@namedef{lcd@sheet@18}{%
  \setlength{\unitlength}{1.0000000000bp}%
  \put(59.25,-102.75){\pdfdest name{tab-strengths} xyz}%
  \put(499.5,-58.5){\pdfannot width 11.2500bp height 10.5000bp depth 0bp {/Subtype/Link/Border[0 0 0]/F 4/A<</S/GoTo/D(app-reference-sweep)>>}}%
  \put(529.5,-92.25){\pdfannot width 11.2500bp height 10.5000bp depth 0bp {/Subtype/Link/Border[0 0 0]/F 4/A<</S/GoTo/D(tab-reduced-settings)>>}}%
  \put(453.75,-372.75){\pdfannot width 10.5000bp height 9.7500bp depth 0bp {/Subtype/Link/Border[0 0 0]/F 4/A<</S/GoTo/D(tab-reduced-settings)>>}}%
  \put(221.25,-433.5){\pdfannot width 11.2500bp height 10.5000bp depth 0bp {/Subtype/Link/Border[0 0 0]/F 4/A<</S/GoTo/D(sec-main)>>}}%
  \put(257.25,-433.5){\pdfannot width 12.7500bp height 10.5000bp depth 0bp {/Subtype/Link/Border[0 0 0]/F 4/A<</S/GoTo/D(sec-generation)>>}}%
  \put(110.25,-456){\pdfannot width 13.5000bp height 10.5000bp depth 0bp {/Subtype/Link/Border[0 0 0]/F 4/A<</S/GoTo/D(app-selection)>>}}%
  \put(519.75,-522){\pdfannot width 5.2500bp height 10.5000bp depth 0bp {/Subtype/Link/Border[0 0 0]/F 4/A<</S/GoTo/D(tab-gen)>>}}%
  \put(84.75,-550.5){\pdfannot width 11.2500bp height 10.5000bp depth 0bp {/Subtype/Link/Border[0 0 0]/F 4/A<</S/GoTo/D(tab-mmlupro)>>}}%
  \put(220.5,-627.75){\pdfannot width 3.7500bp height 10.5000bp depth 0bp {/Subtype/Link/Border[0 0 0]/F 4/A<</S/GoTo/D(tab-reason)>>}}%
  \put(95.25,-639){\pdfannot width 5.2500bp height 10.5000bp depth 0bp {/Subtype/Link/Border[0 0 0]/F 4/A<</S/GoTo/D(tab-gen)>>}}%
  \put(518.25,-678.75){\pdfannot width 11.2500bp height 10.5000bp depth 0bp {/Subtype/Link/Border[0 0 0]/F 4/A<</S/GoTo/D(tab-qwen-reasoning)>>}}%
}
\@namedef{lcd@sheet@19}{%
  \setlength{\unitlength}{1.0000000000bp}%
  \put(59.25,-57.75){\pdfdest name{tab-mmlupro} xyz}%
  \put(59.25,-321){\pdfdest name{tab-qwen-reasoning} xyz}%
  \put(462,-256.5){\pdfannot width 12.0000bp height 9.7500bp depth 0bp {/Subtype/Link/Border[0 0 0]/F 4/A<</S/GoTo/D(app-benchmarks)>>}}%
  \put(282.75,-462.75){\pdfannot width 3.7500bp height 9.7500bp depth 0bp {/Subtype/Link/Border[0 0 0]/F 4/A<</S/GoTo/D(tab-reason)>>}}%
  \put(213,-523.5){\pdfannot width 12.7500bp height 10.5000bp depth 0bp {/Subtype/Link/Border[0 0 0]/F 4/A<</S/GoTo/D(sec-reduced-loop)>>}}%
  \put(250.5,-645.75){\pdfannot width 10.5000bp height 10.5000bp depth 0bp {/Subtype/Link/Border[0 0 0]/F 4/A<</S/GoTo/D(tab-flops)>>}}%
}
\@namedef{lcd@sheet@20}{%
  \setlength{\unitlength}{1.0000000000bp}%
  \put(59.25,-57.75){\pdfdest name{tab-flops} xyz}%
  \put(59.25,-354){\pdfdest name{tab-reduced-settings} xyz}%
  \put(59.25,-555.75){\pdfdest name{app-reduced-details} xyz}%
  \put(345.75,-373.5){\pdfannot width 6.0000bp height 11.2500bp depth 0bp {/Subtype/Link/Border[0 0 0]/F 4/A<</S/GoTo/D(fig-halfloop)>>}}%
  \put(188.25,-597){\pdfannot width 11.2500bp height 10.5000bp depth 0bp {/Subtype/Link/Border[0 0 0]/F 4/A<</S/GoTo/D(tab-reduced-settings)>>}}%
  \put(544.5,-597){\pdfannot width 5.2500bp height 10.5000bp depth 0bp {/Subtype/Link/Border[0 0 0]/F 4/A<</S/GoTo/D(fig-halfloop)>>}}%
  \put(153,-670.5){\pdfannot width 12.7500bp height 10.5000bp depth 0bp {/Subtype/Link/Border[0 0 0]/F 4/A<</S/GoTo/D(sec-reduced-loop)>>}}%
}
\@namedef{lcd@sheet@21}{%
  \setlength{\unitlength}{1.0000000000bp}%
  \put(59.25,-57.75){\pdfdest name{fig-anatomy} xyz}%
  \put(59.25,-540){\pdfdest name{app-form-comparison} xyz}%
  \put(357.75,-430.5){\pdfannot width 9.7500bp height 10.5000bp depth 0bp {/Subtype/Link/Border[0 0 0]/F 4/A<</S/GoTo/D(fig-anatomy)>>}}%
  \put(413.25,-464.25){\pdfannot width 9.7500bp height 10.5000bp depth 0bp {/Subtype/Link/Border[0 0 0]/F 4/A<</S/GoTo/D(fig-anatomy)>>}}%
  \put(84,-515.25){\pdfannot width 11.2500bp height 10.5000bp depth 0bp {/Subtype/Link/Border[0 0 0]/F 4/A<</S/GoTo/D(tab-fractions)>>}}%
  \put(507,-515.25){\pdfannot width 5.2500bp height 10.5000bp depth 0bp {/Subtype/Link/Border[0 0 0]/F 4/A<</S/GoTo/D(fig-halfloop)>>}}%
  \put(350.25,-621){\pdfannot width 5.2500bp height 10.5000bp depth 0bp {/Subtype/Link/Border[0 0 0]/F 4/A<</S/GoTo/D(tab-hidden)>>}}%
  \put(395.25,-621){\pdfannot width 5.2500bp height 10.5000bp depth 0bp {/Subtype/Link/Border[0 0 0]/F 4/A<</S/GoTo/D(tab-gen)>>}}%
  \put(364.5,-632.25){\pdfannot width 11.2500bp height 10.5000bp depth 0bp {/Subtype/Link/Border[0 0 0]/F 4/A<</S/GoTo/D(fig-space-tradeoff)>>}}%
}
\@namedef{lcd@sheet@22}{%
  \setlength{\unitlength}{1.0000000000bp}%
  \put(59.25,-57.75){\pdfdest name{tab-fractions} xyz}%
  \put(59.25,-270){\pdfdest name{fig-space-tradeoff} xyz}%
  \put(261,-546.75){\pdfannot width 5.2500bp height 9.7500bp depth 0bp {/Subtype/Link/Border[0 0 0]/F 4/A<</S/GoTo/D(tab-gen)>>}}%
  \put(286.5,-546.75){\pdfannot width 5.2500bp height 9.7500bp depth 0bp {/Subtype/Link/Border[0 0 0]/F 4/A<</S/GoTo/D(tab-hidden)>>}}%
  \put(381.75,-651.75){\pdfannot width 13.5000bp height 10.5000bp depth 0bp {/Subtype/Link/Border[0 0 0]/F 4/A<</S/GoTo/D(app-selection)>>}}%
  \put(181.5,-685.5){\pdfannot width 11.2500bp height 10.5000bp depth 0bp {/Subtype/Link/Border[0 0 0]/F 4/A<</S/GoTo/D(fig-scoring-benchmarks)>>}}%
}
\@namedef{lcd@sheet@23}{%
  \setlength{\unitlength}{1.0000000000bp}%
  \put(59.25,-57.75){\pdfdest name{fig-scoring-benchmarks} xyz}%
  \put(59.25,-339){\pdfdest name{tab-step-sizes} xyz}%
  \put(402,-307.5){\pdfannot width 4.5000bp height 9.7500bp depth 0bp {/Subtype/Link/Border[0 0 0]/F 4/A<</S/GoTo/D(tab-gen)>>}}%
  \put(427.5,-307.5){\pdfannot width 4.5000bp height 9.7500bp depth 0bp {/Subtype/Link/Border[0 0 0]/F 4/A<</S/GoTo/D(tab-hidden)>>}}%
  \put(408.75,-563.25){\pdfannot width 11.2500bp height 10.5000bp depth 0bp {/Subtype/Link/Border[0 0 0]/F 4/A<</S/GoTo/D(fig-space-tradeoff)>>}}%
  \put(253.5,-637.5){\pdfannot width 14.2500bp height 10.5000bp depth 0bp {/Subtype/Link/Border[0 0 0]/F 4/A<</S/GoTo/D(tab-step-sizes)>>}}%
  \put(300.75,-637.5){\pdfannot width 11.2500bp height 10.5000bp depth 0bp {/Subtype/Link/Border[0 0 0]/F 4/A<</S/GoTo/D(fig-space-step)>>}}%
  \put(166.5,-699.75){\pdfannot width 13.5000bp height 10.5000bp depth 0bp {/Subtype/Link/Border[0 0 0]/F 4/A<</S/GoTo/D(app-selection)>>}}%
}
\@namedef{lcd@sheet@24}{%
  \setlength{\unitlength}{1.0000000000bp}%
  \put(59.25,-91.5){\pdfdest name{fig-space-step} xyz}%
  \put(59.25,-528){\pdfdest name{app-reference-sweep} xyz}%
  \put(304.5,-81){\pdfannot width 11.2500bp height 10.5000bp depth 0bp {/Subtype/Link/Border[0 0 0]/F 4/A<</S/GoTo/D(fig-verdicts)>>}}%
  \put(437.25,-81){\pdfannot width 11.2500bp height 10.5000bp depth 0bp {/Subtype/Link/Border[0 0 0]/F 4/A<</S/GoTo/D(app-reference-sweep)>>}}%
  \put(228,-480.75){\pdfannot width 11.2500bp height 10.5000bp depth 0bp {/Subtype/Link/Border[0 0 0]/F 4/A<</S/GoTo/D(sec-weak)>>}}%
  \put(311.25,-558){\pdfannot width 13.5000bp height 10.5000bp depth 0bp {/Subtype/Link/Border[0 0 0]/F 4/A<</S/GoTo/D(tab-reference-sweep)>>}}%
  \put(198.75,-683.25){\pdfannot width 5.2500bp height 10.5000bp depth 0bp {/Subtype/Link/Border[0 0 0]/F 4/A<</S/GoTo/D(fig-reference)>>}}%
}
\@namedef{lcd@sheet@25}{%
  \setlength{\unitlength}{1.0000000000bp}%
  \put(59.25,-102.75){\pdfdest name{tab-reference-sweep} xyz}%
  \put(229.5,-525){\pdfannot width 10.5000bp height 9.7500bp depth 0bp {/Subtype/Link/Border[0 0 0]/F 4/A<</S/GoTo/D(tab-fewshot)>>}}%
  \put(363,-575.25){\pdfannot width 10.5000bp height 10.5000bp depth 0bp {/Subtype/Link/Border[0 0 0]/F 4/A<</S/GoTo/D(fig-geometry)>>}}%
  \put(396.75,-575.25){\pdfannot width 11.2500bp height 10.5000bp depth 0bp {/Subtype/Link/Border[0 0 0]/F 4/A<</S/GoTo/D(fig-cosmatrix)>>}}%
  \put(401.25,-699.75){\pdfannot width 11.2500bp height 10.5000bp depth 0bp {/Subtype/Link/Border[0 0 0]/F 4/A<</S/GoTo/D(fig-geometry)>>}}%
}
\@namedef{lcd@sheet@26}{%
  \setlength{\unitlength}{1.0000000000bp}%
  \put(59.25,-57.75){\pdfdest name{fig-geometry} xyz}%
  \put(59.25,-321){\pdfdest name{fig-cosmatrix} xyz}%
  \put(306,-300.75){\pdfannot width 3.7500bp height 9.7500bp depth 0bp {/Subtype/Link/Border[0 0 0]/F 4/A<</S/GoTo/D(tab-reason)>>}}%
  \put(328.5,-300.75){\pdfannot width 4.5000bp height 9.7500bp depth 0bp {/Subtype/Link/Border[0 0 0]/F 4/A<</S/GoTo/D(tab-gen)>>}}%
}
\@namedef{lcd@sheet@27}{%
  \setlength{\unitlength}{1.0000000000bp}%
  \put(59.25,-258.75){\pdfdest name{fig-contrast} xyz}%
  \put(88.5,-119.25){\pdfannot width 10.5000bp height 10.5000bp depth 0bp {/Subtype/Link/Border[0 0 0]/F 4/A<</S/GoTo/D(fig-contrast)>>}}%
  \put(375.75,-644.25){\pdfannot width 82.5000bp height 10.5000bp depth 0bp {/Subtype/Link/Border[0 0 0]/F 4/A<</S/GoTo/D(ref-chuang2023dola)>>}}%
  \put(501,-684){\pdfannot width 11.2500bp height 10.5000bp depth 0bp {/Subtype/Link/Border[0 0 0]/F 4/A<</S/GoTo/D(fig-contrast)>>}}%
  \put(90,-729){\pdfannot width 13.5000bp height 10.5000bp depth 0bp {/Subtype/Link/Border[0 0 0]/F 4/A<</S/GoTo/D(app-reduced-details)>>}}%
}
\@namedef{lcd@sheet@28}{%
  \setlength{\unitlength}{1.0000000000bp}%
  \put(59.25,-170.25){\pdfdest name{app-close-decisions} xyz}%
  \put(59.25,-318.75){\pdfdest name{tab-close-decisions} xyz}%
  \put(59.25,-639){\pdfdest name{app-rerank-identity} xyz}%
  \put(216.75,-123){\pdfannot width 12.7500bp height 10.5000bp depth 0bp {/Subtype/Link/Border[0 0 0]/F 4/A<</S/GoTo/D(sec-pair-analysis)>>}}%
  \put(84,-240){\pdfannot width 15.0000bp height 10.5000bp depth 0bp {/Subtype/Link/Border[0 0 0]/F 4/A<</S/GoTo/D(tab-close-decisions)>>}}%
  \put(146.25,-526.5){\pdfannot width 9.7500bp height 9.7500bp depth 0bp {/Subtype/Link/Border[0 0 0]/F 4/A<</S/GoTo/D(tab-fewshot)>>}}%
  \put(332.25,-558){\pdfannot width 15.0000bp height 10.5000bp depth 0bp {/Subtype/Link/Border[0 0 0]/F 4/A<</S/GoTo/D(tab-worked-example)>>}}%
  \put(159.75,-669){\pdfannot width 12.7500bp height 10.5000bp depth 0bp {/Subtype/Link/Border[0 0 0]/F 4/A<</S/GoTo/D(sec-pair-analysis)>>}}%
}
\@namedef{lcd@sheet@29}{%
  \setlength{\unitlength}{1.0000000000bp}%
  \put(59.25,-131.25){\pdfdest name{tab-worked-example} xyz}%
  \put(59.25,-387.75){\pdfdest name{fig-verdicts} xyz}%
  \put(279.75,-326.25){\pdfannot width 11.2500bp height 10.5000bp depth 0bp {/Subtype/Link/Border[0 0 0]/F 4/A<</S/GoTo/D(fig-verdicts)>>}}%
  \put(484.5,-354.75){\pdfannot width 11.2500bp height 10.5000bp depth 0bp {/Subtype/Link/Border[0 0 0]/F 4/A<</S/GoTo/D(fig-verdicts)>>}}%
}
\@namedef{lcd@sheet@30}{%
  \setlength{\unitlength}{1.0000000000bp}%
  \put(59.25,-350.25){\pdfdest name{app-strength-sweeps} xyz}%
  \put(363,-69.75){\pdfannot width 12.0000bp height 10.5000bp depth 0bp {/Subtype/Link/Border[0 0 0]/F 4/A<</S/GoTo/D(app-reference-sweep)>>}}%
  \put(224.25,-314.25){\pdfannot width 12.7500bp height 10.5000bp depth 0bp {/Subtype/Link/Border[0 0 0]/F 4/A<</S/GoTo/D(sec-strength-analysis)>>}}%
  \put(88.5,-379.5){\pdfannot width 11.2500bp height 10.5000bp depth 0bp {/Subtype/Link/Border[0 0 0]/F 4/A<</S/GoTo/D(fig-gen-sweeps)>>}}%
  \put(314.25,-379.5){\pdfannot width 6.0000bp height 10.5000bp depth 0bp {/Subtype/Link/Border[0 0 0]/F 4/A<</S/GoTo/D(fig-strength-sweep)>>}}%
  \put(252.75,-487.5){\pdfannot width 6.0000bp height 10.5000bp depth 0bp {/Subtype/Link/Border[0 0 0]/F 4/A<</S/GoTo/D(fig-weak)>>}}%
}
\@namedef{lcd@sheet@31}{%
  \setlength{\unitlength}{1.0000000000bp}%
  \put(59.25,-57.75){\pdfdest name{fig-gen-sweeps} xyz}%
  \put(59.25,-305.25){\pdfdest name{app-quintile-reach} xyz}%
  \put(59.25,-402){\pdfdest name{tab-quintile-reach-app} xyz}%
  \put(171.75,-335.25){\pdfannot width 12.0000bp height 10.5000bp depth 0bp {/Subtype/Link/Border[0 0 0]/F 4/A<</S/GoTo/D(sec-strength-analysis)>>}}%
  \put(84.75,-346.5){\pdfannot width 15.7500bp height 10.5000bp depth 0bp {/Subtype/Link/Border[0 0 0]/F 4/A<</S/GoTo/D(tab-quintile-reach-app)>>}}%
  \put(155.25,-381){\pdfannot width 15.0000bp height 10.5000bp depth 0bp {/Subtype/Link/Border[0 0 0]/F 4/A<</S/GoTo/D(tab-quintile-reach-app)>>}}%
}
\@namedef{lcd@sheet@32}{%
  \setlength{\unitlength}{1.0000000000bp}%
  \put(207,-92.25){\pdfannot width 12.7500bp height 10.5000bp depth 0bp {/Subtype/Link/Border[0 0 0]/F 4/A<</S/GoTo/D(eq-tokenwise)>>}}%
  \put(357.75,-168.75){\pdfannot width 88.5000bp height 10.5000bp depth 0bp {/Subtype/Link/Border[0 0 0]/F 4/A<</S/GoTo/D(ref-dehghani2019universal)>>}}%
  \put(525,-180){\pdfannot width 27.7500bp height 10.5000bp depth 0bp {/Subtype/Link/Border[0 0 0]/F 4/A<</S/GoTo/D(ref-hao2025training)>>}}%
  \put(59.25,-191.25){\pdfannot width 36.7500bp height 10.5000bp depth 0bp {/Subtype/Link/Border[0 0 0]/F 4/A<</S/GoTo/D(ref-hao2025training)>>}}%
  \put(102,-191.25){\pdfannot width 72.0000bp height 10.5000bp depth 0bp {/Subtype/Link/Border[0 0 0]/F 4/A<</S/GoTo/D(ref-chen2025reasoning)>>}}%
  \put(179.25,-191.25){\pdfannot width 71.2500bp height 10.5000bp depth 0bp {/Subtype/Link/Border[0 0 0]/F 4/A<</S/GoTo/D(ref-ICLR2026_843e1d8a)>>}}%
  \put(256.5,-191.25){\pdfannot width 62.2500bp height 10.5000bp depth 0bp {/Subtype/Link/Border[0 0 0]/F 4/A<</S/GoTo/D(ref-ICLR2026_a4b1a25e)>>}}%
  \put(63,-213.75){\pdfannot width 84.0000bp height 10.5000bp depth 0bp {/Subtype/Link/Border[0 0 0]/F 4/A<</S/GoTo/D(ref-giannou2023programmable)>>}}%
  \put(152.25,-213.75){\pdfannot width 70.5000bp height 10.5000bp depth 0bp {/Subtype/Link/Border[0 0 0]/F 4/A<</S/GoTo/D(ref-yang2024learningalgorithms)>>}}%
  \put(228,-213.75){\pdfannot width 65.2500bp height 10.5000bp depth 0bp {/Subtype/Link/Border[0 0 0]/F 4/A<</S/GoTo/D(ref-fan2025length)>>}}%
  \put(298.5,-213.75){\pdfannot width 82.5000bp height 10.5000bp depth 0bp {/Subtype/Link/Border[0 0 0]/F 4/A<</S/GoTo/D(ref-saunshi2025latentthoughts)>>}}%
  \put(386.25,-213.75){\pdfannot width 72.0000bp height 10.5000bp depth 0bp {/Subtype/Link/Border[0 0 0]/F 4/A<</S/GoTo/D(ref-zhu2025virtualdepth)>>}}%
  \put(463.5,-213.75){\pdfannot width 62.2500bp height 10.5000bp depth 0bp {/Subtype/Link/Border[0 0 0]/F 4/A<</S/GoTo/D(ref-liu2026latent)>>}}%
  \put(189.75,-225){\pdfannot width 69.7500bp height 10.5000bp depth 0bp {/Subtype/Link/Border[0 0 0]/F 4/A<</S/GoTo/D(ref-ouro2025)>>}}%
  \put(303,-225){\pdfannot width 80.2500bp height 10.5000bp depth 0bp {/Subtype/Link/Border[0 0 0]/F 4/A<</S/GoTo/D(ref-geiping2025huginn)>>}}%
  \put(443.25,-225){\pdfannot width 75.7500bp height 10.5000bp depth 0bp {/Subtype/Link/Border[0 0 0]/F 4/A<</S/GoTo/D(ref-parcae2026)>>}}%
  \put(63,-258.75){\pdfannot width 63.7500bp height 10.5000bp depth 0bp {/Subtype/Link/Border[0 0 0]/F 4/A<</S/GoTo/D(ref-fan2026lotus)>>}}%
  \put(386.25,-258.75){\pdfannot width 83.2500bp height 10.5000bp depth 0bp {/Subtype/Link/Border[0 0 0]/F 4/A<</S/GoTo/D(ref-williams2026rltt)>>}}%
  \put(256.5,-270){\pdfannot width 69.7500bp height 10.5000bp depth 0bp {/Subtype/Link/Border[0 0 0]/F 4/A<</S/GoTo/D(ref-yang2026loopcoderv2)>>}}%
  \put(519,-321){\pdfannot width 33.7500bp height 10.5000bp depth 0bp {/Subtype/Link/Border[0 0 0]/F 4/A<</S/GoTo/D(ref-li2022contrastive)>>}}%
  \put(59.25,-332.25){\pdfannot width 21.7500bp height 10.5000bp depth 0bp {/Subtype/Link/Border[0 0 0]/F 4/A<</S/GoTo/D(ref-li2022contrastive)>>}}%
  \put(87,-332.25){\pdfannot width 95.2500bp height 10.5000bp depth 0bp {/Subtype/Link/Border[0 0 0]/F 4/A<</S/GoTo/D(ref-obrien2023reasoning)>>}}%
  \put(433.5,-343.5){\pdfannot width 65.2500bp height 10.5000bp depth 0bp {/Subtype/Link/Border[0 0 0]/F 4/A<</S/GoTo/D(ref-shi2023trusting)>>}}%
  \put(505.5,-343.5){\pdfannot width 47.2500bp height 10.5000bp depth 0bp {/Subtype/Link/Border[0 0 0]/F 4/A<</S/GoTo/D(ref-tang2026subtraction)>>}}%
  \put(59.25,-354.75){\pdfannot width 21.7500bp height 10.5000bp depth 0bp {/Subtype/Link/Border[0 0 0]/F 4/A<</S/GoTo/D(ref-tang2026subtraction)>>}}%
  \put(116.25,-354.75){\pdfannot width 81.0000bp height 10.5000bp depth 0bp {/Subtype/Link/Border[0 0 0]/F 4/A<</S/GoTo/D(ref-chuang2023dola)>>}}%
  \put(326.25,-354.75){\pdfannot width 69.7500bp height 10.5000bp depth 0bp {/Subtype/Link/Border[0 0 0]/F 4/A<</S/GoTo/D(ref-zhou2025alw)>>}}%
  \put(489.75,-462.75){\pdfannot width 63.0000bp height 10.5000bp depth 0bp {/Subtype/Link/Border[0 0 0]/F 4/A<</S/GoTo/D(ref-ho2022cfg)>>}}%
  \put(59.25,-474){\pdfannot width 21.7500bp height 10.5000bp depth 0bp {/Subtype/Link/Border[0 0 0]/F 4/A<</S/GoTo/D(ref-ho2022cfg)>>}}%
  \put(87,-474){\pdfannot width 72.7500bp height 10.5000bp depth 0bp {/Subtype/Link/Border[0 0 0]/F 4/A<</S/GoTo/D(ref-chen2022analog)>>}}%
  \put(501,-485.25){\pdfannot width 51.7500bp height 10.5000bp depth 0bp {/Subtype/Link/Border[0 0 0]/F 4/A<</S/GoTo/D(ref-karras2024autoguidance)>>}}%
  \put(59.25,-496.5){\pdfannot width 21.7500bp height 10.5000bp depth 0bp {/Subtype/Link/Border[0 0 0]/F 4/A<</S/GoTo/D(ref-karras2024autoguidance)>>}}%
  \put(147,-507.75){\pdfannot width 85.5000bp height 10.5000bp depth 0bp {/Subtype/Link/Border[0 0 0]/F 4/A<</S/GoTo/D(ref-sanchez2023stay)>>}}%
}
\makeatother

%% file: html-body.tex
\title{Decoding Looped Transformers Better for (Almost) Free}
\author{Weihao Liu Huangjie Zheng Tianrong Chen Rohit Dilip Richard He Bai Yizhu Jiao Yuyang Wang Ruixiang Zhang}
\date{Apple}
\maketitle
\begin{abstract}

Looped Transformers achieve parameter efficiency by repeatedly executing a shared block across recurrent loops. Each loop yields an intermediate representation decodable for the same next token, yet standard decoding discards earlier states. Because earlier loops embody less computation, recurrence inherently supplies aligned weak-and-strong prediction pairs without auxiliary models or external training. We introduce \textbf{LoopCD}, a training-free contrastive decoding framework that guides token selection by contrasting the final prediction with an earlier recurrent pass, operating either in logit space with one extra output pass (LoopCD-Logits) or in hidden-state space with zero output overhead (LoopCD-Hidden). Across four looped Transformer families, LoopCD delivers substantial, consistent gains at full recurrent depth: LoopCD-Logits raises Ouro-2.6B-Thinking's AIME 2024 pass@1 \textbf{from 61.88\% to 73.33\%}, while LoopCD-Hidden lifts Huginn's HumanEval pass@1 \textbf{from 22.56\% to 31.71\%}. Crucially, these performance gains enable halving the number of recurrent loops while still matching or exceeding full-depth unguided baselines, reducing forward FLOPs by \textbf{22.5\% to 48.2\%}. By transforming intermediate recurrent states into effective guidance signals, LoopCD achieves superior decoding quality while substantially reducing inference compute.

\end{abstract}

Correspondence to: Weihao Liu (\href{mailto:wliu681@uic.edu}{wliu681@uic.edu}; work done at Apple internship) Ruixiang Zhang \href{mailto:ruixiangz@apple.com}{(ruixiangz@apple.com)}

\begin{figure}
\centering
\includegraphics[width=\linewidth,alt={LoopCD lets an earlier loop guide the final one: more accuracy at full depth, less compute at half.}]{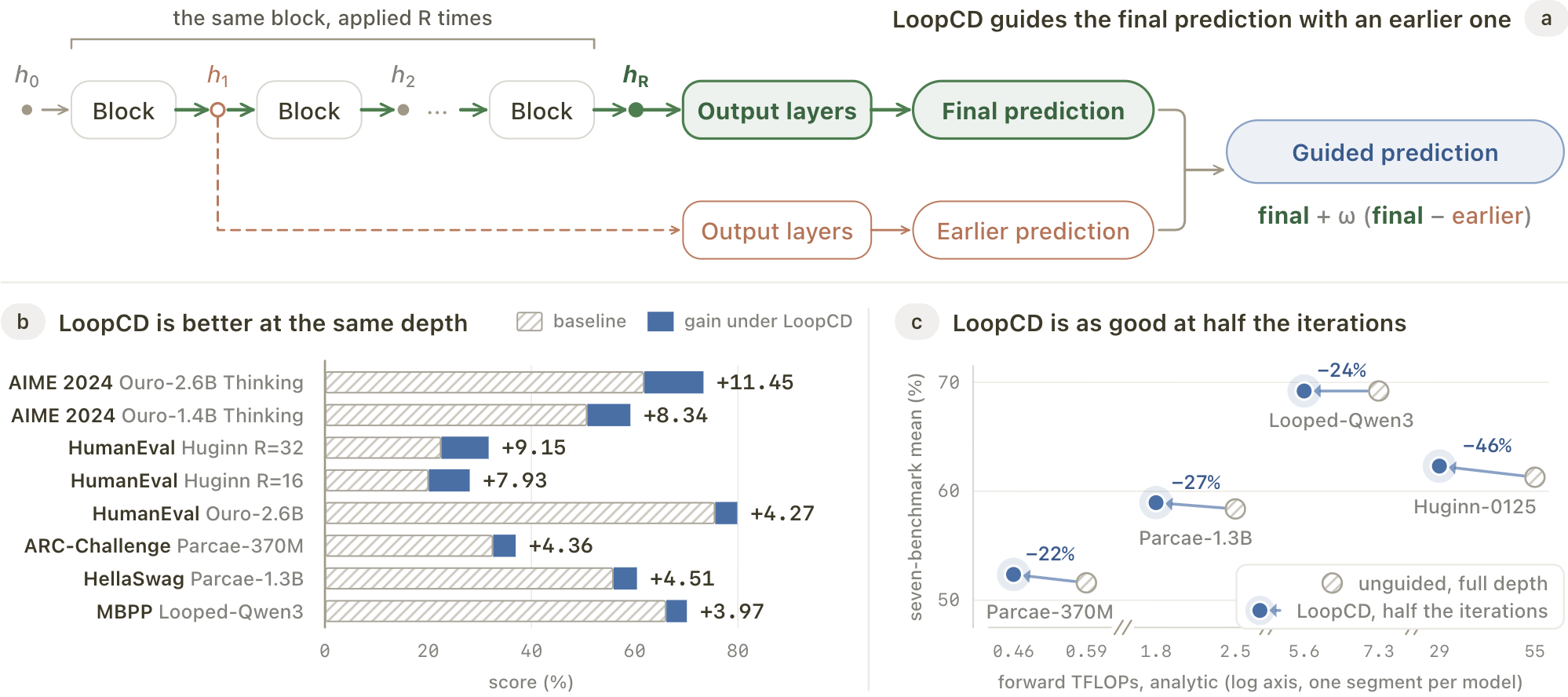}
\caption{\textbf{LoopCD lets an earlier loop guide the final one: more accuracy at full depth, less compute at half.} \textbf{(a)} LoopCD reads an earlier state through the same output layers and uses it to guide the final prediction. \textbf{(b)} Accuracy at full depth without LoopCD (hatched) and the gain with it (solid), for eight model and benchmark pairs; Huginn at \(R=32\) uses hidden-state guidance. \textbf{(c)} Seven-benchmark mean against forward compute, on a log axis broken into one segment per model: the unguided model at full depth (hatched) and LoopCD at half the iterations (filled), with the compute saved; all six settings are in Figure~\hyperref[fig-halfloop]{4}.}\label{fig-hero}
\end{figure}

\section{Introduction}\label{sec-intro}

Looped Transformers offer an appealing architecture for language modeling and reasoning by decoupling depth from parameter count: they repeatedly execute a block of shared layers, allowing effective computation to scale with recurrent iterations without increasing parameter count (\hyperlink{ref-dehghani2019universal}{Dehghani et al., 2019}; \hyperlink{ref-ouro2025}{Zhu et al., 2025a}; \hyperlink{ref-geiping2025huginn}{Geiping et al., 2025}; \hyperlink{ref-parcae2026}{Prairie et al., 2026}). During generation, the model refines its representation across several recurrent passes before emitting a token. Existing methods focus exclusively on the state emitted by the final iteration. However, each intermediate pass also produces a hidden state that can be decoded into a valid token distribution, which standard decoding typically discards.

Contrastive decoding guides a strong predictor away from an aligned weaker reference (\hyperlink{ref-li2022contrastive}{Li et al., 2023}). Similar principles appear in diffusion models, where autoguidance improves sample fidelity by contrasting against a degraded version of the same network (\hyperlink{ref-karras2024autoguidance}{Karras et al., 2024}). In standard language models, obtaining an aligned weak predictor typically requires an auxiliary smaller model, a perturbed context, or an intermediate layer (\hyperlink{ref-li2022contrastive}{Li et al., 2023}; \hyperlink{ref-shi2023trusting}{Shi et al., 2023}; \hyperlink{ref-chuang2023dola}{Chuang et al., 2024}). In contrast, because a looped Transformer repeatedly executes a shared parameter block over depth, each iteration adds compute while operating in the same representation space. The recurrent trajectory thus inherently produces an aligned weak-to-strong prediction pair for the same prefix, without auxiliary models or modified prompts.

We introduce \textbf{LoopCD}, which turns this weak-to-strong trajectory into a direct inference-time decoding method. Because looped Transformers execute the same shared layer block repeatedly, intermediate states lie in the same semantic representation space as the final state and can be decoded by the shared output head. LoopCD contrasts the final prediction with the first iteration to guide token selection along the direction of recurrent refinement, requiring no additional training and no extra recurrent loops. We present two variants of this method: \textbf{LoopCD-Logits}, which evaluates both states through the output layers with one extra output pass, and \textbf{LoopCD-Hidden}, which combines the representations in hidden-state space before the output layers to preserve a single output pass with no extra overhead. We evaluate both representation spaces under two guidance schemes: fixed guidance strength as well as a token-level adaptive guidance method that dynamically scales guidance with prediction uncertainty.

Our experiments demonstrate that LoopCD provides substantial, consistent gains across two distinct computational regimes. At full recurrent depth, LoopCD delivers improvements across all four evaluated model families---Ouro (\hyperlink{ref-ouro2025}{Zhu et al., 2025a}), Huginn (\hyperlink{ref-geiping2025huginn}{Geiping et al., 2025}), Parcae (\hyperlink{ref-parcae2026}{Prairie et al., 2026}), and Looped-Qwen3 (\hyperlink{ref-tflt2026}{Chen et al., 2026})---spanning mathematical reasoning (AIME, OlympiadBench), code generation (HumanEval, MBPP), and multiple-choice benchmarks (MMLU, ARC). LoopCD-Logits lifts Ouro-2.6B-Thinking's (\hyperlink{ref-ouro2025}{Zhu et al., 2025a}) AIME 2024 pass@1 from 61.88\% to 73.33\% (Figure~\hyperref[fig-hero]{1}), while in hidden space, LoopCD-Hidden achieves these gains with zero output overhead, raising Huginn's HumanEval pass@1 from 22.56\% to 31.71\%. Crucially, these substantial performance gains enable reducing the number of recurrent iterations: operating at only half the recurrent depth, LoopCD still matches or exceeds the unguided full-depth baseline across evaluated model families, reducing forward FLOPs by 22.5\% to 48.2\%.

\textbf{Summary of contributions.} This work makes three primary contributions. First, building on the observation that looped Transformers inherently generate aligned weak-and-strong representations across recurrent depth, we propose \textbf{LoopCD}, a training-free contrastive decoding method requiring no auxiliary models or external training. Second, we establish two guidance spaces, showing that LoopCD-Hidden combines representations before the coda layers to eliminate extra output projection overhead while preserving accuracy gains. Third, we demonstrate across four model families that LoopCD consistently improves reasoning and generation at full recurrent depth, and that these substantial gains enable halving recurrent iterations while matching or exceeding full-depth baseline accuracy, reducing forward FLOPs by 22.5\% to 48.2\%.

\section{Preliminaries: Looped Transformers}\label{sec-preliminaries}

A looped Transformer updates its hidden representations by repeatedly applying a block with shared parameters (\hyperlink{ref-dehghani2019universal}{Dehghani et al., 2019}). Each application is a \emph{recurrent iteration}: increasing the iteration count increases effective depth and computation while keeping the stored parameters fixed.

\begin{figure}
\centering
\includegraphics[width=\linewidth,alt={Looped Transformers differ in which part of the network loops.}]{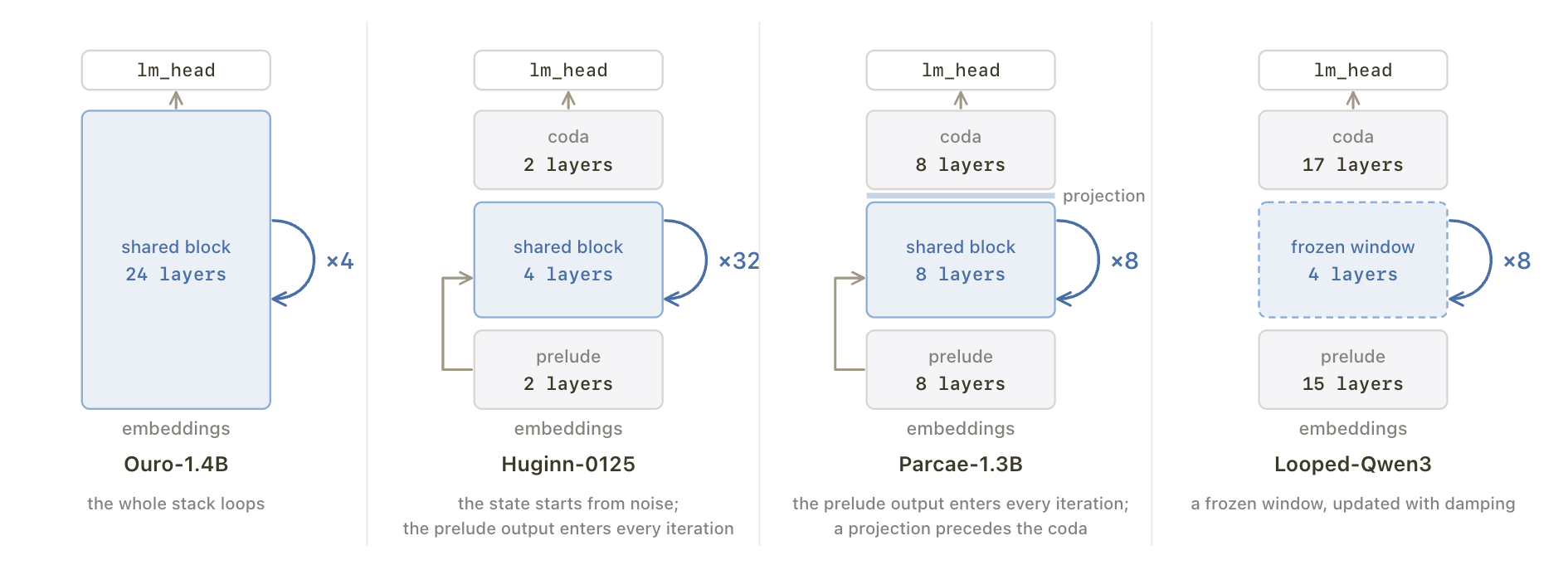}
\caption{\textbf{Looped Transformers differ in which part of the network loops.} Ouro repeats its full stack; Huginn, Parcae, and Looped-Qwen3 repeat a shared block (blue) between fixed prelude and coda layers (gray). Layer and iteration counts are for Ouro-1.4B, Huginn-0125, Parcae-1.3B, and Looped-Qwen3 (Table~\hyperref[tab-arch]{A1}).}\label{fig-looped-background}
\end{figure}

Looped Transformer architectures vary in how they integrate the shared block within the overall network (Figure~\hyperref[fig-looped-background]{2}). Ouro repeats its full decoder stack (\hyperlink{ref-ouro2025}{Zhu et al., 2025a}). Huginn places the recurrent block between input-processing layers, called the \emph{prelude}, and post-loop layers, called the \emph{coda}; its updates also receive a fixed representation from the prelude (\hyperlink{ref-geiping2025huginn}{Geiping et al., 2025}). Parcae and Looped-Qwen3 also retain layers before and after the recurrent block, although their update mechanisms differ: Parcae learns its recurrent dynamics, while Looped-Qwen3 applies damped updates to a frozen middle-layer window (\hyperlink{ref-parcae2026}{Prairie et al., 2026}; \hyperlink{ref-tflt2026}{Chen et al., 2026}). Appendix~\hyperref[app-architecture-models]{A.2} provides architectural details for each evaluated model family.

For a fixed input prefix, let \(h_0\) be the initial state of the recurrent block and \(h_k\) its state after \(k\) iterations. The block applies an update \(f\) for \(R\) iterations, after which the coda layers and language modeling head map the final state \(h_R\) to token logits:

\begin{equation}
\begin{gathered}
h_0 \xrightarrow{\;f\;} h_1 \xrightarrow{\;f\;} \cdots \xrightarrow{\;f\;} h_R,\\
z_R=\operatorname{\mathtt{lm\_head}}\bigl(\operatorname{\mathtt{coda}}(h_R)\bigr).
\end{gathered}\label{eq-recurrence}
\end{equation}

Here, \(\operatorname{\mathtt{lm\_head}}\) consists of the final layer normalization and vocabulary projection. Standard autoregressive decoding samples the next token from \(p_R=\operatorname{softmax}(z_R)\).

\section{LoopCD: Contrastive Decoding from Recurrent States}\label{sec-method}

\textbf{LoopCD} uses the first recurrent state \(\textcolor[HTML]{B37356}{h_1}\) in Eq.~\hyperref[eq-recurrence]{(1)} as a weak reference for the stronger final state \(\textcolor[HTML]{3B6440}{\boldsymbol{h_R}}\). The two states represent the same input prefix after different amounts of recurrent computation. Standard decoding from \(h_R\) is the \emph{unguided} baseline. LoopCD produces a \emph{guided} prediction by extrapolating away from the reference, either in hidden-state space or in logit space. This contrastive update (\hyperlink{ref-li2022contrastive}{Li et al., 2023}) shares the principle of autoguidance and classifier-free guidance in diffusion models: a prediction is guided away from a degraded or unconditional reference (\hyperlink{ref-ho2022cfg}{Ho \& Salimans, 2022}; \hyperlink{ref-karras2024autoguidance}{Karras et al., 2024}).

LoopCD has two design choices: the guidance space (Section~\hyperref[sec-guidance]{3.1}) and the guidance strength \(\omega\) (Section~\hyperref[sec-guidance-strength]{3.2}).

\subsection{Guidance Space}\label{sec-guidance}

\begin{figure}
\centering
\includegraphics[width=\linewidth,alt={LoopCD applies the contrast before or after the coda and language modeling head.}]{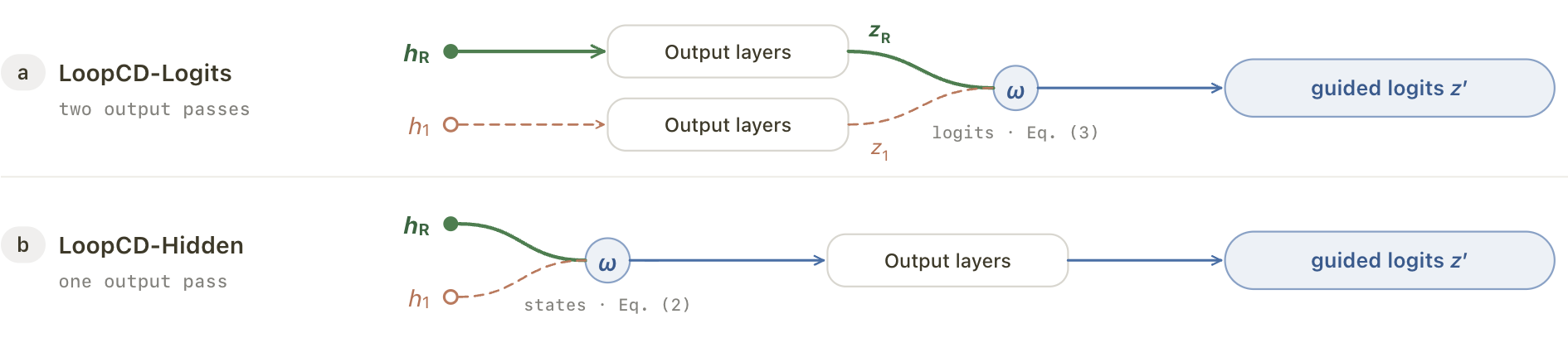}
\caption{\textbf{LoopCD applies the contrast before or after the coda and language modeling head.} LoopCD-Hidden combines \(h_1\) and \(h_R\) (Eq.~\hyperref[eq-hidden-guidance]{(2)}) and evaluates the coda and \texttt{lm\_head} once; LoopCD-Logits evaluates both states through them and combines the logits (Eq.~\hyperref[eq-loopcd]{(3)}), in two passes. Rust marks the reference, sage the final state, steel the guided result.}\label{fig-looped-architecture}
\end{figure}

LoopCD applies the contrast before the coda layers or after the language modeling head (Figure~\hyperref[fig-looped-architecture]{3}).

\textbf{LoopCD-Hidden.} LoopCD combines the reference and final hidden states before the coda layers:

\begin{equation}
\textcolor[HTML]{3A5A8C}{h'}=\textcolor[HTML]{3B6440}{\boldsymbol{h_R}}+\omega\,\bigl(\textcolor[HTML]{3B6440}{\boldsymbol{h_R}}-\textcolor[HTML]{B37356}{h_1}\bigr),\qquad \omega\geq0.\label{eq-hidden-guidance}
\end{equation}

The coefficient \(\omega\) sets the strength of the contrast, with \(\omega=0\) recovering the unguided state \(h_R\). The combined state then passes through the coda layers and language modeling head: \(\textcolor[HTML]{3A5A8C}{z'}=\operatorname{\mathtt{lm\_head}}(\operatorname{\mathtt{coda}}(\textcolor[HTML]{3A5A8C}{h'}))\). The coda layers and \(\operatorname{\mathtt{lm\_head}}\) each execute once, preserving the baseline's single pass through these layers (Section~\hyperref[sec-reduced-loop]{4.3}).

\textbf{LoopCD-Logits.} LoopCD contrasts the logits obtained from the reference and final states after the coda layers and language modeling head:

\begin{equation}
\begin{aligned}
\textcolor[HTML]{B37356}{z_1}&=\operatorname{\mathtt{lm\_head}}\bigl(\operatorname{\mathtt{coda}}(\textcolor[HTML]{B37356}{h_1})\bigr),\qquad
\textcolor[HTML]{3B6440}{\boldsymbol{z_R}}=\operatorname{\mathtt{lm\_head}}\bigl(\operatorname{\mathtt{coda}}(\textcolor[HTML]{3B6440}{\boldsymbol{h_R}})\bigr),\\
\textcolor[HTML]{3A5A8C}{z'}&=\textcolor[HTML]{3B6440}{\boldsymbol{z_R}}+\omega\,\bigl(\textcolor[HTML]{3B6440}{\boldsymbol{z_R}}-\textcolor[HTML]{B37356}{z_1}\bigr),\qquad \omega\geq0.
\end{aligned}\label{eq-loopcd}
\end{equation}

Computing the reference logits \(z_1\) requires one additional pass through the coda layers and \(\operatorname{\mathtt{lm\_head}}\). Appendix~\hyperref[app-probability-guidance]{A.3} gives the equivalent update in probability space. Both forms decode the next token from \(\operatorname{softmax}(z')\).

\subsection{Guidance Strength}\label{sec-guidance-strength}

The coefficient \(\omega\) controls how far guidance extrapolates beyond the unguided prediction.

\textbf{Fixed strength.} A constant \(\omega\) is used for every token within a given evaluation setting (Appendix~\hyperref[app-selection]{B.4}).

\textbf{Adaptive strength.} LoopCD can adjust \(\omega\) at each token using the uncertainty of the unguided prediction. Let \(p_{R,(1)}\) and \(p_{R,(2)}\) be the two largest token probabilities in \(p_R=\operatorname{softmax}(z_R)\). We set

\begin{equation}
\omega=\omega_{\max}\bigl[1-(p_{R,(1)}-p_{R,(2)})\bigr],\label{eq-tokenwise}
\end{equation}

where \(\omega_{\max}\) is the maximum strength. The rule approaches this maximum as the two leading probabilities become equal and reduces guidance as one candidate becomes dominant. The probability margin measures the uncertainty between the leading candidates directly, without computing entropy over the full vocabulary.

LoopCD-Logits already computes the unguided logits \(z_R\) needed by the adaptive rule. Applying the same rule to LoopCD-Hidden would require a coda and head pass on \(h_R\) to obtain \(p_R\), followed by another pass on \(h'\). We therefore use fixed and adaptive strength for LoopCD-Logits, and fixed strength for LoopCD-Hidden.

\section{Experiments}\label{sec-experiments}

We evaluate LoopCD across four looped Transformer families under three evaluation protocols and two computational regimes. First, we evaluate LoopCD at full recurrent depth matching the unguided baselines, establishing substantial performance gains across mathematical reasoning and code generation (Section~\hyperref[sec-main]{4.1}) as well as consistent improvements across multiple-choice benchmarks (Section~\hyperref[sec-generation]{4.2}). Second, given these strong performance gains, we investigate whether LoopCD enables reducing the number of recurrent loops: we show that executing with only half the recurrent loops, LoopCD matches or surpasses full-depth unguided baselines while saving substantial forward FLOPs (Section~\hyperref[sec-reduced-loop]{4.3}).

\textbf{Models and comparisons.} We evaluate Ouro (\hyperlink{ref-ouro2025}{Zhu et al., 2025a}), Huginn (\hyperlink{ref-geiping2025huginn}{Geiping et al., 2025}), Parcae (\hyperlink{ref-parcae2026}{Prairie et al., 2026}), and Looped-Qwen3 (\hyperlink{ref-tflt2026}{Chen et al., 2026}), built from Qwen3-4B (\hyperlink{ref-qwen3}{Qwen Team, 2025}). Guided and unguided runs share checkpoints and prompts; full-depth comparisons also match recurrent depth. Appendix~\hyperref[app-model-settings]{B.1} details model configurations and paired comparisons, and Appendix~\hyperref[app-benchmarks]{B.2} specifies the benchmark protocols.

\subsection{Gains on Mathematical Reasoning and Code Generation}\label{sec-main}

\textbf{LoopCD substantially improves mathematical reasoning pass rates.} Both Ouro-Thinking models improve pass@1 and pass@10 on AIME 2024 (\hyperlink{ref-h42025aime2024}{HuggingFaceH4, 2025}), AIME 2025 (\hyperlink{ref-opencompass2025aime2025}{OpenCompass, 2025}), and OlympiadBench (\hyperlink{ref-he2024olympiadbench}{He et al., 2024}) at the same recurrent depth as their unguided baselines (Table~\hyperref[tab-reason]{1}(a)). For Ouro-2.6B-Thinking, adaptive LoopCD-Logits raises AIME 2024 pass@1 from 61.88\% to 73.33\%, AIME 2025 pass@1 from 49.58\% to 56.88\%, and OlympiadBench pass@1 from 64.05\% to 67.29\%. Fixed and adaptive guidance rules both improve all evaluated benchmarks across pass@1 (gains of 5.27 to 7.33 points) and pass@10 (gains of 2.53 to 4.12 points), with the adaptive rule achieving the highest mean accuracy by concentrating guidance on difficult decision boundaries.

\textbf{On code generation, LoopCD consistently improves execution pass rates across model families and scales.} We evaluate HumanEval (\hyperlink{ref-chen2021humaneval}{Chen et al., 2021}) and MBPP (\hyperlink{ref-austin2021mbpp}{Austin et al., 2021}) with base and extended tests from EvalPlus (\hyperlink{ref-liu2023evalplus}{Liu et al., 2023}) (Table~\hyperref[tab-reason]{1}(b)). For Huginn-0125 at \(R=32\), adaptive LoopCD-Logits lifts HumanEval pass@1 from 23.17\% to 28.66\%, and at \(R=16\) from 20.12\% to 28.05\%. Combining the two hidden states before the output layers achieves even stronger results: at \(R=32\), LoopCD-Hidden raises HumanEval pass@1 from 22.56\% to 31.71\%, and at \(R=16\) matches the adaptive logit score, moving from 21.95\% to 28.05\% (Table~\hyperref[tab-hidden]{3}(b)).

\begin{table}
\caption{\textbf{LoopCD improves mathematical reasoning and code generation at full recurrent depth.} Panel (a): Ouro-Thinking, pass@1 and pass@10 from sixteen samples per problem; subscripts give the change from the baseline. Panel (b): five configurations, base and extended tests; a guided score is red below its baseline.}\label{tab-reason}

\textbf{(a) Mathematical reasoning}\par

\begin{tabular}{lrrrrrrrr}
\toprule
\lxBeginTableHead
\lxTableColumnHead\textbf{} & \multicolumn{2}{c}{\lxTableColumnHead\textbf{AIME 2024}} & \multicolumn{2}{c}{\lxTableColumnHead\textbf{AIME 2025}} & \multicolumn{2}{c}{\lxTableColumnHead\textbf{OlympiadBench}} & \multicolumn{2}{c}{\lxTableColumnHead\textbf{Mean \(\Delta\)}} \\
\lxTableColumnHead\textbf{Model} & \lxTableColumnHead\textbf{Pass@1} & \lxTableColumnHead\textbf{Pass@10} & \lxTableColumnHead\textbf{Pass@1} & \lxTableColumnHead\textbf{Pass@10} & \lxTableColumnHead\textbf{Pass@1} & \lxTableColumnHead\textbf{Pass@10} & \lxTableColumnHead\textbf{Pass@1} & \lxTableColumnHead\textbf{Pass@10} \\
\lxEndTableHead
\midrule
\textcolor[HTML]{83827D}{Ouro-1.4B-Thinking} & \textcolor[HTML]{83827D}{50.83} & \textcolor[HTML]{83827D}{80.33} & \textcolor[HTML]{83827D}{38.96} & \textcolor[HTML]{83827D}{68.09} & \textcolor[HTML]{83827D}{61.27} & \textcolor[HTML]{83827D}{76.69} & \textcolor[HTML]{83827D}{} & \textcolor[HTML]{83827D}{} \\
LoopCD, fixed & 56.88\textsubscript{\textcolor[HTML]{2E7D32}{+6.05}} & 84.82\textsubscript{\textcolor[HTML]{2E7D32}{+4.49}} & 46.25\textsubscript{\textcolor[HTML]{2E7D32}{+7.29}} & 69.23\textsubscript{\textcolor[HTML]{2E7D32}{+1.14}} & 63.75\textsubscript{\textcolor[HTML]{2E7D32}{+2.48}} & 78.66\textsubscript{\textcolor[HTML]{2E7D32}{+1.97}} & \textcolor[HTML]{2E7D32}{+5.27} & \textcolor[HTML]{2E7D32}{+2.53} \\
LoopCD, adaptive & 59.17\textsubscript{\textcolor[HTML]{2E7D32}{+8.34}} & 88.19\textsubscript{\textcolor[HTML]{2E7D32}{+7.86}} & 47.08\textsubscript{\textcolor[HTML]{2E7D32}{+8.12}} & 70.30\textsubscript{\textcolor[HTML]{2E7D32}{+2.21}} & 63.48\textsubscript{\textcolor[HTML]{2E7D32}{+2.21}} & 78.99\textsubscript{\textcolor[HTML]{2E7D32}{+2.30}} & \textbf{\textcolor[HTML]{2E7D32}{+6.22}} & \textbf{\textcolor[HTML]{2E7D32}{+4.12}} \\
\textcolor[HTML]{83827D}{Ouro-2.6B-Thinking} & \textcolor[HTML]{83827D}{61.88} & \textcolor[HTML]{83827D}{82.65} & \textcolor[HTML]{83827D}{49.58} & \textcolor[HTML]{83827D}{75.95} & \textcolor[HTML]{83827D}{64.05} & \textcolor[HTML]{83827D}{78.22} & \textcolor[HTML]{83827D}{} & \textcolor[HTML]{83827D}{} \\
LoopCD, fixed & 70.62\textsubscript{\textcolor[HTML]{2E7D32}{+8.74}} & 88.33\textsubscript{\textcolor[HTML]{2E7D32}{+5.68}} & 58.33\textsubscript{\textcolor[HTML]{2E7D32}{+8.75}} & 81.21\textsubscript{\textcolor[HTML]{2E7D32}{+5.26}} & 66.66\textsubscript{\textcolor[HTML]{2E7D32}{+2.61}} & 79.52\textsubscript{\textcolor[HTML]{2E7D32}{+1.30}} & \textcolor[HTML]{2E7D32}{+6.70} & \textbf{\textcolor[HTML]{2E7D32}{+4.08}} \\
LoopCD, adaptive & 73.33\textsubscript{\textcolor[HTML]{2E7D32}{+11.45}} & 88.33\textsubscript{\textcolor[HTML]{2E7D32}{+5.68}} & 56.88\textsubscript{\textcolor[HTML]{2E7D32}{+7.30}} & 79.53\textsubscript{\textcolor[HTML]{2E7D32}{+3.58}} & 67.29\textsubscript{\textcolor[HTML]{2E7D32}{+3.24}} & 80.08\textsubscript{\textcolor[HTML]{2E7D32}{+1.86}} & \textbf{\textcolor[HTML]{2E7D32}{+7.33}} & \textcolor[HTML]{2E7D32}{+3.71} \\
\bottomrule
\end{tabular}

\textbf{(b) Code generation}\par

\begin{tabular}{lrrrrr}
\toprule
\lxBeginTableHead
\lxTableColumnHead\textbf{} & \multicolumn{2}{c}{\lxTableColumnHead\textbf{HumanEval}} & \multicolumn{2}{c}{\lxTableColumnHead\textbf{MBPP}} & \lxTableColumnHead\textbf{Mean \(\Delta\)} \\
\lxTableColumnHead\textbf{Model} & \lxTableColumnHead\textbf{base} & \lxTableColumnHead\textbf{ext.} & \lxTableColumnHead\textbf{base} & \lxTableColumnHead\textbf{ext.} & \lxTableColumnHead\textbf{} \\
\lxEndTableHead
\midrule
\textcolor[HTML]{83827D}{Ouro-1.4B} & \textcolor[HTML]{83827D}{69.51} & \textcolor[HTML]{83827D}{65.85} & \textcolor[HTML]{83827D}{73.28} & \textcolor[HTML]{83827D}{61.11} & \textcolor[HTML]{83827D}{} \\
LoopCD, fixed & \textcolor[HTML]{2E7D32}{71.95} & \textcolor[HTML]{2E7D32}{66.46} & \textcolor[HTML]{C62828}{72.75} & \textcolor[HTML]{C62828}{60.05} & \textcolor[HTML]{2E7D32}{+0.37} \\
LoopCD, adaptive & \textcolor[HTML]{2E7D32}{72.56} & \textcolor[HTML]{2E7D32}{67.07} & \textcolor[HTML]{2E7D32}{74.07} & \textcolor[HTML]{83827D}{61.11} & \textbf{\textcolor[HTML]{2E7D32}{+1.27}} \\
\textcolor[HTML]{83827D}{Ouro-2.6B} & \textcolor[HTML]{83827D}{75.61} & \textcolor[HTML]{83827D}{70.12} & \textcolor[HTML]{83827D}{77.78} & \textcolor[HTML]{83827D}{64.02} & \textcolor[HTML]{83827D}{} \\
LoopCD, fixed & \textcolor[HTML]{2E7D32}{77.44} & \textcolor[HTML]{2E7D32}{71.34} & \textcolor[HTML]{2E7D32}{78.31} & \textcolor[HTML]{2E7D32}{64.81} & \textcolor[HTML]{2E7D32}{+1.09} \\
LoopCD, adaptive & \textcolor[HTML]{2E7D32}{79.88} & \textcolor[HTML]{2E7D32}{72.56} & \textcolor[HTML]{2E7D32}{79.10} & \textcolor[HTML]{2E7D32}{66.14} & \textbf{\textcolor[HTML]{2E7D32}{+2.54}} \\
\bottomrule
\end{tabular}

\begin{tabular}{lrrrrr}
\toprule
\lxBeginTableHead
\lxTableColumnHead\textbf{} & \multicolumn{2}{c}{\lxTableColumnHead\textbf{HumanEval}} & \multicolumn{2}{c}{\lxTableColumnHead\textbf{MBPP}} & \lxTableColumnHead\textbf{Mean \(\Delta\)} \\
\lxTableColumnHead\textbf{Model} & \lxTableColumnHead\textbf{base} & \lxTableColumnHead\textbf{ext.} & \lxTableColumnHead\textbf{base} & \lxTableColumnHead\textbf{ext.} & \lxTableColumnHead\textbf{} \\
\lxEndTableHead
\midrule
\textcolor[HTML]{83827D}{Huginn, \(R{=}32\)} & \textcolor[HTML]{83827D}{23.17} & \textcolor[HTML]{83827D}{20.12} & \textcolor[HTML]{83827D}{42.06} & \textcolor[HTML]{83827D}{32.54} & \textcolor[HTML]{83827D}{} \\
LoopCD, fixed & \textcolor[HTML]{2E7D32}{27.44} & \textcolor[HTML]{2E7D32}{23.78} & \textcolor[HTML]{2E7D32}{44.97} & \textcolor[HTML]{2E7D32}{37.57} & \textbf{\textcolor[HTML]{2E7D32}{+3.97}} \\
LoopCD, adaptive & \textcolor[HTML]{2E7D32}{28.66} & \textcolor[HTML]{2E7D32}{24.39} & \textcolor[HTML]{C62828}{41.27} & \textcolor[HTML]{2E7D32}{33.86} & \textcolor[HTML]{2E7D32}{+2.57} \\
\textcolor[HTML]{83827D}{Huginn, \(R{=}16\)} & \textcolor[HTML]{83827D}{20.12} & \textcolor[HTML]{83827D}{17.68} & \textcolor[HTML]{83827D}{40.74} & \textcolor[HTML]{83827D}{32.01} & \textcolor[HTML]{83827D}{} \\
LoopCD, fixed & \textcolor[HTML]{2E7D32}{26.83} & \textcolor[HTML]{2E7D32}{20.73} & \textcolor[HTML]{2E7D32}{41.01} & \textcolor[HTML]{2E7D32}{34.13} & \textcolor[HTML]{2E7D32}{+3.04} \\
LoopCD, adaptive & \textcolor[HTML]{2E7D32}{28.05} & \textcolor[HTML]{2E7D32}{25.61} & \textcolor[HTML]{C62828}{39.68} & \textcolor[HTML]{2E7D32}{32.80} & \textbf{\textcolor[HTML]{2E7D32}{+3.90}} \\
\textcolor[HTML]{83827D}{Looped-Qwen3} & \textcolor[HTML]{83827D}{79.88} & \textcolor[HTML]{83827D}{71.95} & \textcolor[HTML]{83827D}{66.14} & \textcolor[HTML]{83827D}{56.08} & \textcolor[HTML]{83827D}{} \\
LoopCD, fixed & \textcolor[HTML]{C62828}{79.27} & \textcolor[HTML]{C62828}{71.34} & \textcolor[HTML]{2E7D32}{70.11} & \textcolor[HTML]{2E7D32}{58.99} & \textbf{\textcolor[HTML]{2E7D32}{+1.42}} \\
LoopCD, adaptive & \textcolor[HTML]{2E7D32}{80.49} & \textcolor[HTML]{2E7D32}{73.78} & \textcolor[HTML]{2E7D32}{67.99} & \textcolor[HTML]{2E7D32}{57.41} & \textcolor[HTML]{2E7D32}{+1.41} \\
\bottomrule
\end{tabular}

\end{table}

\subsection{Consistent Improvements Across Architectures and Suites}\label{sec-generation}

\textbf{LoopCD improves average multiple-choice accuracy across all four model families.} We score candidate answers by likelihood on seven benchmarks: ARC-Challenge and ARC-Easy (\hyperlink{ref-clark2018arc}{Clark et al., 2018}), SciQ (\hyperlink{ref-welbl2017sciq}{Welbl et al., 2017}), MMLU (\hyperlink{ref-hendrycks2021mmlu}{Hendrycks et al., 2021}), HellaSwag (\hyperlink{ref-zellers2019hellaswag}{Zellers et al., 2019}), WinoGrande (\hyperlink{ref-sakaguchi2020winogrande}{Sakaguchi et al., 2020}), and PIQA (\hyperlink{ref-bisk2020piqa}{Bisk et al., 2020}). Every evaluated model improves its benchmark mean under LoopCD-Logits, with the adaptive strength method leading across nearly all configurations (Table~\hyperref[tab-gen]{2}).

\begin{table}
\caption{\textbf{Every logit configuration improves its seven-benchmark mean under LoopCD.} Seven multiple-choice benchmarks, each option scored by likelihood. Baselines in gray, then LoopCD at fixed and adaptive strength; subscripts give the change from the baseline, and bold marks the better rule's mean (Appendix \hyperref[app-benchmarks]{B.2}).}\label{tab-gen}

\begin{tabular}{lrrrrrrrr}
\toprule
\lxBeginTableHead
\lxTableColumnHead\textbf{Model} & \lxTableColumnHead\textbf{ARC-C} & \lxTableColumnHead\textbf{ARC-E} & \lxTableColumnHead\textbf{SciQ} & \lxTableColumnHead\textbf{MMLU} & \lxTableColumnHead\textbf{HellaSwag} & \lxTableColumnHead\textbf{WinoGrande} & \lxTableColumnHead\textbf{PIQA} & \lxTableColumnHead\textbf{Mean \(\Delta\)} \\
\lxEndTableHead
\midrule
\textcolor[HTML]{83827D}{Ouro-1.4B} & \textcolor[HTML]{83827D}{60.41} & \textcolor[HTML]{83827D}{84.09} & \textcolor[HTML]{83827D}{94.70} & \textcolor[HTML]{83827D}{44.45} & \textcolor[HTML]{83827D}{74.48} & \textcolor[HTML]{83827D}{70.80} & \textcolor[HTML]{83827D}{78.51} & \textcolor[HTML]{83827D}{} \\
LoopCD, fixed & 62.29\textsubscript{\textcolor[HTML]{2E7D32}{+1.88}} & 84.39\textsubscript{\textcolor[HTML]{2E7D32}{+0.30}} & 95.30\textsubscript{\textcolor[HTML]{2E7D32}{+0.60}} & 45.35\textsubscript{\textcolor[HTML]{2E7D32}{+0.90}} & 75.30\textsubscript{\textcolor[HTML]{2E7D32}{+0.82}} & 69.85\textsubscript{\textcolor[HTML]{C62828}{\textminus{}0.95}} & 79.05\textsubscript{\textcolor[HTML]{2E7D32}{+0.54}} & \textbf{\textcolor[HTML]{2E7D32}{+0.58}} \\
LoopCD, adaptive & 61.69\textsubscript{\textcolor[HTML]{2E7D32}{+1.28}} & 84.26\textsubscript{\textcolor[HTML]{2E7D32}{+0.17}} & 95.20\textsubscript{\textcolor[HTML]{2E7D32}{+0.50}} & 45.67\textsubscript{\textcolor[HTML]{2E7D32}{+1.22}} & 75.57\textsubscript{\textcolor[HTML]{2E7D32}{+1.09}} & 68.35\textsubscript{\textcolor[HTML]{C62828}{\textminus{}2.45}} & 78.73\textsubscript{\textcolor[HTML]{2E7D32}{+0.22}} & \textcolor[HTML]{2E7D32}{+0.29} \\
\textcolor[HTML]{83827D}{Ouro-2.6B} & \textcolor[HTML]{83827D}{66.21} & \textcolor[HTML]{83827D}{88.01} & \textcolor[HTML]{83827D}{94.20} & \textcolor[HTML]{83827D}{51.09} & \textcolor[HTML]{83827D}{79.52} & \textcolor[HTML]{83827D}{77.66} & \textcolor[HTML]{83827D}{80.25} & \textcolor[HTML]{83827D}{} \\
LoopCD, fixed & 68.60\textsubscript{\textcolor[HTML]{2E7D32}{+2.39}} & 89.10\textsubscript{\textcolor[HTML]{2E7D32}{+1.09}} & 94.60\textsubscript{\textcolor[HTML]{2E7D32}{+0.40}} & 51.82\textsubscript{\textcolor[HTML]{2E7D32}{+0.73}} & 81.07\textsubscript{\textcolor[HTML]{2E7D32}{+1.55}} & 76.87\textsubscript{\textcolor[HTML]{C62828}{\textminus{}0.79}} & 80.74\textsubscript{\textcolor[HTML]{2E7D32}{+0.49}} & \textcolor[HTML]{2E7D32}{+0.84} \\
LoopCD, adaptive & 69.71\textsubscript{\textcolor[HTML]{2E7D32}{+3.50}} & 89.02\textsubscript{\textcolor[HTML]{2E7D32}{+1.01}} & 94.30\textsubscript{\textcolor[HTML]{2E7D32}{+0.10}} & 52.30\textsubscript{\textcolor[HTML]{2E7D32}{+1.21}} & 81.66\textsubscript{\textcolor[HTML]{2E7D32}{+2.14}} & 75.93\textsubscript{\textcolor[HTML]{C62828}{\textminus{}1.73}} & 80.69\textsubscript{\textcolor[HTML]{2E7D32}{+0.44}} & \textbf{\textcolor[HTML]{2E7D32}{+0.95}} \\
\textcolor[HTML]{83827D}{Huginn-0125 \(R{=}32\)} & \textcolor[HTML]{83827D}{37.63} & \textcolor[HTML]{83827D}{64.31} & \textcolor[HTML]{83827D}{90.40} & \textcolor[HTML]{83827D}{33.47} & \textcolor[HTML]{83827D}{66.65} & \textcolor[HTML]{83827D}{60.77} & \textcolor[HTML]{83827D}{75.79} & \textcolor[HTML]{83827D}{} \\
LoopCD, fixed & 39.51\textsubscript{\textcolor[HTML]{2E7D32}{+1.88}} & 64.10\textsubscript{\textcolor[HTML]{C62828}{\textminus{}0.21}} & 91.10\textsubscript{\textcolor[HTML]{2E7D32}{+0.70}} & 33.57\textsubscript{\textcolor[HTML]{2E7D32}{+0.10}} & 69.00\textsubscript{\textcolor[HTML]{2E7D32}{+2.35}} & 61.96\textsubscript{\textcolor[HTML]{2E7D32}{+1.19}} & 74.97\textsubscript{\textcolor[HTML]{C62828}{\textminus{}0.82}} & \textcolor[HTML]{2E7D32}{+0.74} \\
LoopCD, adaptive & 39.42\textsubscript{\textcolor[HTML]{2E7D32}{+1.79}} & 64.69\textsubscript{\textcolor[HTML]{2E7D32}{+0.38}} & 91.20\textsubscript{\textcolor[HTML]{2E7D32}{+0.80}} & 33.61\textsubscript{\textcolor[HTML]{2E7D32}{+0.14}} & 69.03\textsubscript{\textcolor[HTML]{2E7D32}{+2.38}} & 61.33\textsubscript{\textcolor[HTML]{2E7D32}{+0.56}} & 75.79\textsubscript{\textcolor[HTML]{83827D}{+0.00}} & \textbf{\textcolor[HTML]{2E7D32}{+0.86}} \\
\textcolor[HTML]{83827D}{Huginn-0125 \(R{=}16\)} & \textcolor[HTML]{83827D}{37.63} & \textcolor[HTML]{83827D}{64.31} & \textcolor[HTML]{83827D}{90.30} & \textcolor[HTML]{83827D}{33.51} & \textcolor[HTML]{83827D}{66.28} & \textcolor[HTML]{83827D}{60.30} & \textcolor[HTML]{83827D}{75.52} & \textcolor[HTML]{83827D}{} \\
LoopCD, fixed & 39.76\textsubscript{\textcolor[HTML]{2E7D32}{+2.13}} & 64.86\textsubscript{\textcolor[HTML]{2E7D32}{+0.55}} & 90.90\textsubscript{\textcolor[HTML]{2E7D32}{+0.60}} & 33.48\textsubscript{\textcolor[HTML]{C62828}{\textminus{}0.03}} & 68.90\textsubscript{\textcolor[HTML]{2E7D32}{+2.62}} & 59.75\textsubscript{\textcolor[HTML]{C62828}{\textminus{}0.55}} & 74.54\textsubscript{\textcolor[HTML]{C62828}{\textminus{}0.98}} & \textcolor[HTML]{2E7D32}{+0.62} \\
LoopCD, adaptive & 39.33\textsubscript{\textcolor[HTML]{2E7D32}{+1.70}} & 64.94\textsubscript{\textcolor[HTML]{2E7D32}{+0.63}} & 91.20\textsubscript{\textcolor[HTML]{2E7D32}{+0.90}} & 33.46\textsubscript{\textcolor[HTML]{C62828}{\textminus{}0.05}} & 68.82\textsubscript{\textcolor[HTML]{2E7D32}{+2.54}} & 59.67\textsubscript{\textcolor[HTML]{C62828}{\textminus{}0.63}} & 74.86\textsubscript{\textcolor[HTML]{C62828}{\textminus{}0.66}} & \textbf{\textcolor[HTML]{2E7D32}{+0.63}} \\
\textcolor[HTML]{83827D}{Parcae-370M} & \textcolor[HTML]{83827D}{32.59} & \textcolor[HTML]{83827D}{54.42} & \textcolor[HTML]{83827D}{78.60} & \textcolor[HTML]{83827D}{29.34} & \textcolor[HTML]{83827D}{44.10} & \textcolor[HTML]{83827D}{55.09} & \textcolor[HTML]{83827D}{66.97} & \textcolor[HTML]{83827D}{} \\
LoopCD, fixed & 35.15\textsubscript{\textcolor[HTML]{2E7D32}{+2.56}} & 55.05\textsubscript{\textcolor[HTML]{2E7D32}{+0.63}} & 79.40\textsubscript{\textcolor[HTML]{2E7D32}{+0.80}} & 29.58\textsubscript{\textcolor[HTML]{2E7D32}{+0.24}} & 46.66\textsubscript{\textcolor[HTML]{2E7D32}{+2.56}} & 53.59\textsubscript{\textcolor[HTML]{C62828}{\textminus{}1.50}} & 68.01\textsubscript{\textcolor[HTML]{2E7D32}{+1.04}} & \textcolor[HTML]{2E7D32}{+0.90} \\
LoopCD, adaptive & 36.95\textsubscript{\textcolor[HTML]{2E7D32}{+4.36}} & 54.63\textsubscript{\textcolor[HTML]{2E7D32}{+0.21}} & 79.80\textsubscript{\textcolor[HTML]{2E7D32}{+1.20}} & 29.63\textsubscript{\textcolor[HTML]{2E7D32}{+0.29}} & 47.86\textsubscript{\textcolor[HTML]{2E7D32}{+3.76}} & 53.83\textsubscript{\textcolor[HTML]{C62828}{\textminus{}1.26}} & 67.41\textsubscript{\textcolor[HTML]{2E7D32}{+0.44}} & \textbf{\textcolor[HTML]{2E7D32}{+1.29}} \\
\textcolor[HTML]{83827D}{Parcae-1.3B} & \textcolor[HTML]{83827D}{41.21} & \textcolor[HTML]{83827D}{65.07} & \textcolor[HTML]{83827D}{83.70} & \textcolor[HTML]{83827D}{32.63} & \textcolor[HTML]{83827D}{55.93} & \textcolor[HTML]{83827D}{57.85} & \textcolor[HTML]{83827D}{72.03} & \textcolor[HTML]{83827D}{} \\
LoopCD, fixed & 43.86\textsubscript{\textcolor[HTML]{2E7D32}{+2.65}} & 65.91\textsubscript{\textcolor[HTML]{2E7D32}{+0.84}} & 85.80\textsubscript{\textcolor[HTML]{2E7D32}{+2.10}} & 33.19\textsubscript{\textcolor[HTML]{2E7D32}{+0.56}} & 59.19\textsubscript{\textcolor[HTML]{2E7D32}{+3.26}} & 58.72\textsubscript{\textcolor[HTML]{2E7D32}{+0.87}} & 71.44\textsubscript{\textcolor[HTML]{C62828}{\textminus{}0.59}} & \textcolor[HTML]{2E7D32}{+1.38} \\
LoopCD, adaptive & 45.31\textsubscript{\textcolor[HTML]{2E7D32}{+4.10}} & 66.16\textsubscript{\textcolor[HTML]{2E7D32}{+1.09}} & 85.30\textsubscript{\textcolor[HTML]{2E7D32}{+1.60}} & 33.52\textsubscript{\textcolor[HTML]{2E7D32}{+0.89}} & 60.44\textsubscript{\textcolor[HTML]{2E7D32}{+4.51}} & 58.48\textsubscript{\textcolor[HTML]{2E7D32}{+0.63}} & 70.35\textsubscript{\textcolor[HTML]{C62828}{\textminus{}1.68}} & \textbf{\textcolor[HTML]{2E7D32}{+1.59}} \\
\textcolor[HTML]{83827D}{Looped-Qwen3} & \textcolor[HTML]{83827D}{62.12} & \textcolor[HTML]{83827D}{79.25} & \textcolor[HTML]{83827D}{95.30} & \textcolor[HTML]{83827D}{47.97} & \textcolor[HTML]{83827D}{69.14} & \textcolor[HTML]{83827D}{65.90} & \textcolor[HTML]{83827D}{75.19} & \textcolor[HTML]{83827D}{} \\
LoopCD, fixed & 62.54\textsubscript{\textcolor[HTML]{2E7D32}{+0.42}} & 79.00\textsubscript{\textcolor[HTML]{C62828}{\textminus{}0.25}} & 95.00\textsubscript{\textcolor[HTML]{C62828}{\textminus{}0.30}} & 48.81\textsubscript{\textcolor[HTML]{2E7D32}{+0.84}} & 71.39\textsubscript{\textcolor[HTML]{2E7D32}{+2.25}} & 65.75\textsubscript{\textcolor[HTML]{C62828}{\textminus{}0.15}} & 74.65\textsubscript{\textcolor[HTML]{C62828}{\textminus{}0.54}} & \textcolor[HTML]{2E7D32}{+0.32} \\
LoopCD, adaptive & 64.08\textsubscript{\textcolor[HTML]{2E7D32}{+1.96}} & 78.58\textsubscript{\textcolor[HTML]{C62828}{\textminus{}0.67}} & 95.10\textsubscript{\textcolor[HTML]{C62828}{\textminus{}0.20}} & 49.07\textsubscript{\textcolor[HTML]{2E7D32}{+1.10}} & 72.31\textsubscript{\textcolor[HTML]{2E7D32}{+3.17}} & 66.14\textsubscript{\textcolor[HTML]{2E7D32}{+0.24}} & 75.14\textsubscript{\textcolor[HTML]{C62828}{\textminus{}0.05}} & \textbf{\textcolor[HTML]{2E7D32}{+0.79}} \\
\bottomrule
\end{tabular}

\end{table}

LoopCD is robust against different training methods: while Ouro supervises readouts after every pass, Huginn and Parcae sample recurrent depths during pre-training, and Looped-Qwen3 retrofits recurrence onto a frozen non-recurrent model. LoopCD delivers consistent improvements across all four regimes, demonstrating that its benefit does not depend on a specific recurrent training objective. These improvements persist across diverse prompting formats, from zero-shot evaluation on Huginn to twenty-five demonstrations on Ouro, as well as under generative chain-of-thought decoding on MMLU-Pro (\hyperlink{ref-wang2024mmlupro}{Wang et al., 2024}) (Table~\hyperref[tab-mmlupro]{A5}).

\begin{table}
\caption{\textbf{LoopCD-Hidden raises every configuration's mean at no added output pass.} LoopCD-Hidden at fixed strength. Baselines in gray; subscripts give the change from the baseline.}\label{tab-hidden}

\textbf{(a) Multiple-choice scoring}\par

\begin{tabular}{lrrrrrr}
\toprule
\lxBeginTableHead
\lxTableColumnHead\textbf{} & \multicolumn{2}{c}{\lxTableColumnHead\textbf{Huginn-0125\newline{}\(R{=}32\)}} & \multicolumn{2}{c}{\lxTableColumnHead\textbf{Huginn-0125\newline{}\(R{=}16\)}} & \multicolumn{2}{c}{\lxTableColumnHead\textbf{Parcae-1.3B\newline{}\(R{=}8\)}} \\
\lxTableColumnHead\textbf{Benchmark} & \lxTableColumnHead\textbf{Baseline} & \lxTableColumnHead\textbf{LoopCD} & \lxTableColumnHead\textbf{Baseline} & \lxTableColumnHead\textbf{LoopCD} & \lxTableColumnHead\textbf{Baseline} & \lxTableColumnHead\textbf{LoopCD} \\
\lxEndTableHead
\midrule
ARC-C & 37.97 & 38.74\textsubscript{\textcolor[HTML]{2E7D32}{+0.77}} & 37.37 & 37.97\textsubscript{\textcolor[HTML]{2E7D32}{+0.60}} & 40.36 & 42.66\textsubscript{\textcolor[HTML]{2E7D32}{+2.30}} \\
ARC-E & 64.18 & 64.52\textsubscript{\textcolor[HTML]{2E7D32}{+0.34}} & 64.14 & 64.35\textsubscript{\textcolor[HTML]{2E7D32}{+0.21}} & 64.60 & 65.36\textsubscript{\textcolor[HTML]{2E7D32}{+0.76}} \\
HellaSwag & 66.56 & 68.86\textsubscript{\textcolor[HTML]{2E7D32}{+2.30}} & 66.32 & 68.16\textsubscript{\textcolor[HTML]{2E7D32}{+1.84}} & 55.99 & 58.19\textsubscript{\textcolor[HTML]{2E7D32}{+2.20}} \\
WinoGrande & 61.17 & 62.35\textsubscript{\textcolor[HTML]{2E7D32}{+1.18}} & 61.01 & 62.83\textsubscript{\textcolor[HTML]{2E7D32}{+1.82}} & 58.96 & 59.12\textsubscript{\textcolor[HTML]{2E7D32}{+0.16}} \\
PIQA & 75.52 & 75.57\textsubscript{\textcolor[HTML]{2E7D32}{+0.05}} & 75.19 & 75.08\textsubscript{\textcolor[HTML]{C62828}{\textminus{}0.11}} & 72.63 & 71.11\textsubscript{\textcolor[HTML]{C62828}{\textminus{}1.52}} \\
SciQ & 90.30 & 91.00\textsubscript{\textcolor[HTML]{2E7D32}{+0.70}} & 90.00 & 90.70\textsubscript{\textcolor[HTML]{2E7D32}{+0.70}} & 83.80 & 85.30\textsubscript{\textcolor[HTML]{2E7D32}{+1.50}} \\
MMLU & 33.58 & 34.08\textsubscript{\textcolor[HTML]{2E7D32}{+0.50}} & 33.54 & 33.74\textsubscript{\textcolor[HTML]{2E7D32}{+0.20}} & 32.48 & 33.06\textsubscript{\textcolor[HTML]{2E7D32}{+0.58}} \\
Mean \(\Delta\) &  & \textcolor[HTML]{2E7D32}{+0.83} &  & \textcolor[HTML]{2E7D32}{+0.75} &  & \textcolor[HTML]{2E7D32}{+0.85} \\
\bottomrule
\end{tabular}

\textbf{(b) Code generation}\par

\begin{tabular}{lrrrr}
\toprule
\lxBeginTableHead
\lxTableColumnHead\textbf{} & \multicolumn{2}{c}{\lxTableColumnHead\textbf{Huginn-0125\newline{}\(R{=}32\)}} & \multicolumn{2}{c}{\lxTableColumnHead\textbf{Huginn-0125\newline{}\(R{=}16\)}} \\
\lxTableColumnHead\textbf{Benchmark} & \lxTableColumnHead\textbf{Baseline} & \lxTableColumnHead\textbf{LoopCD} & \lxTableColumnHead\textbf{Baseline} & \lxTableColumnHead\textbf{LoopCD} \\
\lxEndTableHead
\midrule
\multicolumn{5}{c}{\textbf{HumanEval}} \\
Base & 22.56 & 31.71\textsubscript{\textcolor[HTML]{2E7D32}{+9.15}} & 21.95 & 28.05\textsubscript{\textcolor[HTML]{2E7D32}{+6.10}} \\
Extended & 19.51 & 29.27\textsubscript{\textcolor[HTML]{2E7D32}{+9.76}} & 18.29 & 23.78\textsubscript{\textcolor[HTML]{2E7D32}{+5.49}} \\
\multicolumn{5}{c}{\textbf{MBPP}} \\
Base & 40.74 & 41.27\textsubscript{\textcolor[HTML]{2E7D32}{+0.53}} & 41.53 & 42.86\textsubscript{\textcolor[HTML]{2E7D32}{+1.33}} \\
Extended & 33.07 & 34.13\textsubscript{\textcolor[HTML]{2E7D32}{+1.06}} & 33.07 & 35.19\textsubscript{\textcolor[HTML]{2E7D32}{+2.12}} \\
Mean \(\Delta\) &  & \textcolor[HTML]{2E7D32}{+5.13} &  & \textcolor[HTML]{2E7D32}{+3.76} \\
\bottomrule
\end{tabular}

\end{table}

\textbf{LoopCD-Hidden provides zero-overhead improvements.} Table~\hyperref[tab-hidden]{3} evaluates the hidden-state form at full recurrent depth. The two states are combined before the output layers, so guidance adds no extra forward computation (Section~\hyperref[sec-guidance]{3.1}). In multiple-choice scoring (Table~\hyperref[tab-hidden]{3}(a)), the seven-benchmark mean rises by \textcolor[HTML]{2E7D32}{+0.83} and \textcolor[HTML]{2E7D32}{+0.75} points on Huginn at \(R=32\) and \(R=16\), exceeding fixed LoopCD-Logits gains (\textcolor[HTML]{2E7D32}{+0.74} and \textcolor[HTML]{2E7D32}{+0.62}, Table~\hyperref[tab-gen]{2}), and by \textcolor[HTML]{2E7D32}{+0.85} points on Parcae-1.3B. In code generation (Table~\hyperref[tab-hidden]{3}(b)), the four-column mean improves by \textcolor[HTML]{2E7D32}{+5.13} points at \(R=32\), ahead of both logit rules (\textcolor[HTML]{2E7D32}{+3.97} and \textcolor[HTML]{2E7D32}{+2.57}, Table~\hyperref[tab-reason]{1}(b)), and by \textcolor[HTML]{2E7D32}{+3.76} at \(R=16\), with every code column improving at both depths. Appendix~\hyperref[app-form-comparison]{D.3} analyzes how much of the logit gain the hidden form retains behind a deeper coda. Across benchmarks, the largest score increases concentrate on code generation and complex reasoning tasks.

\subsection{Fewer Iterations, Matching or Superior Performance}\label{sec-reduced-loop}

\begin{figure}
\centering
\includegraphics[width=\linewidth,alt={With LoopCD, half the recurrent iterations match or beat full depth for less compute.}]{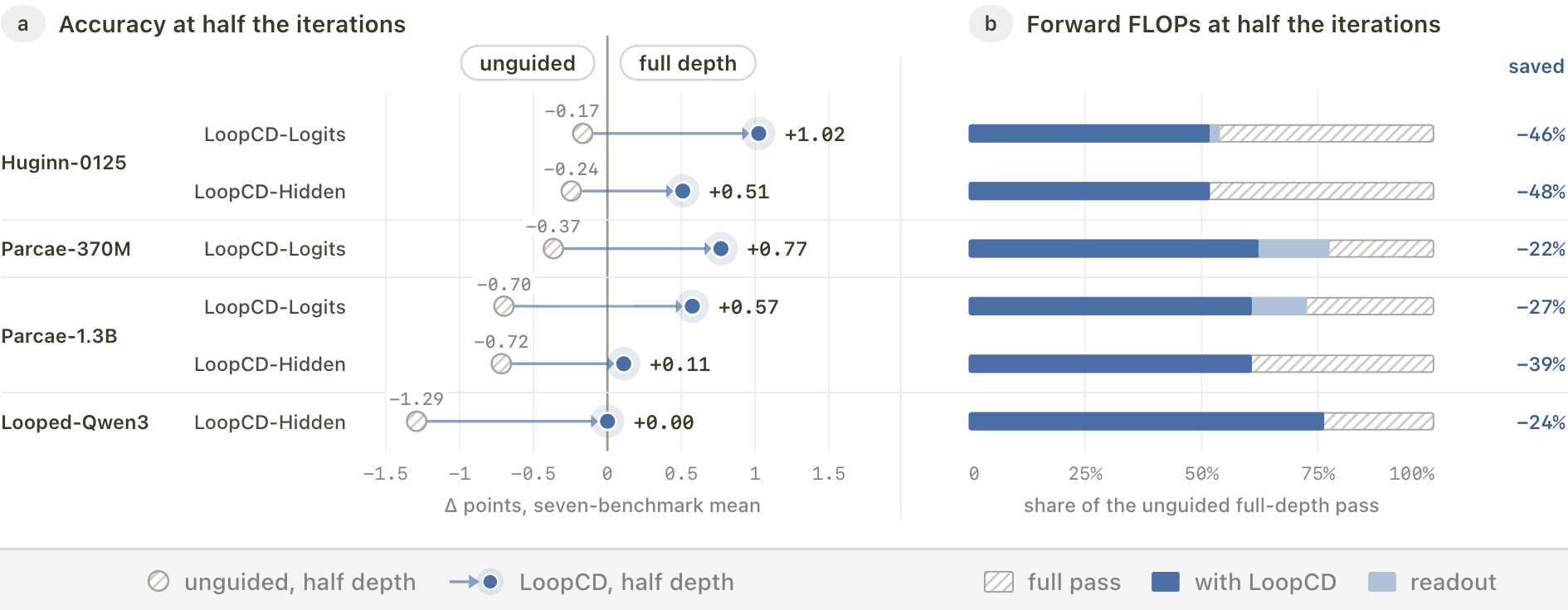}
\caption{\textbf{With LoopCD, half the recurrent iterations match or beat full depth for less compute.} \textbf{(a)} Change in the seven-benchmark mean from the unguided model at full depth: the unguided model at half depth (hatched) and LoopCD at half depth (filled). \textbf{(b)} Forward FLOPs at half depth (solid) as a share of the unguided full-depth pass (hatched); the lighter segment is the readout of logit guidance.}\label{fig-halfloop}
\end{figure}

\textbf{LoopCD can reduce the number of loop iterations while matching or surpassing baseline models at full depth.} Given the performance gains at fixed depth, we investigate whether guidance can substitute for recurrent iterations to reduce inference compute. Reducing recurrent iterations by half incurs an accuracy penalty of 0.17 to 1.29 points on the unguided seven-benchmark multiple-choice mean. Applying LoopCD at this halved depth adds 0.75 to 1.29 points, closing the deficit across all six evaluated settings spanning Huginn, Parcae, and Looped-Qwen3 (Figure~\hyperref[fig-halfloop]{4}). In particular, Huginn-0125 evaluated at sixteen of its thirty-two iterations outperforms its full-depth unguided baseline by 1.02 points under LoopCD-Logits and 0.51 points under LoopCD-Hidden; Looped-Qwen3 matches its full-depth baseline at half iterations under LoopCD-Hidden (+0.00 points).

\textbf{LoopCD shifts the accuracy-compute Pareto frontier, saving up to half the forward FLOPs at equal accuracy.} Figure~\hyperref[fig-halfloop]{4}(b) details forward FLOP requirements as a fraction of the unguided full-depth forward pass, accounting for all guidance overhead. Halving the iterations removes the loop's share of the pass, while LoopCD-Logits re-evaluates the post-loop readout layers once. Consequently, the guided model at halved depth requires only 0.52 to 0.78 of the original forward FLOPs while matching or surpassing full-depth accuracy. Across evaluated reduced-depth settings, LoopCD eliminates 22.5\% to 48.2\% of total forward FLOPs at equal or superior accuracy.

\section{How and Why LoopCD Works}\label{sec-analysis}

Having established that LoopCD improves accuracy across architectures and enables substantial compute reductions, we now investigate the mechanisms underlying these empirical gains.

\subsection{Built-in Weak and Strong Predictions Across Recurrence}\label{sec-weak}

\begin{figure}
\centering
\includegraphics[width=\linewidth,alt={A Looped Transformer's first recurrent iteration is much weaker than its final one.}]{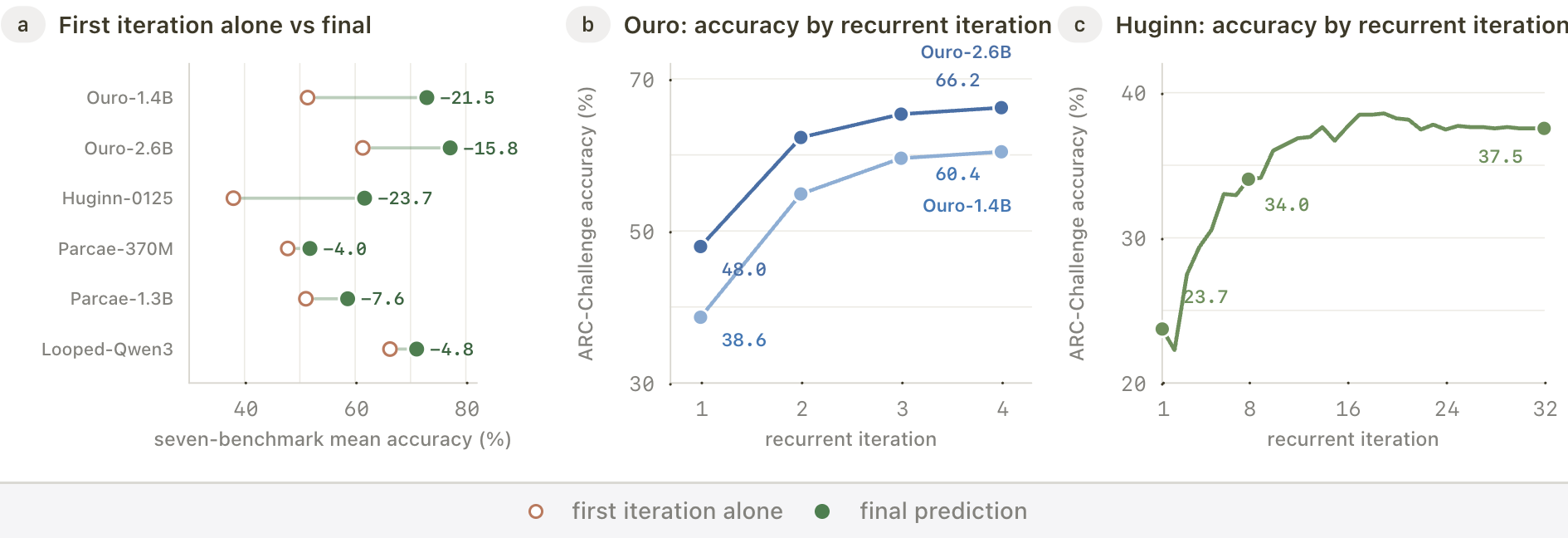}
\caption{\textbf{A Looped Transformer's first recurrent iteration is much weaker than its final one.} \textbf{(a)} After the first iteration (open) against the final one (filled), seven-benchmark mean accuracy, six models. \textbf{(b)} Ouro after each recurrent iteration, ARC-Challenge, 25-shot. \textbf{(c)} Huginn after each of its 32 recurrent iterations, ARC-Challenge, zero-shot.}\label{fig-weak}
\end{figure}

\textbf{Empirical validation of the weak-to-strong recurrence trajectory.} While LoopCD relies on the premise that recurrent passes produce natural weak-to-strong prediction pairs, we empirically validate this progression across model families. Because a looped Transformer repeatedly applies the same parameter block over depth, intermediate recurrent states form aligned, weaker predictors sharing the vocabulary and feature space of the final layer. Evaluated alone, the prediction after the first iteration trails the final converged prediction by 4.0 to 23.7 points on the seven-benchmark mean (Figure~\hyperref[fig-weak]{5}(a)), with accuracy steadily climbing across iterations (Figure~\hyperref[fig-weak]{5}(b, c); Appendix~\hyperref[app-reference-sweep]{E.1}). The recurrent trajectory thus exposes a built-in sequence of progressively stronger models from a single network.

\begin{figure}
\centering
\includegraphics[width=\linewidth,alt={LoopCD gains most where the weak prediction disagrees most with the final prediction.}]{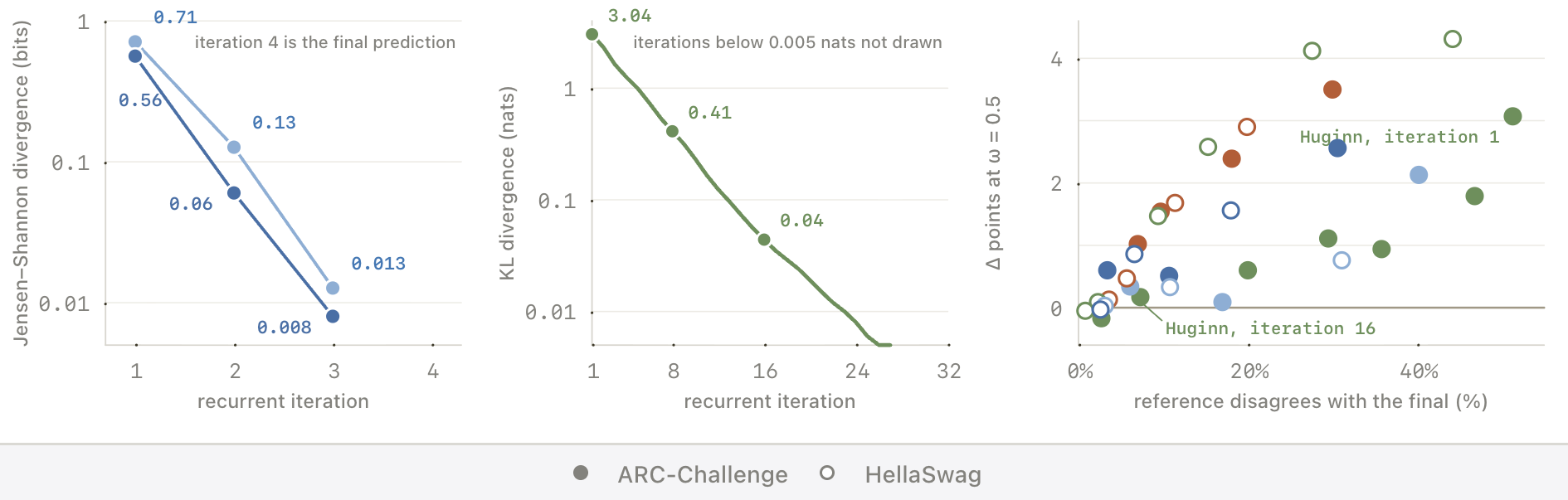}
\caption{\textbf{LoopCD gains most where the weak prediction disagrees most with the final prediction.} \textbf{(a)} Jensen--Shannon divergence (bits) between Ouro's prediction at each recurrent iteration and its final one, ARC-Challenge. \textbf{(b)} Kullback--Leibler divergence (nats) between Huginn's prediction at each recurrent iteration and its final one. \textbf{(c)} The gain of LoopCD-Logits at \(\omega=0.5\) against the share of questions on which the reference alone picks a different option from the final prediction, four models, ARC-Challenge (filled) and HellaSwag (open).}\label{fig-contrast-gain}
\end{figure}

\textbf{LoopCD gains scale directly with how sharply the early reference disagrees with the settled prediction.} Recurrence moves the prediction far in its early iterations and little in its late ones: the divergence from the final prediction falls from 0.71 bits after Ouro-1.4B's first pass to 0.013 after its third, and from 3.0 nats after Huginn's first step to 0.04 after its sixteenth (Figure~\hyperref[fig-contrast-gain]{6}(a, b)). The gain tracks that disagreement (Figure~\hyperref[fig-contrast-gain]{6}(c)): Huginn's first step picks a different option from the final prediction on 51\% of the ARC-Challenge questions and gives 3.07 points at \(\omega=0.5\), its sixteenth disagrees on 7\% and gives 0.17, and Parcae and Ouro run the same course. What the reference supplies is a direction; a late iteration that has converged on the final state leaves little to continue, and a deeper reference needs a larger strength to make up for it.

\begin{figure}
\centering
\includegraphics[width=\linewidth,alt={LoopCD gains most from the first recurrent iteration as its weak prediction.}]{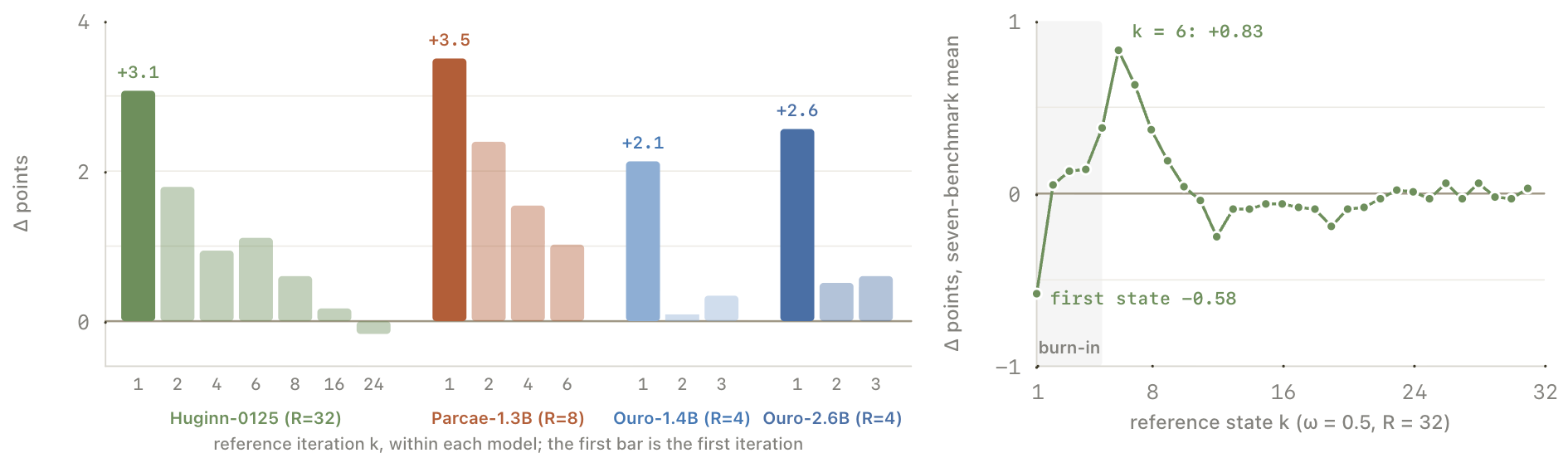}
\caption{\textbf{LoopCD gains most from the first recurrent iteration as its weak prediction.} \textbf{(a)} The gain of LoopCD-Logits at \(\omega=0.5\) on ARC-Challenge by reference iteration \(k\), grouped by model, the first iteration in full colour. \textbf{(b)} Huginn at \(R=32\), LoopCD-Hidden at \(\omega=0.5\): seven-benchmark mean change by reference state, burn-in shaded.}\label{fig-reference}
\end{figure}

\textbf{The first recurrent step provides the most effective logit reference, while Huginn's hidden-state reference requires post-burn-in representations.} Across all swept models, LoopCD-Logits achieves its largest gain using the first recurrent state as reference (e.g., on ARC-Challenge, gains drop from 3.07 to 0.17 points as reference depth increases for Huginn, and from 3.50 to 1.02 for Parcae-1.3B; Figure~\hyperref[fig-reference]{7}(a)). For LoopCD-Hidden, Huginn initializes from Gaussian noise that its coda layers project away for logits but which degrades unprojected hidden contrasts, causing early hidden states to yield \(-0.58\) points before peaking at the sixth step with \(+0.83\) points (Figure~\hyperref[fig-reference]{7}(b)). Parcae and Looped-Qwen3 start from deterministic representations and effectively use their first state in both guidance spaces (Appendix~\hyperref[app-selection]{B.4}).

\subsection{Re-ranking Close Decisions Under Uncertainty}\label{sec-pair-analysis}

\begin{figure}
\centering
\includegraphics[width=\linewidth,alt={The re-ranking carries the gain, and the gain lands where the model is uncertain.}]{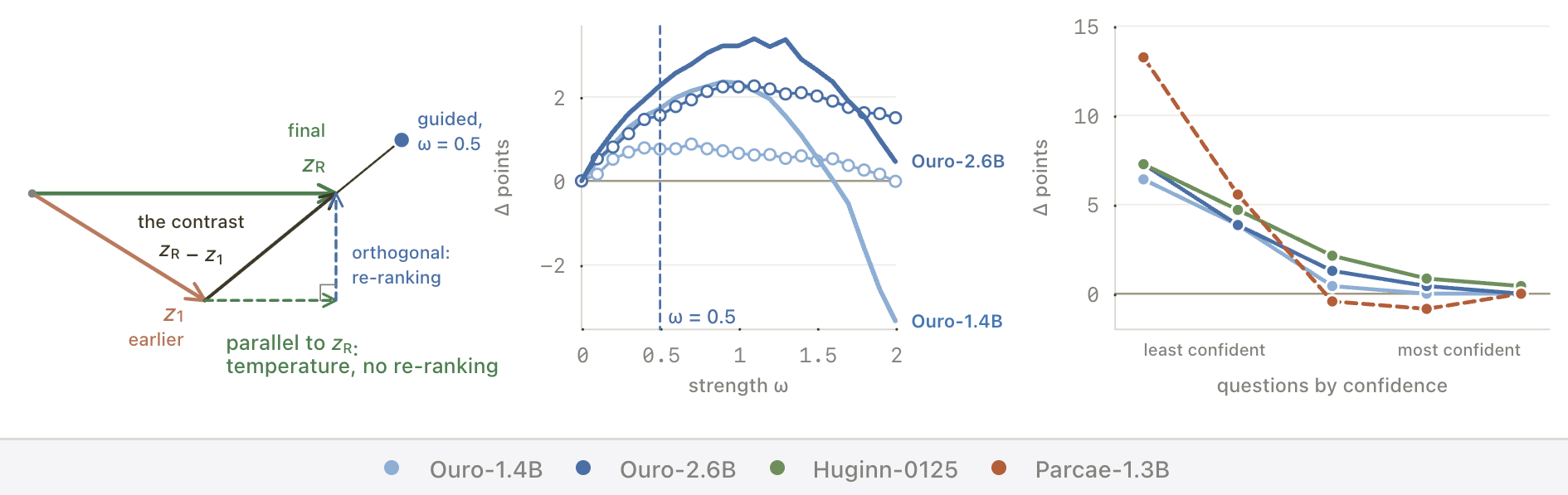}
\caption{\textbf{The re-ranking carries the gain, and the gain lands where the model is uncertain.} \textbf{(a)} An illustrative contrast \(z_R-z_1\) splits into a part parallel to \(z_R\), which rescales the logits like a temperature, and an orthogonal part, which re-ranks tokens. \textbf{(b)} HellaSwag, the two Ouro models: accuracy change as the strength grows, re-ranking part alone (thick) and full update (thin). \textbf{(c)} ARC-Challenge questions in five equal groups by the unguided prediction's confidence, least confident first: accuracy change per group, LoopCD-Logits at \(\omega=0.5\), four models.}\label{fig-decomp}
\end{figure}

\textbf{The contrast vector decomposes into parallel temperature scaling and orthogonal re-ranking, with the re-ranking component driving performance gains.} Projecting the logit contrast \(\textcolor[HTML]{3B6440}{\boldsymbol{z_R}}-\textcolor[HTML]{B37356}{z_1}\) into components parallel and orthogonal to \(\textcolor[HTML]{3B6440}{\boldsymbol{z_R}}\) (Figure~\hyperref[fig-decomp]{8}(a)) separates their functional roles. The parallel component rescales logits uniformly, acting as a temperature adjustment that preserves token order. The orthogonal component alters relative token distances, acting as a pure re-ranking that directly changes top-token selection. On HellaSwag, applying the orthogonal re-ranking component alone achieves gains matching or exceeding the full update up to \(\omega=1.5\), yielding +1.72 points against +0.76 for Ouro-1.4B and +2.27 against +1.56 for Ouro-2.6B at \(\omega=0.5\) (Figure~\hyperref[fig-decomp]{8}(b)). LoopCD-Hidden produces equivalent re-ranking decisions in representation space (Appendix~\hyperref[app-rerank-identity]{F.2}).

\textbf{Guidance helps most where the model is uncertain.} A re-ranking can only change a decision it can reach. The update flips an answer only when the push it adds exceeds the gap between the final prediction's two best options, so confident decisions stay as they are and the undecided ones move. Grouping ARC-Challenge questions by that gap, guidance at \(\omega=0.5\) adds between +6.4 and +13.3 points on the least confident fifth of questions for four models and at most +0.4 on the most confident fifth (Figure~\hyperref[fig-decomp]{8}(c); Appendix~\hyperref[app-close-decisions]{F.1}). Overall, relatively few answers flip (between 5.8\% and 14.9\%), and the net gain of 2.1 to 3.5 points comes strictly from resolving close decisions. This dynamic explains why the adaptive rule of Section~\hyperref[sec-guidance-strength]{3.2} focuses guidance strength where the probability margin is narrow, concentrating updates where they can act.

\subsection{Tuning Guidance Strength Across Scoring and Generation}\label{sec-strength-analysis}

\begin{figure}
\centering
\includegraphics[width=\linewidth,alt={Multiple-choice scoring tolerates about twice the fixed strength that generation does.}]{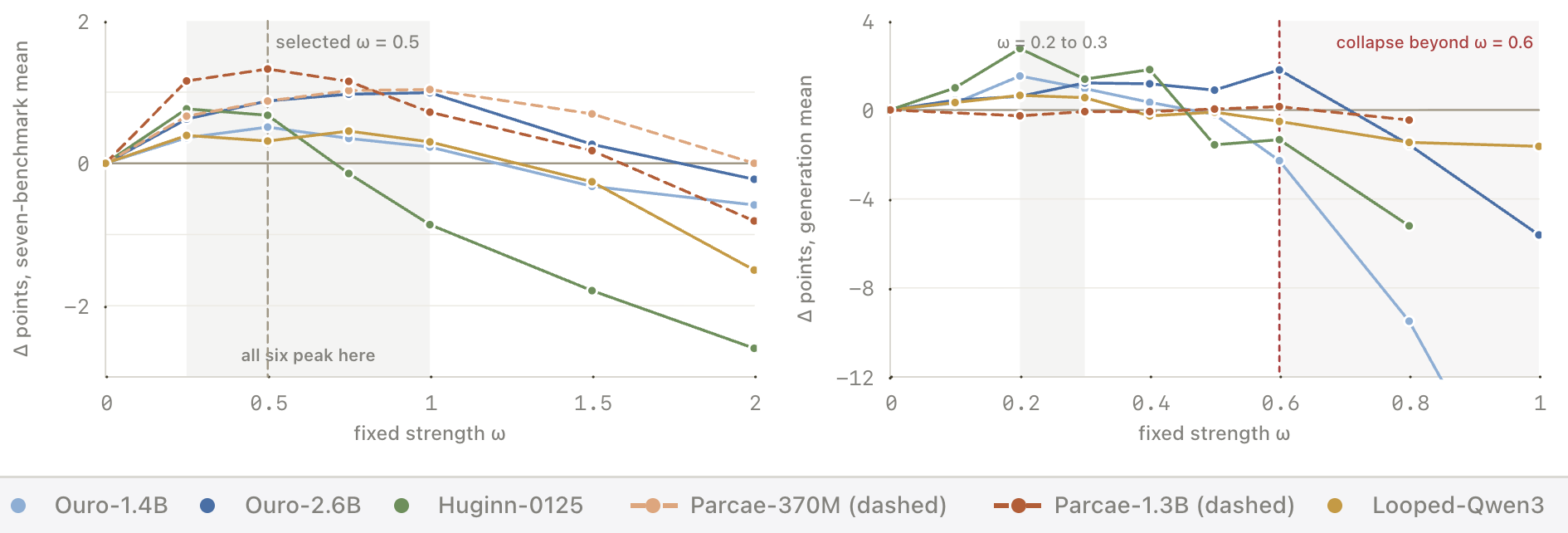}
\caption{\textbf{Multiple-choice scoring tolerates about twice the fixed strength that generation does.} \textbf{(a)} Seven-benchmark mean change against the fixed logit strength, six models; the band spans the strengths at which they peak, and the dashed line marks the selected \(\omega=0.5\). \textbf{(b)} Generation mean change, five models; the band marks the selected \(0.2\) to \(0.3\), and beyond \(\omega\approx0.6\) the curves fall steeply.}\label{fig-strength-sweep}
\end{figure}

\textbf{Multiple-choice scoring tolerates broad guidance strength, whereas autoregressive generation requires smaller values to prevent compounding errors.} For multiple-choice scoring, accuracy gains remain positive across a wide strength band peaking near \(\omega=0.5\) (Figure~\hyperref[fig-strength-sweep]{9}). In contrast, autoregressive generation requires a narrower operating window (\(\omega \in [0.2, 0.3]\)), as early token shifts compound across generated prefixes and cause steep declines at larger strengths. Moving toward the weaker reference (\(\omega < 0\)) universally degrades accuracy across both task families (Appendix~\hyperref[app-strength-sweeps]{G.1}).

\begin{figure}
\centering
\includegraphics[width=\linewidth,alt={The adaptive rule keeps its gain over a wider range of strength than a fixed value.}]{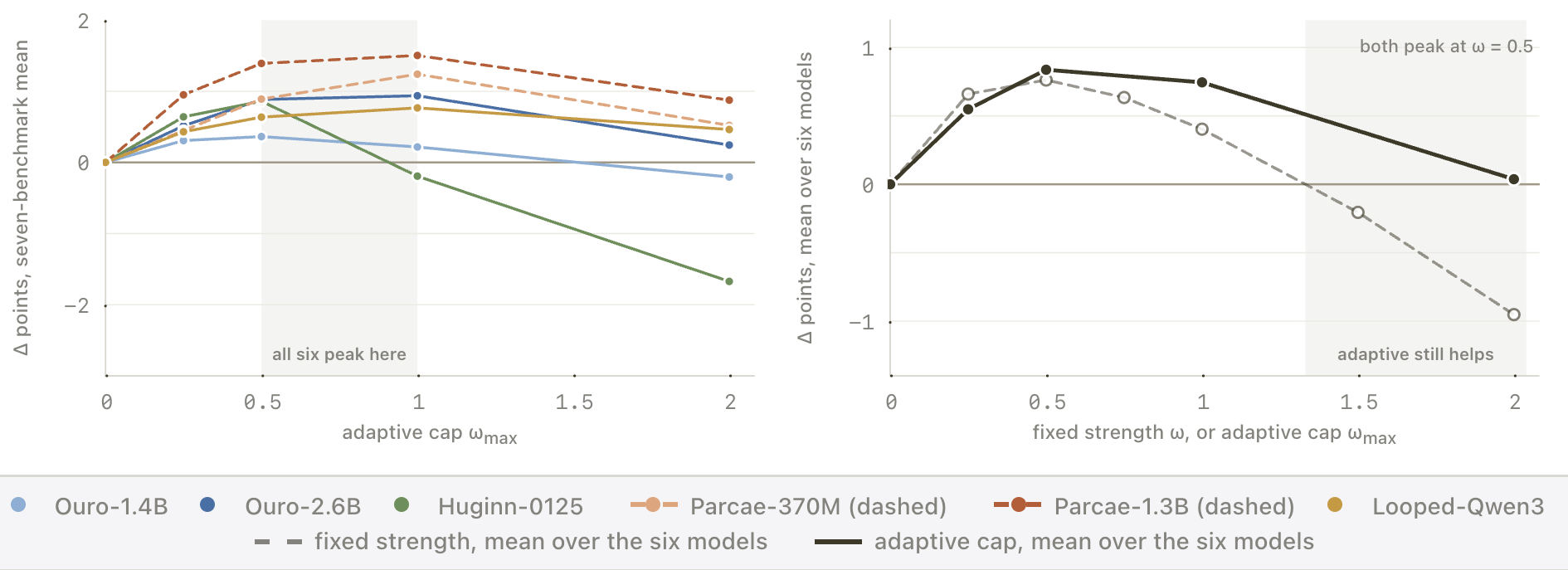}
\caption{\textbf{The adaptive rule keeps its gain over a wider range of strength than a fixed value.} \textbf{(a)} Seven-benchmark mean change against the adaptive cap \(\omega_{\max}\), six models; the band spans the caps at which they peak. \textbf{(b)} Mean over the six models, the fixed rule against its strength (dashed) and the adaptive rule against its cap (solid); the band is where the adaptive mean is still above the baseline and the fixed mean is not.}\label{fig-strength-adaptive}
\end{figure}

\textbf{Adaptive margin gating stabilizes decoding at large strengths by scaling guidance with prediction uncertainty.} By modulating strength via the top-two probability margin (Eq.~\hyperref[eq-tokenwise]{(4)}), adaptive guidance applies maximum strength \(\omega_{\max}\) only to contested tokens and smoothly withholds updates on settled decisions. This margin gating prevents high-confidence drift, maintaining positive accuracy gains across a broad range of caps (\(\omega_{\max} \in [0.5, 1.0]\)) where fixed strength overshoots (Figure~\hyperref[fig-strength-adaptive]{10}; Appendix~\hyperref[app-quintile-reach]{G.2}).

\section{Conclusion}\label{sec-conclusion}

Looped Transformers produce states at different recurrent depths, and those states are the weak and strong predictions that contrastive decoding needs, computed by the same layers for the same token and left unused by ordinary decoding. LoopCD continues the change that recurrence made, from the earlier prediction through the final one, at the cost of one extra output pass or none. At the same depth it raises Ouro's mean pass@1 on mathematical reasoning by five to seven points and improves every multiple-choice and generation suite mean across four families; with half the iterations it matches or exceeds the deeper unguided model in six settings, for 22.5\% to 48.2\% fewer FLOPs. The gain has a simple account: recurrence settles a prediction early and leaves some decisions close, the first iteration's prediction records the direction of that movement, and continuing it re-ranks the close decisions toward the option recurrence was raising. Models trained with their intermediate predictions in mind, as Ouro is, may make this pair more useful still.

\appendix
\setcounter{figure}{0}
\setcounter{table}{0}
\renewcommand{\thefigure}{A\arabic{figure}}
\renewcommand{\thetable}{A\arabic{table}}

\section{Looped Transformers and LoopCD}\label{app-readouts}

This appendix expands Sections~\hyperref[sec-preliminaries]{2} and~\hyperref[sec-method]{3}: the state indexing and output interface behind Eq.~\hyperref[eq-loopcd]{(3)}, the four evaluated families, and the probability form of LoopCD-Logits.

\subsection{States and coda layers}\label{app-output-interface}

LoopCD uses states at the end of complete iterations and the model's existing coda layers. The reference \(h_b\) and final state \(h_{R}\) enter the same output interface. LoopCD-Logits passes both states through this interface and combines the resulting logits. LoopCD-Hidden combines the states first and passes their result through it once. In each case, the modified logits feed the evaluation's existing scoring or token-selection rule.

Looped Transformers with a sandwich structure generally include their prediction head in the coda, whereas we count the coda separately from the final normalization and prediction head to better count the layers. Table~\hyperref[tab-arch]{A1} lists the resulting layer counts. The location of the shared block determines which coda layers must be repeated to obtain earlier logits.

\begin{table}
\caption{\textbf{Layer structure of the evaluated models.} \(\#\mathcal P\), \(\#\mathcal R\), and \(\#\mathcal C\) count prelude, recurrent-block, and coda layers; \(R\) is the iteration count, and \(L_{\mathrm{eff}}=\#\mathcal P+R\#\mathcal R+\#\mathcal C\) counts layer applications. The Ouro-Thinking checkpoints share the Ouro architectures listed; Looped-Qwen3 is built from the dense Qwen3-4B. Layer counts do not give FLOPs, as layers differ in width (Table \hyperref[tab-flops]{A7}).}\label{tab-arch}

\begin{tabular}{lrrrrrr}
\toprule
\lxBeginTableHead
\lxTableColumnHead\textbf{Model} & \lxTableColumnHead\textbf{\# Params} & \lxTableColumnHead\textbf{\(\#\mathcal{P}\)} & \lxTableColumnHead\textbf{\(\#\mathcal{R}\)} & \lxTableColumnHead\textbf{\(\#\mathcal{C}\)} & \lxTableColumnHead\textbf{\(R\)} & \lxTableColumnHead\textbf{\(L_{\mathrm{eff}}\)} \\
\lxEndTableHead
\midrule
Ouro-1.4B & 1.4B & 0 & 24 & 0 & 4 & 96 \\
Ouro-2.6B & 2.6B & 0 & 48 & 0 & 4 & 192 \\
Huginn-0125 (\(R\) = 32) & 3.5B & 2 & 4 & 2 & 32 & 132 \\
Huginn-0125 (\(R\) = 16) & 3.5B & 2 & 4 & 2 & 16 & 68 \\
Parcae-370M & 0.37B & 4 & 4 & 4 & 8 & 40 \\
Parcae-1.3B & 1.3B & 8 & 8 & 8 & 8 & 80 \\
Looped-Qwen3 & 4.0B & 15 & 4 & 17 & 8 & 64 \\
Qwen3-4B & 4.0B & 36 & 0 & 0 & -- & 36 \\
\bottomrule
\end{tabular}

\end{table}

\subsection{Model architectures}\label{app-architecture-models}

\textbf{Ouro.} Ouro repeats its decoder stack and trains predictions at intermediate depths (\hyperlink{ref-ouro2025}{Zhu et al., 2025a}). The earlier state passes through the final normalization and vocabulary head to produce logits. There are no prelude or coda Transformer layers. A hidden-state update could be applied before the normalization and head; our reported Ouro experiments use logits.

\textbf{Huginn.} Huginn computes a prelude representation, initializes a separate recurrent state with Gaussian noise, and conditions each recurrent update on the prelude representation (\hyperlink{ref-geiping2025huginn}{Geiping et al., 2025}). The state denoted \(h_{1}\) in this paper is the result of the first update, and \(h_0\) is the noise initialization. Earlier logits require a pass through the coda and head. LoopCD-Hidden combines recurrent outputs before this pass.

\textbf{Parcae.} Parcae also initializes a separate recurrent state and repeatedly conditions on the normalized prelude representation. Its update includes learned linear terms together with the recurrent Transformer block (\hyperlink{ref-parcae2026}{Prairie et al., 2026}). The recurrent output is projected before entering the coda. LoopCD-Logits applies this projection, coda, and head to the earlier state; the hidden-state form combines states before the projection and uses the coda layers once.

\textbf{Looped-Qwen3.} The Training-Free Looped Transformers construction repeats a frozen window with damped residual updates. A substep has the form \(u_{j+1}=(1-1/K)u_j+(1/K)g(u_j)\), where \(g\) is the frozen window and \(K\) is the number of substeps in the wrapper (\hyperlink{ref-tflt2026}{Chen et al., 2026}). For the evaluated model, the window contains four layers, with fifteen layers before it and seventeen after it. Earlier logits require the frozen tail and head. The reduced-depth result in Figure~\hyperref[fig-halfloop]{4} uses the first four of eight substeps, following the reported experiment; changing the wrapper to four substeps with a different damping coefficient would define a different computation.

\subsection{The probability form of LoopCD-Logits}\label{app-probability-guidance}

Let \(Z_s\) and \(Z_w\) be the softmax normalizers of the final and earlier logits. Since \(\log p_s(v)=z_s(v)-\log Z_s\) and \(\log p_w(v)=z_w(v)-\log Z_w\), the quantity \((1+\omega)\log p_s(v)-\omega\log p_w(v)\) equals the guided logit \(z'(v)\) of Eq.~\hyperref[eq-loopcd]{(3)} plus a constant shared by all vocabulary entries. Exponentiating and normalizing therefore gives the probability form Section~\hyperref[sec-guidance]{3.1} states in words, \(p'(v)\propto p_s(v)\,(p_s(v)/p_w(v))^{\omega}\). The identity holds at each token position for its selected coefficient, including the adaptive rule. It describes the guided logits before any temperature or truncation used by an evaluation protocol. Nonlinear coda layers prevent applying this identity directly to LoopCD-Hidden.

\section{Detailed Experimental Setup}\label{app-protocol}

This appendix further details the evaluated model configurations, benchmark protocols, prompting strategies, and hyperparameter selection.

\subsection{Model configurations and paired comparisons}\label{app-model-settings}

Huginn is evaluated at both \(R=16\) and \(R=32\); the Ouro-Thinking checkpoints are used for mathematical reasoning and the Ouro base checkpoints for the other tasks. Looped-Qwen3 utilizes a frozen Qwen3-4B checkpoint with damped recurrent updates applied to a four-layer middle window (detailed in Appendix~\hyperref[app-architecture-models]{A.2} and Table~\hyperref[tab-arch]{A1}).

To ensure rigorous paired comparisons, a LoopCD run and its unguided baseline always share the identical checkpoint, prompt, shot count, generation limit, stopping rule, answer extractor, and evaluator. Recurrent depths are also strictly matched, except in explicit ablation studies comparing standard and reduced budgets. For the reduced-budget study, standard-budget baselines average repeated logit runs where available, while hidden-state runs are paired with their own dedicated baselines.

\subsection{Benchmarks and table conventions}\label{app-benchmarks}

All reported scores are percentages. A subscript on a guided score denotes its signed percentage-point change (\(\Delta\)) from the strictly matched unguided baseline. The ``Mean~\(\Delta\)'' aggregates these changes with equal weight across the specific columns relevant to each evaluation suite.

\textbf{Multiple-choice benchmarks.} Evaluated on ARC-Challenge, ARC-Easy (\hyperlink{ref-clark2018arc}{Clark et al., 2018}), HellaSwag (\hyperlink{ref-zellers2019hellaswag}{Zellers et al., 2019}), WinoGrande (\hyperlink{ref-sakaguchi2020winogrande}{Sakaguchi et al., 2020}), PIQA (\hyperlink{ref-bisk2020piqa}{Bisk et al., 2020}), SciQ (\hyperlink{ref-welbl2017sciq}{Welbl et al., 2017}), and MMLU (\hyperlink{ref-hendrycks2021mmlu}{Hendrycks et al., 2021}). Candidates are scored by likelihood using the language model evaluation harness (\hyperlink{ref-gao2023lmeval}{Gao et al., 2023}).

\textbf{Greedy generation.} Evaluated on HumanEval (\hyperlink{ref-chen2021humaneval}{Chen et al., 2021}), MBPP (\hyperlink{ref-austin2021mbpp}{Austin et al., 2021}), GSM8K (\hyperlink{ref-cobbe2021gsm8k}{Cobbe et al., 2021}), and MMLU-Pro (\hyperlink{ref-wang2024mmlupro}{Wang et al., 2024}) with greedy decoding. Code generation is assessed via EvalPlus (\hyperlink{ref-liu2023evalplus}{Liu et al., 2023}) using both base and extended tests. GSM8K-CoT uses a strict-match extractor for the \texttt{\#\#\#\# N} format.

\textbf{Mathematical reasoning.} Evaluated on AIME 2024 (\hyperlink{ref-h42025aime2024}{HuggingFaceH4, 2025}), AIME 2025 (\hyperlink{ref-opencompass2025aime2025}{OpenCompass, 2025}) (30 problems each, integer matching), and the 674-problem English mathematics subset of OlympiadBench (\hyperlink{ref-he2024olympiadbench}{He et al., 2024}) (symbolic equivalence). Each AIME set combines parts I and II of that year's examination. We estimate pass@1 and pass@10 separately from 16 sampled solutions per problem using the standard unbiased estimator (\hyperlink{ref-chen2021humaneval}{Chen et al., 2021}; \hyperlink{ref-ouro2025}{Zhu et al., 2025a}).

\subsection{Prompts and sampling}\label{app-prompts}

Tables~\hyperref[tab-fewshot]{A2} and~\hyperref[tab-fewshot-gen]{A3} document the few-shot configurations for multiple-choice scoring and generation. Code benchmarks operate zero-shot while retaining the standard EvalPlus chat instruction. MMLU-Pro uses five chain-of-thought (CoT) demonstrations. GSM8K is evaluated with three demonstrations by default, except for GSM8K-CoT (which uses eight) and all hidden-state generation runs. Mathematical reasoning strictly uses zero-shot sampling.

\begin{table}
\caption{\textbf{Demonstrations per prompt on the multiple-choice benchmarks.} 0 means zero-shot. A guided run always uses its baseline's count, so every change in Tables~\hyperref[tab-gen]{2} and~\hyperref[tab-hidden]{3} is measured at a fixed prompt. Where the two forms differ, as for Looped-Qwen3, compare their changes rather than their absolute scores.}\label{tab-fewshot}

\begin{tabular}{lrrrrrrr}
\toprule
\lxBeginTableHead
\lxTableColumnHead\textbf{Model} & \lxTableColumnHead\textbf{ARC-C} & \lxTableColumnHead\textbf{ARC-E} & \lxTableColumnHead\textbf{HellaSwag} & \lxTableColumnHead\textbf{WinoGrande} & \lxTableColumnHead\textbf{PIQA} & \lxTableColumnHead\textbf{SciQ} & \lxTableColumnHead\textbf{MMLU} \\
\lxEndTableHead
\midrule
\multicolumn{8}{c}{\emph{LoopCD-Logits}} \\
Ouro-1.4B / 2.6B & 25 & 8 & 10 & 5 & 0 & 0 & 5 \\
Huginn-0125 \(R{=}32\) / \(R{=}16\) & 0 & 0 & 0 & 0 & 0 & 0 & 0 \\
Parcae-370M / 1.3B & 25 & 0 & 0 & 0 & 0 & 0 & 5 \\
Looped-Qwen3 & 25 & 0 & 0 & 0 & 0 & 0 & 5 \\
\multicolumn{8}{c}{\emph{LoopCD-Hidden}} \\
Huginn-0125 & 0 & 0 & 0 & 0 & 0 & 0 & 0 \\
Parcae-1.3B & 25 & 0 & 0 & 0 & 0 & 0 & 5 \\
Looped-Qwen3 & 0 & 0 & 0 & 0 & 0 & 0 & 0 \\
\bottomrule
\end{tabular}

\end{table}

\begin{table}
\caption{\textbf{Prompting configurations across generation benchmarks.} \(\dagger\) denotes GSM8K-CoT. Code benchmarks use no worked examples. HumanEval(+) and MBPP(+) include both base and extended tests.}\label{tab-fewshot-gen}

\begin{tabular}{lrrrr}
\toprule
\lxBeginTableHead
\lxTableColumnHead\textbf{Model} & \lxTableColumnHead\textbf{GSM8K} & \lxTableColumnHead\textbf{MMLU-Pro} & \lxTableColumnHead\textbf{HumanEval(+)} & \lxTableColumnHead\textbf{MBPP(+)} \\
\lxEndTableHead
\midrule
\multicolumn{5}{c}{\emph{LoopCD-Logits}} \\
Ouro-1.4B / 2.6B & 3 & 5 & 0 & 0 \\
Huginn-0125 \(R{=}32\) / \(R{=}16\) & 8\(^{\dagger}\) & 5 & 0 & 0 \\
Looped-Qwen3 & 3 & 5 & 0 & 0 \\
\multicolumn{5}{c}{\emph{LoopCD-Hidden}} \\
Huginn-0125 \(R{=}32\) / \(R{=}16\) & -- & -- & 0 & 0 \\
\bottomrule
\end{tabular}

\end{table}

\subsection{Strengths and reference states}\label{app-selection}

Hyperparameters are determined via screening sweeps prior to full-suite evaluations (see Appendix~\hyperref[app-strength-sweeps]{G.1}, Figures~\hyperref[fig-strength-sweep]{9}, and~\hyperref[fig-strength-adaptive]{10}). Table~\hyperref[tab-strengths]{A4} records the selected values for experiments at standard recurrent depth.

Regarding strength-allocation strategies, we evaluate both fixed and adaptive strengths for regular-depth logit-space setups (multiple-choice scoring, greedy generation, and reasoning on Ouro-Thinking and Looped-Qwen3). Conversely, fixed strength alone is applied to all hidden-state configurations and to all experiments operating under a reduced recurrent depth.

For LoopCD-Logits, the fixed strength (\(\omega\)) is \(0.5\) for multiple-choice scoring. For answer and code generation, it is \(0.3\) for Ouro and Looped-Qwen3 and \(0.2\) for Huginn. Reasoning strengths and adaptive caps (\(\omega_{\max}\)) are listed in Table~\hyperref[tab-strengths]{A4}.

For LoopCD-Hidden, the reference state (\(h_b\)) is chosen with the model's initialization in mind (Appendix~\hyperref[app-reference-sweep]{E.1}). Parcae-1.3B has a deterministic prelude, so its contrast uses \(h_{1}\) at \(R=8\) with \(\omega=1.0\). Huginn starts from Gaussian noise; after its burn-in phase, multiple-choice scoring uses \(h_{6}\) at \(R=32\) and \(h_{7}\) at \(R=16\), both with \(\omega=0.5\). Generation uses \(h_{7}\) and \(h_{6}\) at those respective depths, with \(\omega=0.3\). Reduced-depth experiments use the distinct settings in Table~\hyperref[tab-reduced-settings]{A8}.

\begin{table}
\caption{\textbf{Selected hyperparameters for standard-budget experiments.} \(\omega\) is fixed strength, and \(\omega_{\max}\) is the adaptive cap. LoopCD-Hidden rows specify the reference state \(h_b\) to avoid initialization noise; LoopCD-Logits rows use \(h_{1}\). Dashes denote untested modes. Reduced-depth configurations follow Table~\hyperref[tab-reduced-settings]{A8}.}\label{tab-strengths}

\begin{tabular}{lrr}
\toprule
\lxBeginTableHead
\lxTableColumnHead\textbf{Configuration} & \lxTableColumnHead\textbf{Fixed \(\omega\)} & \lxTableColumnHead\textbf{Cap \(\omega_{\max}\)} \\
\lxEndTableHead
\midrule
\multicolumn{3}{c}{\emph{Multiple-choice scoring, LoopCD-Logits}} \\
Ouro-1.4B & 0.5 & 1.0 \\
Ouro-2.6B & 0.5 & 1.0 \\
Huginn-0125 \(R{=}32\) & 0.5 & 0.5 \\
Huginn-0125 \(R{=}16\) & 0.5 & 0.5 \\
Parcae-370M & 0.5 & 1.0 \\
Parcae-1.3B & 0.5 & 1.0 \\
Looped-Qwen3 & 0.5 & 1.0 \\
\multicolumn{3}{c}{\emph{Multiple-choice scoring, LoopCD-Hidden}} \\
Huginn-0125 \(R{=}32\) & 0.5, \(h_b=h_{6}\) & -- \\
Huginn-0125 \(R{=}16\) & 0.5, \(h_b=h_{7}\) & -- \\
Parcae-1.3B \(R{=}8\) & 1.0, \(h_b=h_{1}\) & -- \\
\bottomrule
\end{tabular}

\begin{tabular}{lrr}
\toprule
\lxBeginTableHead
\lxTableColumnHead\textbf{Configuration} & \lxTableColumnHead\textbf{Fixed \(\omega\)} & \lxTableColumnHead\textbf{Cap \(\omega_{\max}\)} \\
\lxEndTableHead
\midrule
\multicolumn{3}{c}{\emph{Generation, LoopCD-Logits}} \\
Ouro-1.4B & 0.3 & 0.5 \\
Ouro-2.6B & 0.3 & 1.0 \\
Huginn-0125 \(R{=}32\) & 0.2 & 0.25 \\
Huginn-0125 \(R{=}16\) & 0.2 & 0.25 \\
Looped-Qwen3 & 0.3 & 0.5 \\
\multicolumn{3}{c}{\emph{Generation, LoopCD-Hidden}} \\
Huginn-0125 \(R{=}32\) & 0.3, \(h_b=h_{7}\) & -- \\
Huginn-0125 \(R{=}16\) & 0.3, \(h_b=h_{6}\) & -- \\
\multicolumn{3}{c}{\emph{Mathematical reasoning, LoopCD-Logits}} \\
Ouro-1.4B-Thinking & 0.5 & 1.0 \\
Ouro-2.6B-Thinking & 0.5 & 1.5 \\
Looped-Qwen3 & 0.4 & 0.5 \\
\bottomrule
\end{tabular}

\end{table}

\section{Extended Results in Full Recurrent Depth}\label{app-results}

This appendix supplements Sections~\hyperref[sec-main]{4.1} and~\hyperref[sec-generation]{4.2} with GSM8K and MMLU-Pro results under LoopCD-Logits and a comparison of Looped-Qwen3's reasoning metrics. The LoopCD-Hidden reference states and strengths are specified in Appendix~\hyperref[app-selection]{B.4}.

\subsection{GSM8K and MMLU-Pro under LoopCD-Logits}\label{app-mmlupro}

Unlike the likelihood-based or code-execution evaluations, GSM8K and MMLU-Pro are scored based on the final generated answer. We evaluate GSM8K using a three-shot setup (eight-shot with a chain of thought for Huginn). For MMLU-Pro, which covers MMLU's subject matter, we use five-shot chain-of-thought prompting, contrasting with Table~\hyperref[tab-gen]{2} where MMLU is scored by likelihood.

Table~\hyperref[tab-mmlupro]{A5} presents the five generation configurations under LoopCD-Logits. MMLU-Pro improves in nine out of the ten rows, gaining up to \(+1.42\) points on Ouro-2.6B under the fixed rule. In contrast, GSM8K exhibits mixed results: it improves in six of the ten rows (by up to \(+2.04\) points on Ouro-1.4B), but declines by up to \(−1.29\) on Huginn-0125 at \(R=32\).

\begin{table}
\caption{\textbf{LoopCD-Logits holds or raises MMLU-Pro in every configuration; GSM8K changes are small and mixed.} Five configurations at full depth. Baselines in gray; subscripts give the change from the baseline (Appendix \hyperref[app-benchmarks]{B.2}).}\label{tab-mmlupro}

\begin{tabular}{lrrrrrr}
\toprule
\lxBeginTableHead
\lxTableColumnHead\textbf{} & \multicolumn{3}{c}{\lxTableColumnHead\textbf{GSM8K}} & \multicolumn{3}{c}{\lxTableColumnHead\textbf{MMLU-Pro}} \\
\lxTableColumnHead\textbf{Model} & \lxTableColumnHead\textbf{Baseline} & \lxTableColumnHead\textbf{LoopCD, fixed} & \lxTableColumnHead\textbf{LoopCD, adaptive} & \lxTableColumnHead\textbf{Baseline} & \lxTableColumnHead\textbf{LoopCD, fixed} & \lxTableColumnHead\textbf{LoopCD, adaptive} \\
\lxEndTableHead
\midrule
Ouro-1.4B & 75.82 & 77.48\textsubscript{\textcolor[HTML]{2E7D32}{+1.66}} & 77.86\textsubscript{\textcolor[HTML]{2E7D32}{+2.04}} & 48.87 & 49.14\textsubscript{\textcolor[HTML]{2E7D32}{+0.27}} & 48.88\textsubscript{\textcolor[HTML]{2E7D32}{+0.01}} \\
Ouro-2.6B & 81.27 & 82.34\textsubscript{\textcolor[HTML]{2E7D32}{+1.07}} & 80.44\textsubscript{\textcolor[HTML]{C62828}{\textminus{}0.83}} & 56.08 & 57.50\textsubscript{\textcolor[HTML]{2E7D32}{+1.42}} & 56.99\textsubscript{\textcolor[HTML]{2E7D32}{+0.91}} \\
Huginn-0125 \(R{=}32\) & 42.91 & 41.62\textsubscript{\textcolor[HTML]{C62828}{\textminus{}1.29}} & 43.06\textsubscript{\textcolor[HTML]{2E7D32}{+0.15}} & 13.59 & 14.50\textsubscript{\textcolor[HTML]{2E7D32}{+0.91}} & 14.23\textsubscript{\textcolor[HTML]{2E7D32}{+0.64}} \\
Huginn-0125 \(R{=}16\) & 38.21 & 39.80\textsubscript{\textcolor[HTML]{2E7D32}{+1.59}} & 38.44\textsubscript{\textcolor[HTML]{2E7D32}{+0.23}} & 13.66 & 14.26\textsubscript{\textcolor[HTML]{2E7D32}{+0.60}} & 14.18\textsubscript{\textcolor[HTML]{2E7D32}{+0.52}} \\
Looped-Qwen3 & 85.29 & 84.15\textsubscript{\textcolor[HTML]{C62828}{\textminus{}1.14}} & 84.23\textsubscript{\textcolor[HTML]{C62828}{\textminus{}1.06}} & 58.86 & 58.85\textsubscript{\textcolor[HTML]{C62828}{\textminus{}0.01}} & 59.32\textsubscript{\textcolor[HTML]{2E7D32}{+0.46}} \\
\bottomrule
\end{tabular}

\end{table}

\subsection{Reasoning metrics: pass@1 and pass@10 across sampled trajectories}\label{app-token-compounding}

The reasoning gains observed in Table~\hyperref[tab-reason]{1} are an order of magnitude larger than the changes on the multiple-choice suite in Table~\hyperref[tab-gen]{2}. The evaluation protocol offers a plausible explanation: a multiple-choice question alters only a single likelihood comparison, whereas a sampled reasoning solution involves hundreds of sequential token decisions, each of which guidance can shift. Furthermore, pass@\(k\) rewards finding a correct answer among the \(k\) generated samples.

This compounding helps explain the divergence in Looped-Qwen3's mathematical-reasoning metrics (Table~\hyperref[tab-qwen-reasoning]{A6}), the only block in the reasoning suite where pass@1 declines while pass@10 rises. On AIME 2024, fixed-strength guidance lowers pass@1 from 64.79 to 61.88: a single sampled solution succeeds less often. Yet pass@10 rises from 82.91 to 86.17---a three-point increase on a thirty-problem set, or roughly one additional problem solved at least once among ten samples. This pattern is consistent with guidance broadening coverage across sampled trajectories even as it reduces per-sample success on some problems; the aggregate pass@\(k\) scores do not directly measure trajectory diversity. For the Ouro-Thinking models, both metrics rise across nearly every configuration.

\begin{table}
\caption{\textbf{On mathematical reasoning, LoopCD lowers Looped-Qwen3's pass@1 and raises its pass@10.} Both metrics from sixteen samples per problem, as in Table \hyperref[tab-reason]{1}(a).}\label{tab-qwen-reasoning}

\begin{tabular}{lrrrrrrrr}
\toprule
\lxBeginTableHead
\lxTableColumnHead\textbf{} & \multicolumn{2}{c}{\lxTableColumnHead\textbf{AIME 2024}} & \multicolumn{2}{c}{\lxTableColumnHead\textbf{AIME 2025}} & \multicolumn{2}{c}{\lxTableColumnHead\textbf{OlympiadBench}} & \multicolumn{2}{c}{\lxTableColumnHead\textbf{Mean \(\Delta\)}} \\
\lxTableColumnHead\textbf{Model} & \lxTableColumnHead\textbf{pass@1} & \lxTableColumnHead\textbf{pass@10} & \lxTableColumnHead\textbf{pass@1} & \lxTableColumnHead\textbf{pass@10} & \lxTableColumnHead\textbf{pass@1} & \lxTableColumnHead\textbf{pass@10} & \lxTableColumnHead\textbf{pass@1} & \lxTableColumnHead\textbf{pass@10} \\
\lxEndTableHead
\midrule
\textcolor[HTML]{83827D}{Looped-Qwen3} & \textcolor[HTML]{83827D}{64.79} & \textcolor[HTML]{83827D}{82.91} & \textcolor[HTML]{83827D}{55.00} & \textcolor[HTML]{83827D}{75.68} & \textcolor[HTML]{83827D}{66.52} & \textcolor[HTML]{83827D}{78.91} & \textcolor[HTML]{83827D}{--} & \textcolor[HTML]{83827D}{--} \\
LoopCD, fixed & 61.88\textsubscript{\textcolor[HTML]{C62828}{\textminus{}2.91}} & 86.17\textsubscript{\textcolor[HTML]{2E7D32}{+3.26}} & 54.58\textsubscript{\textcolor[HTML]{C62828}{\textminus{}0.42}} & 77.42\textsubscript{\textcolor[HTML]{2E7D32}{+1.74}} & 66.52\textsubscript{\textcolor[HTML]{83827D}{+0.00}} & 79.60\textsubscript{\textcolor[HTML]{2E7D32}{+0.69}} & \textcolor[HTML]{C62828}{\textminus{}1.11} & \textcolor[HTML]{2E7D32}{+1.90} \\
LoopCD, adaptive & 61.04\textsubscript{\textcolor[HTML]{C62828}{\textminus{}3.75}} & 86.10\textsubscript{\textcolor[HTML]{2E7D32}{+3.19}} & 53.96\textsubscript{\textcolor[HTML]{C62828}{\textminus{}1.04}} & 81.28\textsubscript{\textcolor[HTML]{2E7D32}{+5.60}} & 66.29\textsubscript{\textcolor[HTML]{C62828}{\textminus{}0.23}} & 79.65\textsubscript{\textcolor[HTML]{2E7D32}{+0.74}} & \textcolor[HTML]{C62828}{\textminus{}1.67} & \textcolor[HTML]{2E7D32}{+3.18} \\
\bottomrule
\end{tabular}

\end{table}

\section{Computation and Cost Analysis}\label{app-cost}

This appendix expands upon Section~\hyperref[sec-reduced-loop]{4.3}, detailing the general FLOP accounting for the two guidance forms, followed by a comprehensive evaluation of the reduced-depth settings, covering both their performance recovery and computational tradeoffs. Finally, we examine the underlying representational dynamics that differentiate LoopCD-Logits and LoopCD-Hidden.

\subsection{FLOP accounting}\label{app-flop-accounting}

We calculate theoretical computational cost by assuming two FLOPs per multiply-and-accumulate operation in matrix products during a 512-token prefill. The unguided baseline reflects the standard output computation. LoopCD-Logits introduces model-specific overheads to process the reference state. Specifically, Ouro requires normalization and the language modeling head. Huginn adds the coda and head. Parcae utilizes the projection, coda, and head. Finally, Looped-Qwen3 applies the frozen tail and head. Table~\hyperref[tab-flops]{A7} summarizes the resulting computational multipliers and their architectural breakdown.

Conversely, LoopCD-Hidden only introduces lightweight vector operations and relies on a single standard coda pass. Consequently, its computational multiplier effectively rounds to \(1.000\times\) in this accounting, representing negligible overhead.

\begin{table}
\caption{\textbf{LoopCD-Logits adds computational overhead based on post-loop layers, whereas LoopCD-Hidden adds negligible cost.} Forward TFLOPs represent one unguided 512-token prefill. The logit multiplier is relative to this baseline, with the extra work divided between the head and the coda/frozen tail applied to the reference state. LoopCD-Hidden relies solely on vector operations before a single output pass, yielding a \(1.000\times\) multiplier.}\label{tab-flops}

\begin{tabular}{lrrrrrr}
\toprule
\lxBeginTableHead
\lxTableColumnHead\textbf{} & \lxTableColumnHead\textbf{} & \multicolumn{3}{c}{\lxTableColumnHead\textbf{LoopCD-Logits}} & \lxTableColumnHead\textbf{} & \lxTableColumnHead\textbf{} \\
\lxTableColumnHead\textbf{Model} & \lxTableColumnHead\textbf{Forward\newline{}TFLOPs} & \lxTableColumnHead\textbf{FLOPs\newline{}\(\times\)} & \lxTableColumnHead\textbf{head\newline{}(\%)} & \lxTableColumnHead\textbf{coda or tail\newline{}(\%)} & \lxTableColumnHead\textbf{Layers after\newline{}the loop} & \lxTableColumnHead\textbf{LoopCD-Hidden\newline{}\(\times\)} \\
\lxEndTableHead
\midrule
Ouro-1.4B & 5.360 & 1.019 & 1.92 & 0.00 & 0 & 1.000 \\
Ouro-2.6B & 10.617 & 1.010 & 0.97 & 0.00 & 0 & 1.000 \\
Huginn-0125 \(R{=}32\) & 54.526 & 1.022 & 0.65 & 1.51 & 2 & 1.000 \\
Huginn-0125 \(R{=}16\) & 28.261 & 1.042 & 1.25 & 2.90 & 2 & 1.000 \\
Parcae-370M & 0.593 & 1.152 & 5.80 & 9.42 & 4 & 1.000 \\
Parcae-1.3B & 2.500 & 1.119 & 2.06 & 9.79 & 8 & 1.000 \\
Looped-Qwen3 & 7.287 & 1.306 & 5.47 & 25.11 & 17 & 1.000 \\
\bottomrule
\end{tabular}

\end{table}

Note that these FLOP counts reflect theoretical arithmetic workload rather than end-to-end wall-clock speedups. Additionally, the specific reduced-depth experiments discussed subsequently are evaluated exclusively on the multiple-choice suite.

\begin{table}
\caption{\textbf{Settings of the six half-depth comparisons in Figure \hyperref[fig-halfloop]{4}.} Each guided model runs half the recurrent iterations of its full-depth baseline, on the multiple-choice benchmarks; Looped-Qwen3 exits after four of its eight damped substeps and keeps the coefficient \(1/8\). \(h_{1}\) is the state after the first iteration. The strength is fixed across benchmarks within each setting.}\label{tab-reduced-settings}

\begin{tabular}{lrrrr}
\toprule
\lxBeginTableHead
\lxTableColumnHead\textbf{Model} & \lxTableColumnHead\textbf{Guidance} & \lxTableColumnHead\textbf{Full \(\to\) reduced} & \lxTableColumnHead\textbf{Reference} & \lxTableColumnHead\textbf{Fixed \(\omega\)} \\
\lxEndTableHead
\midrule
Huginn-0125 & Logit & 32 → 16 & \(h_{5}\) & 0.5 \\
Huginn-0125 & Hidden-state & 32 → 16 & \(h_{6}\) & 0.5 \\
Parcae-370M & Logit & 8 → 4 & \(h_{1}\) & 0.75 \\
Parcae-1.3B & Logit & 8 → 4 & \(h_{1}\) & 0.5 \\
Parcae-1.3B & Hidden-state & 8 → 4 & \(h_{1}\) & 0.75 \\
Looped-Qwen3 & Hidden-state & 8 → 4 & \(h_{1}\) & 1.0 \\
\bottomrule
\end{tabular}

\end{table}

\subsection{Reduced-depth evaluation: performance recovery and computational tradeoffs}\label{app-reduced-details}

Having established the baseline computational overheads, we now evaluate configurations where the recurrent iteration budget is strictly halved. Table~\hyperref[tab-reduced-settings]{A8} details the specific operational parameters for the experiments presented in Figure~\hyperref[fig-halfloop]{4}. To ensure a rigorous evaluation, the fixed strength for each model is determined uniformly across the multiple-choice suite and held constant for all benchmarks. Because halving the iteration budget inherently alters the models' recurrent dynamics, these operational parameters can differ from their full-depth counterparts. For instance, to accommodate the shortened trajectory, LoopCD-Logits on Huginn utilizes a later reference state (\(h_{5}\)) compared to the initial state (\(h_{1}\)) used at full depth.

As reported in Section~\hyperref[sec-reduced-loop]{4.3}, halving the recurrent iterations degrades the unguided baseline, but LoopCD successfully recovers this deficit across all evaluated configurations.

\begin{figure}
\centering
\includegraphics[width=\linewidth,alt={Half depth saves the loop's share of the forward pass, and LoopCD-Logits pays for the layers after it.}]{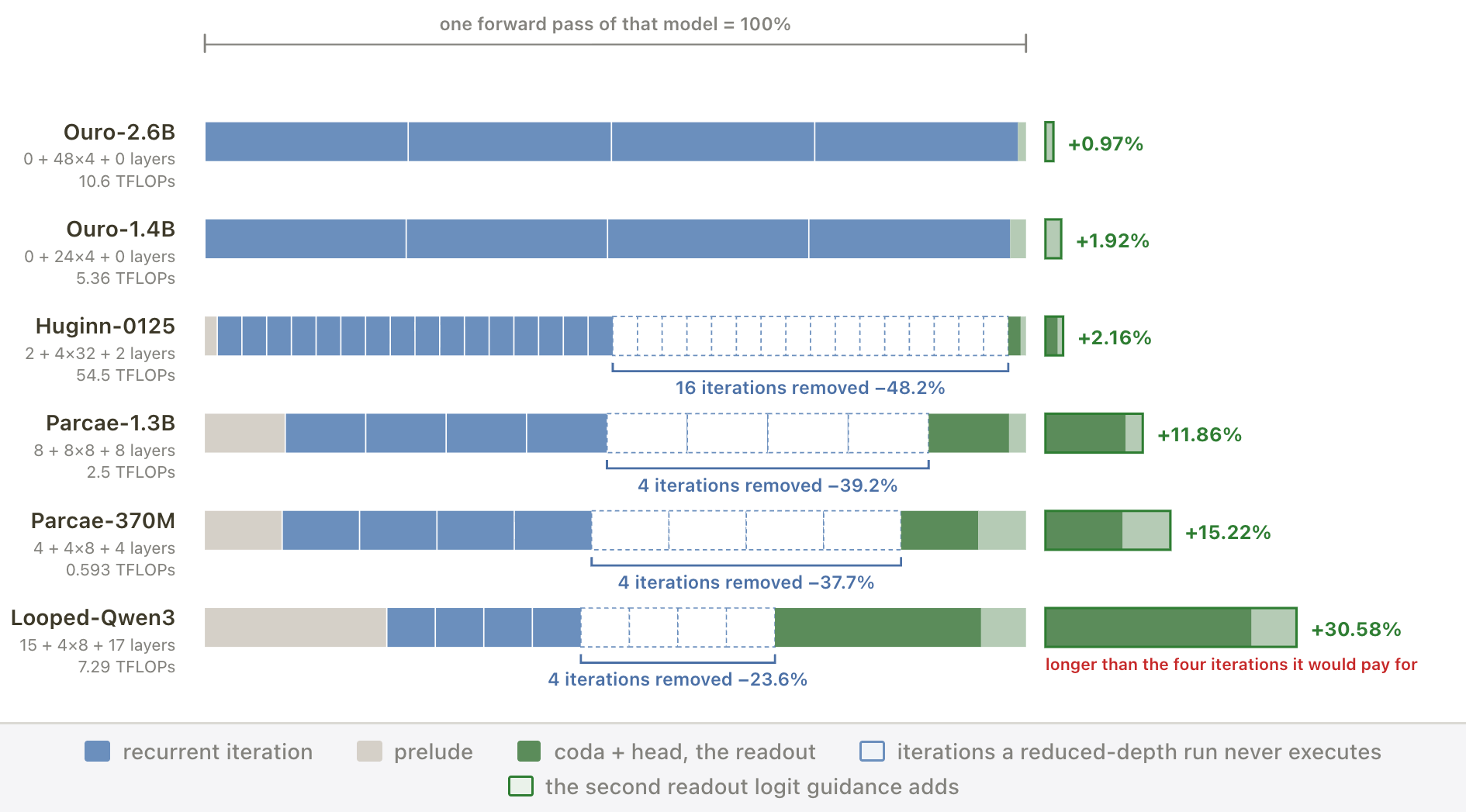}
\caption{\textbf{Half depth saves the loop's share of the forward pass, and LoopCD-Logits pays for the layers after it.} One bar per configuration, scaled to its unguided full-depth pass: prelude, recurrent iterations, then coda and head. Dashed blocks are the iterations a half-depth run skips; the outlined block at the right is LoopCD-Logits' second coda and head. LoopCD-Hidden adds no visible block. Analytic FLOPs at a 512-token prefill.}\label{fig-anatomy}
\end{figure}

To support the computational tradeoff analysis in the main text, Figure~\hyperref[fig-anatomy]{A1} visualizes the structural breakdown of each model's forward pass. Because depth reduction only removes recurrent blocks, the potential savings are strictly bounded by the loop's relative share of the total computation. Conversely, the LoopCD-Logits overhead is dictated entirely by the non-recurrent layers following the loop (the coda and prediction head). Figure~\hyperref[fig-anatomy]{A1} physically illustrates why fully recurrent architectures like Ouro and Huginn benefit immensely from LoopCD-Logits at reduced depths, whereas retrofitted architectures like Looped-Qwen3, which possess massive non-recurrent components, experience readout overheads that exceed their loop savings, necessitating LoopCD-Hidden for effective FLOP reduction.

Table~\hyperref[tab-fractions]{A9} provides the exact computational expenditures for all six reduced-depth settings analyzed in Figure~\hyperref[fig-halfloop]{4}, detailing the specific fractional FLOP costs that define the shifted Pareto frontier.

\begin{table}
\caption{\textbf{Each half-depth setting spends about half to three quarters of the full-depth forward pass, guidance included.} Analytic forward FLOPs at a 512-token prefill, as a fraction of the same checkpoint's unguided full-depth pass; logit settings include their second readout.}\label{tab-fractions}

\begin{tabular}{lrrr}
\toprule
\lxBeginTableHead
\lxTableColumnHead\textbf{Setting} & \lxTableColumnHead\textbf{Iterations} & \lxTableColumnHead\textbf{Forward FLOPs at half depth,\newline{}guidance included (fraction of the full pass)} & \lxTableColumnHead\textbf{Removed\newline{}(\%)} \\
\lxEndTableHead
\midrule
Huginn-0125, logits & 32 → 16 & 0.54 & 46.0\% \\
Huginn-0125, hidden states & 32 → 16 & 0.52 & 48.2\% \\
Parcae-370M, logits & 8 → 4 & 0.78 & 22.5\% \\
Parcae-1.3B, logits & 8 → 4 & 0.73 & 27.3\% \\
Parcae-1.3B, hidden states & 8 → 4 & 0.61 & 39.2\% \\
Looped-Qwen3, hidden states & 8 → 4 & 0.76 & 23.6\% \\
\bottomrule
\end{tabular}

\end{table}

\subsection{Representational dynamics: LoopCD-Logits versus LoopCD-Hidden}\label{app-form-comparison}

While the preceding subsections establish the computational tradeoffs and performance recovery of the two guidance forms, they also differ fundamentally in their representational dynamics. Specifically, they differ in how effectively performance gains are retained across different model architectures, and in how sensitive the final output distribution is to the guidance strength.

\begin{figure}
\centering
\includegraphics[width=\linewidth,alt={LoopCD-Hidden keeps most of the logit form's gain behind a shallow coda and exceeds it in code generation.}]{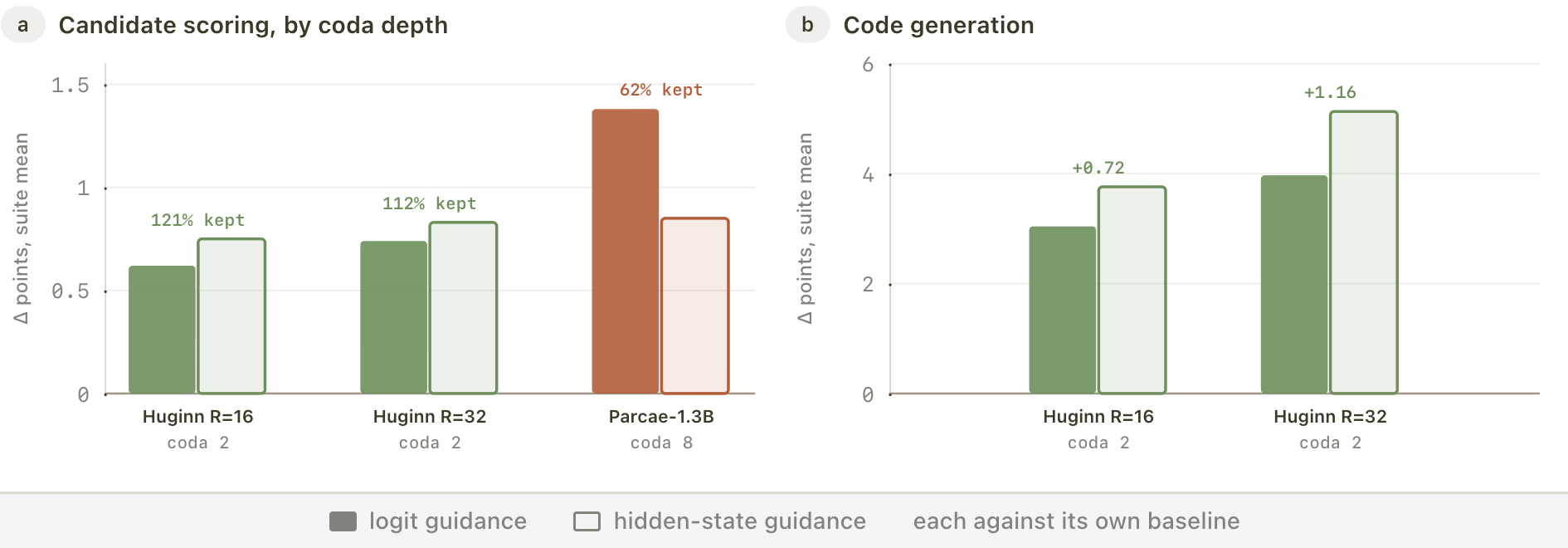}
\caption{\textbf{LoopCD-Hidden keeps most of the logit form's gain behind a shallow coda and exceeds it in code generation.} Mean change at the selected settings of Tables \hyperref[tab-gen]{2} and \hyperref[tab-hidden]{3}, LoopCD-Logits (solid) against LoopCD-Hidden (outlined), each against its own unguided model. \textbf{(a)} Seven-benchmark mean, three configurations in order of coda depth, with the share of the logit gain the hidden-state form keeps. \textbf{(b)} Four-column code-generation mean, Huginn at both depths.}\label{fig-space-tradeoff}
\end{figure}

\textbf{Impact of coda depth on hidden-state retention.} Comparing Table~\hyperref[tab-hidden]{3} to Table~\hyperref[tab-gen]{2} reveals that LoopCD-Hidden retains performance proportional to the depth of the architecture's coda (Figure~\hyperref[fig-space-tradeoff]{A2}(a)). For Huginn, which has a shallow two-layer coda, LoopCD-Hidden achieves nearly identical gains to fixed LoopCD-Logits on the seven-benchmark mean. Specifically, it yields 0.83 versus 0.74 points at \(R=32\), and 0.75 versus 0.62 points at \(R=16\). Conversely, on Parcae-1.3B, which features a deeper eight-layer coda, LoopCD-Hidden retains only roughly sixty percent of the logit gain (0.85 versus 1.38).

Ouro presents the opposite extreme, as it possesses zero layers after the recurrent loop. Consequently, on Ouro, LoopCD-Hidden frequently outperforms LoopCD-Logits. On HellaSwag, hidden-state gains exceed logit gains for both Ouro-1.4B (+1.73 versus +0.76) and Ouro-2.6B (+2.61 versus +1.56). Furthermore, when matched for reference state and strength, the seven-benchmark means consistently favor the hidden-state form (0.94 versus 0.88 for 2.6B, and 0.10 versus 0.51 for 1.4B).

\begin{figure}
\centering
\includegraphics[width=\linewidth,alt={On most benchmarks the hidden-state form matches or exceeds the logit form; Parcae-1.3B is the exception.}]{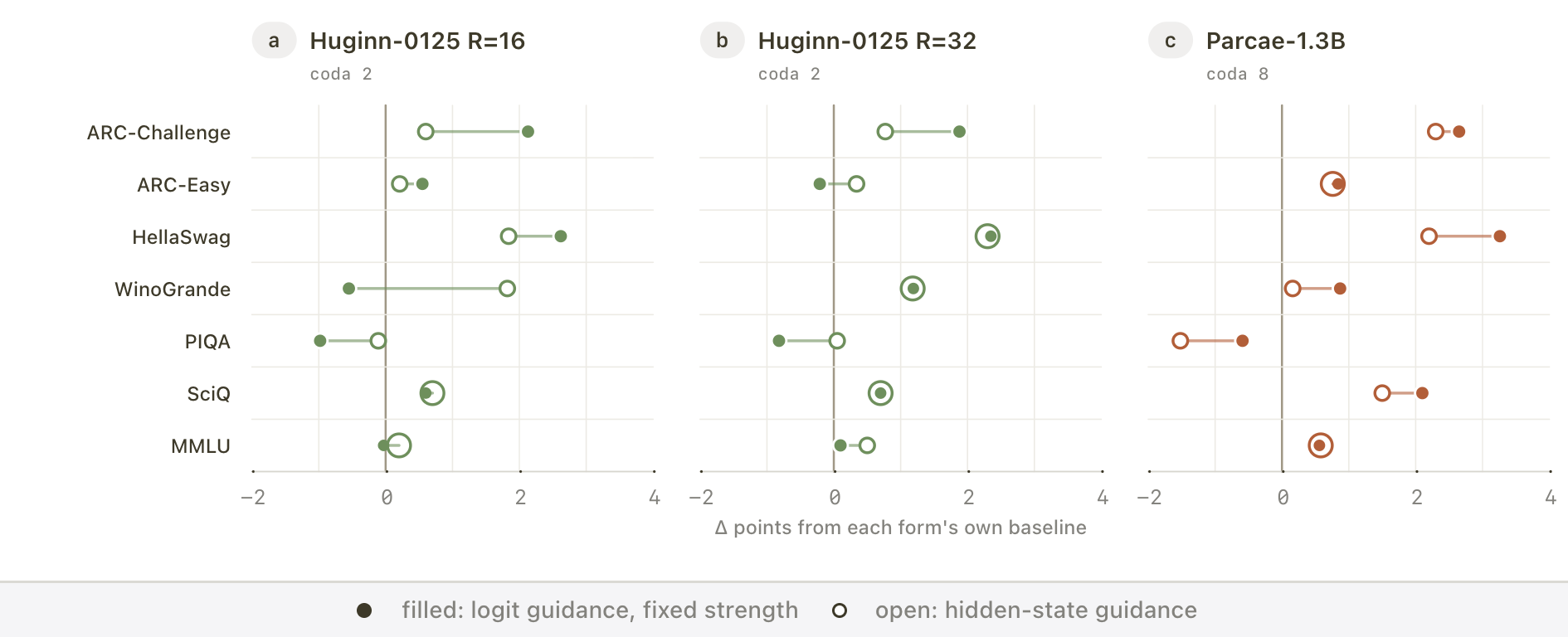}
\caption{\textbf{On most benchmarks the hidden-state form matches or exceeds the logit form; Parcae-1.3B is the exception.} Change from the unguided model per benchmark at the selected settings of Tables \hyperref[tab-gen]{2} and \hyperref[tab-hidden]{3}, LoopCD-Logits (filled) and LoopCD-Hidden (open). Rows are benchmarks; columns are the configurations with a coda, in order of coda depth.}\label{fig-scoring-benchmarks}
\end{figure}

This behavior can be explained by representation attenuation. A guidance contrast injected before the coda is progressively damped by every subsequent layer. Note that while this structural trend holds broadly, the reported hidden-state rows optimize their own references and strengths independently (Appendix~\hyperref[app-selection]{B.4}). Within multiple-choice benchmarks specifically, the logit form maintains its advantage primarily on ARC-Challenge and HellaSwag for all three architectures with a coda. On most other benchmarks, the hidden form matches or exceeds logit performance, with Parcae-1.3B being the notable exception (Figure~\hyperref[fig-scoring-benchmarks]{A3}).

\textbf{Performance retention in generative tasks.} Extending this analysis to generative benchmarks, the hidden-state form actually exceeds the fixed logit form across both evaluated Huginn depths (Figure~\hyperref[fig-space-tradeoff]{A2}(b)). Evaluated against their respective baselines in code generation tasks, LoopCD-Hidden achieves 5.13 versus 3.97 points on the four-column mean at \(R=32\), and 3.76 versus 3.04 points at \(R=16\).

\begin{figure}
\centering
\includegraphics[width=\linewidth,alt={How far guidance moves the output depends on the coda.}]{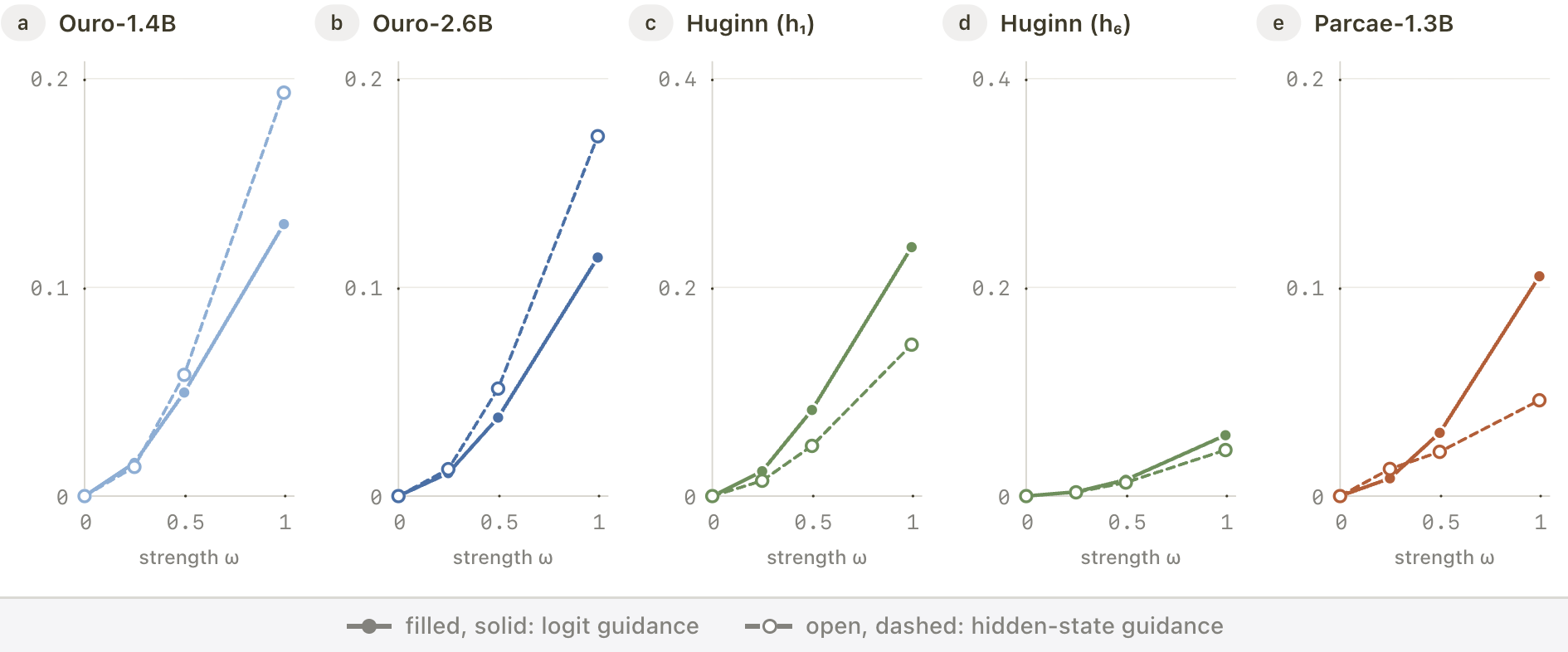}
\caption{\textbf{How far guidance moves the output depends on the coda.} Mean per-position Jensen--Shannon divergence (bits) from the unguided final prediction against the strength, 300 HellaSwag documents; LoopCD-Logits solid with filled markers, LoopCD-Hidden dashed with open markers. \textbf{(a, b)} Ouro, no coda: the hidden-state form moves the output more. \textbf{(c, d)} Huginn at \(R=32\), references \(h_{1}\) and \(h_{6}\). \textbf{(e)} Parcae-1.3B, deep coda: the hidden-state form moves it less. Panels of one family share a y-axis.}\label{fig-space-step}
\end{figure}

\phantomsection\label{app-space-step}\textbf{Output sensitivity to guidance strength.} When analyzing the distributional updates directly, identical strength values (\(\omega\)) yield different magnitudes of output divergence depending on the architecture. Specifically, applying guidance in the hidden-state space produces a larger output divergence for models whose entire stack loops, but a substantially smaller divergence for models with a deep coda (Table~\hyperref[tab-step-sizes]{A10}, Figure~\hyperref[fig-space-step]{A4}).

In Ouro, only layer normalization separates the recurrent state from the vocabulary head. Consequently, LoopCD-Hidden shifts the output distribution much more aggressively than LoopCD-Logits at equivalent strengths. Because of this heightened sensitivity, Ouro's optimal hidden-state gain peaks at lower strength values compared to its logit gain (0.25 versus 0.5 for Ouro-1.4B, and 0.5 versus 1 for Ouro-2.6B). These empirical peaks align perfectly with our selected hyperparameters (Appendix~\hyperref[app-selection]{B.4}).

Conversely, a coda damps the hidden-state update. For Parcae (eight coda layers) and Huginn, LoopCD-Logits moves the distribution further than LoopCD-Hidden at identical strengths, entirely reversing the trend seen in Ouro. Interestingly, Huginn's hidden-state form utilizing the first state (\(h_{1}\)) still peaks early (0.15 versus 0.25). This exception occurs because the initial-state update is only moderately aligned with the true re-ranking direction; instead, it carries a large noise component orthogonal to the final logits (Figure~\hyperref[fig-verdicts]{A8}(b)), as analyzed in Appendix~\hyperref[app-reference-sweep]{E.1}.

\begin{table}
\caption{\textbf{At the same strength, LoopCD-Hidden moves the output more than LoopCD-Logits in Ouro and less behind a coda.} Mean per-position Jensen--Shannon divergence (bits) from the unguided final prediction, 300 HellaSwag documents. Parcae's values sit above a 0.006-bit floor that two identical passes already show.}\label{tab-step-sizes}

\begin{tabular}{lrrrrr}
\toprule
\lxBeginTableHead
\lxTableColumnHead\textbf{} & \lxTableColumnHead\textbf{} & \multicolumn{2}{c}{\lxTableColumnHead\textbf{\(\omega=0.5\)}} & \multicolumn{2}{c}{\lxTableColumnHead\textbf{\(\omega=1\)}} \\
\lxTableColumnHead\textbf{Model} & \lxTableColumnHead\textbf{Reference} & \lxTableColumnHead\textbf{hidden states} & \lxTableColumnHead\textbf{logits} & \lxTableColumnHead\textbf{hidden states} & \lxTableColumnHead\textbf{logits} \\
\lxEndTableHead
\midrule
Ouro-1.4B & \(h_{1}\) & 0.058 & 0.050 & 0.19 & 0.13 \\
Ouro-2.6B & \(h_{1}\) & 0.052 & 0.038 & 0.17 & 0.11 \\
Huginn-0125 \(R{=}32\) & \(h_{1}\) & 0.048 & 0.083 & 0.15 & 0.24 \\
Huginn-0125 \(R{=}32\) & \(h_{6}\) & 0.013 & 0.016 & 0.044 & 0.058 \\
Parcae-1.3B & \(h_{1}\) & 0.021 & 0.030 & 0.046 & 0.105 \\
\bottomrule
\end{tabular}

\end{table}

\section{The Reference Iteration}\label{app-reference}

This appendix expands upon Section~\hyperref[sec-weak]{5.1} by deeply analyzing the dynamics of intermediate iterations. We first examine how standalone accuracy and disagreement drive reference selection, including the specific effects of noise initialization. We then map the geometric convergence of these states in representation space, and finally analyze their intra-iteration dynamics in the output space to explain convergence behavior.

\subsection{Reference selection across iterations}\label{app-reference-sweep}

To understand what makes an effective reference, Table~\hyperref[tab-reference-sweep]{A11} details the standalone accuracy, disagreement with the final prediction, and resulting guidance gain for every evaluated reference iteration \(k\). The empirical results demonstrate that guidance gain scales reliably with the reference's disagreement from the final prediction, rather than its standalone competence.

For instance, Parcae-1.3B's sixth step achieves a highly competent standalone accuracy of 41.04\% compared to the final prediction's 40.36\%. However, because it has largely converged with the final state, it yields only a single point of gain. Conversely, Huginn's second step is just as weak as its first (22.10\% against 22.78\%), yet because its disagreement has already shrunk, it provides only half the guidance gain of the initial step. Consequently, the first iteration universally serves as the optimal logit reference across stable architectures.

This reliance on divergence direction also explains a notable architectural anomaly: Huginn's hidden-state reference performance. As shown in Figure~\hyperref[fig-reference]{7}(b), Huginn's hidden-state sweep is negative at the first state and peaks later at the sixth. This occurs because Huginn initializes its recurrent state from Gaussian noise and subsequently moves it on a hypersphere. Across its 32 steps, the state's norm stays tightly bounded at 76.4 within a spread of 0.05. The first step is nearly as long as the radius itself, making the whole path 6.0 times longer than the net displacement---behaving similarly to a random walk for every one of the 1,172 ARC-Challenge questions.

Therefore, the contrast between the first hidden state and the final one is heavily dominated by the initial noise-removal step, not by a meaningful change in the prediction itself. Because the coda layers and head project this initial noise away, the first logits form an excellent reference. However, LoopCD-Hidden directly amplifies this unprojected noise, rendering the first hidden state detrimental.

\begin{table}
\caption{\textbf{Guidance gain depends on disagreement rather than standalone accuracy.} LoopCD-Logits at \(\omega=0.5\) with reference iteration \(k\). Each block's header gives the unguided model's accuracy; ``alone'' is the reference's own accuracy. Shot counts as in Table \hyperref[tab-fewshot]{A2}.}\label{tab-reference-sweep}

\textbf{ARC-Challenge}\par

\begin{tabular}{lrrr}
\toprule
\lxBeginTableHead
\lxTableColumnHead\textbf{Reference \(k\)} & \lxTableColumnHead\textbf{Alone\newline{}(\%)} & \lxTableColumnHead\textbf{Disagrees\newline{}(\%)} & \lxTableColumnHead\textbf{Gain at\newline{}\(\omega=0.5\)} \\
\lxEndTableHead
\midrule
\multicolumn{4}{c}{\emph{Huginn-0125}, final prediction 37.54\%} \\
1 & 22.78 & 51.3 & 3.07 \\
2 & 22.10 & 46.8 & 1.79 \\
4 & 30.12 & 35.8 & 0.94 \\
6 & 32.25 & 29.5 & 1.11 \\
8 & 33.45 & 20.0 & 0.60 \\
16 & 37.20 & 7.3 & 0.17 \\
24 & 38.05 & 2.7 & 0.17 \\
\multicolumn{4}{c}{\emph{Parcae-1.3B}, final prediction 40.36\%} \\
1 & 32.34 & 30.0 & 3.50 \\
2 & 38.91 & 18.1 & 2.39 \\
4 & 40.36 & 9.7 & 1.54 \\
6 & 41.04 & 7.0 & 1.02 \\
\multicolumn{4}{c}{\emph{Ouro-1.4B}, final prediction 60.41\%} \\
1 & 38.65 & 40.2 & 2.13 \\
2 & 54.86 & 17.0 & 0.09 \\
3 & 59.56 & 6.1 & 0.34 \\
\multicolumn{4}{c}{\emph{Ouro-2.6B}, final prediction 66.21\%} \\
1 & 47.95 & 30.6 & 2.56 \\
2 & 62.29 & 10.7 & 0.51 \\
3 & 65.36 & 3.4 & 0.60 \\
\bottomrule
\end{tabular}

\textbf{HellaSwag}\par

\begin{tabular}{lrrr}
\toprule
\lxBeginTableHead
\lxTableColumnHead\textbf{Reference \(k\)} & \lxTableColumnHead\textbf{Alone\newline{}(\%)} & \lxTableColumnHead\textbf{Disagrees\newline{}(\%)} & \lxTableColumnHead\textbf{Gain at\newline{}\(\omega=0.5\)} \\
\lxEndTableHead
\midrule
\multicolumn{4}{c}{\emph{Huginn-0125}, final prediction 66.74\%} \\
2 & 34.40 & 44.2 & 4.31 \\
4 & 46.35 & 27.6 & 4.12 \\
6 & 56.36 & 15.3 & 2.58 \\
8 & 61.30 & 9.4 & 1.47 \\
16 & 66.33 & 2.3 & 0.09 \\
24 & 66.62 & 0.8 & 0.05 \\
\multicolumn{4}{c}{\emph{Parcae-1.3B}, final prediction 56.28\%} \\
1 & 44.39 & 19.9 & 2.90 \\
2 & 50.76 & 11.4 & 1.68 \\
4 & 55.01 & 5.7 & 0.47 \\
6 & 56.09 & 3.6 & 0.13 \\
\multicolumn{4}{c}{\emph{Ouro-1.4B}, final prediction 74.52\%} \\
1 & 55.39 & 31.1 & 0.76 \\
2 & 71.47 & 10.8 & 0.33 \\
3 & 73.95 & 3.1 & 0.03 \\
\multicolumn{4}{c}{\emph{Ouro-2.6B}, final prediction 79.47\%} \\
1 & 68.89 & 18.0 & 1.56 \\
2 & 77.84 & 6.6 & 0.86 \\
3 & 79.23 & 2.6 & 0.03 \\
\bottomrule
\end{tabular}

\end{table}

\subsection{Geometric convergence in representation space}\label{app-representation-analysis}

\begin{figure}
\centering
\includegraphics[width=\linewidth,alt={Each Ouro recurrent iteration reworks the state; Looped-Qwen3's damped substeps move it gradually.}]{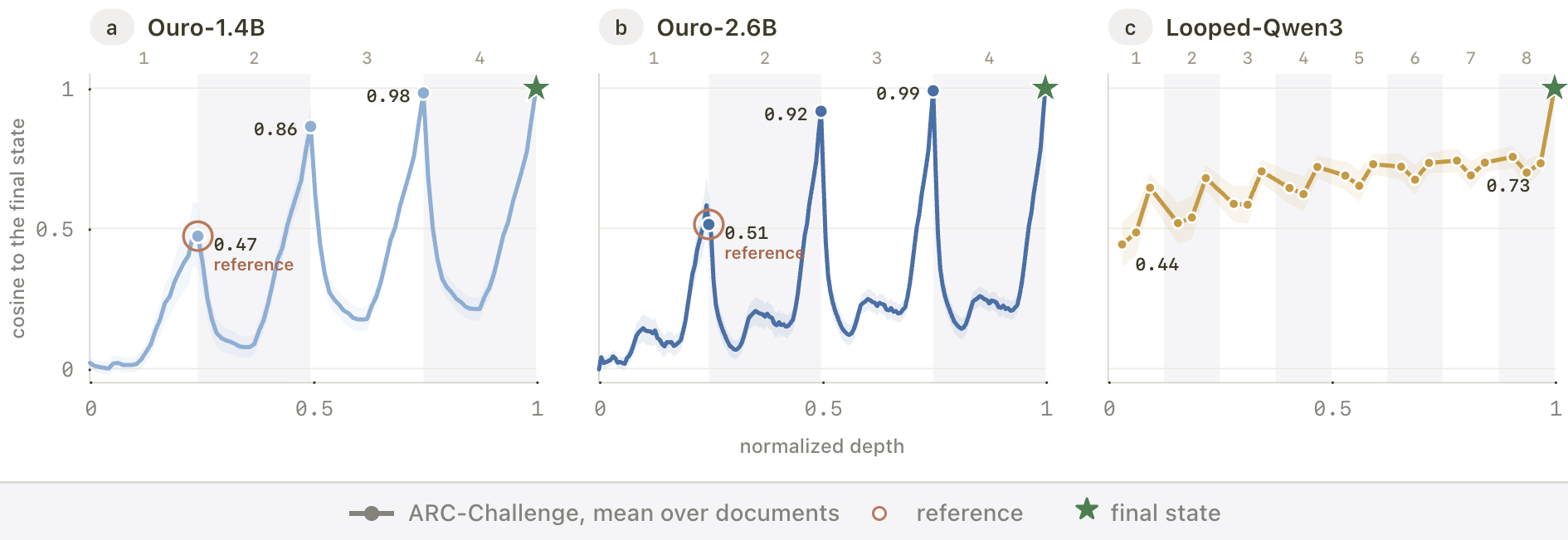}
\caption{\textbf{Each Ouro recurrent iteration reworks the state; Looped-Qwen3's damped substeps move it gradually.} Standardized cosine similarity to the final state across the unrolled layers, ARC-Challenge, with a band for the spread over documents. Alternate shading marks Ouro's recurrent iterations and Looped-Qwen3's substeps; dots mark the ends of Ouro's recurrent iterations, the ring the reference \(h_{1}\) of Tables \hyperref[tab-reason]{1} and \hyperref[tab-gen]{2}, the star the final state.}\label{fig-geometry}
\end{figure}

\begin{figure}
\centering
\includegraphics[width=\linewidth,alt={Across ten configurations, Ouro's similarity drops at each recurrent-iteration boundary and Looped-Qwen3's climbs more smoothly.}]{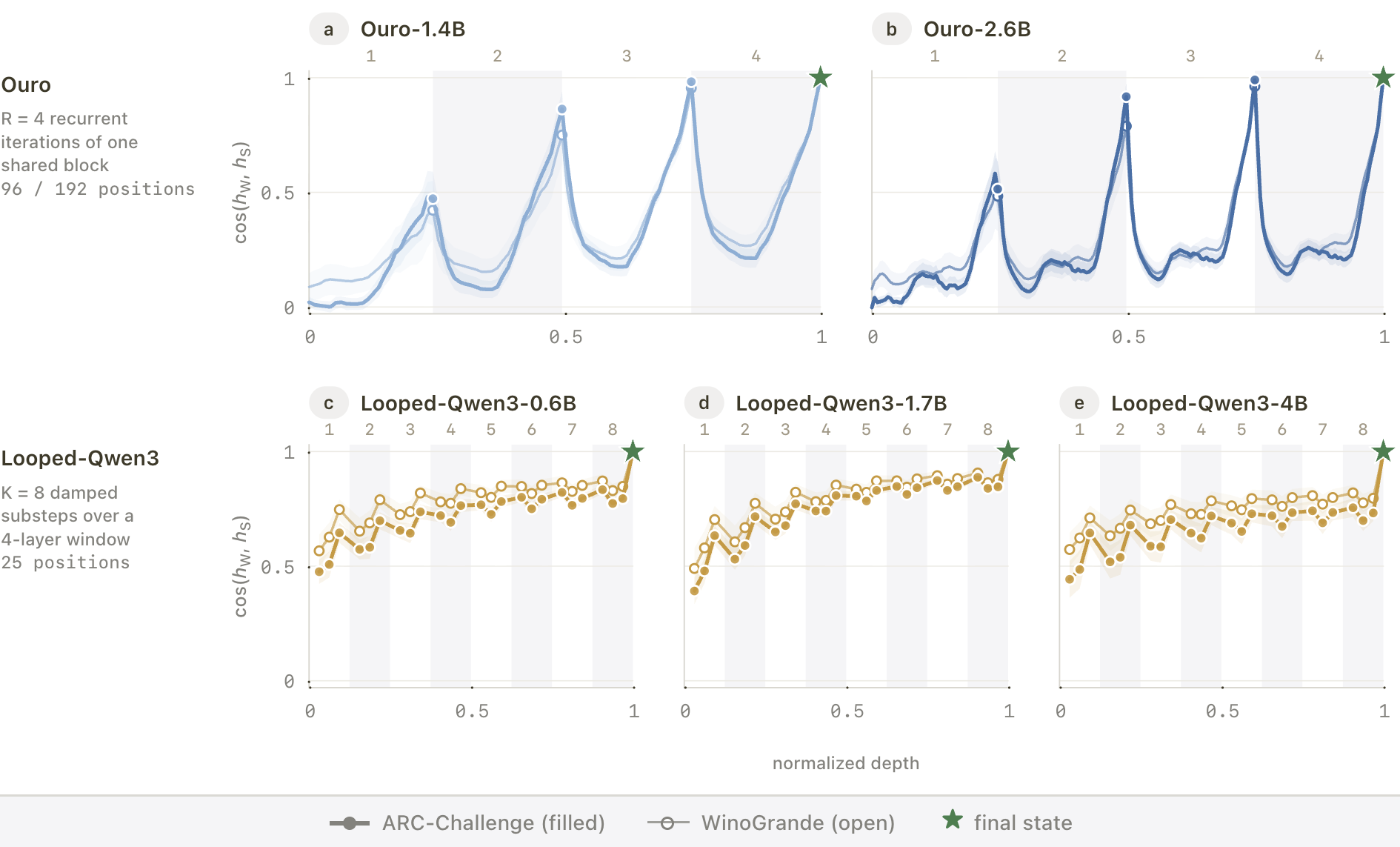}
\caption{\textbf{Across ten configurations, Ouro's similarity drops at each recurrent-iteration boundary and Looped-Qwen3's climbs more smoothly.} Standardized cosine similarity of each intermediate state to the final state against normalized depth. Bands give the spread over documents; the star marks the final state.}\label{fig-cosmatrix}
\end{figure}

Transitioning from output accuracy to internal representations, Figures~\hyperref[fig-geometry]{A5} and~\hyperref[fig-cosmatrix]{A6} visualize how intermediate states converge geometrically toward the final state. To isolate meaningful structural changes, every hidden dimension is standardized using its mean and variance pooled over the scored positions and unrolled layers. Let \(\tilde{h}_W\) and \(\tilde{h}_S\) denote the standardized intermediate state and the standardized final state, respectively. The geometric convergence is measured using cosine similarity:

\begin{equation}
\text{sim}(\tilde{h}_W, \tilde{h}_S) = \frac{\tilde{h}_W \cdot \tilde{h}_S}{\|\tilde{h}_W\|_2 \|\tilde{h}_S\|_2}
\end{equation}

Computing the similarity strictly after this standardization drastically reduces the distortion caused by dimensions with excessively large variances.

These visualizations highlight distinct architectural dynamics. As shown in Figure~\hyperref[fig-geometry]{A5}, Ouro exhibits a sharp geometric reset. Its similarity to the final state drops dramatically after every pass boundary, followed by a steady recovery within the pass itself. In contrast, Looped-Qwen3 incorporates damped substeps that constrain the updates, resulting in a much smoother, gradual climb toward the final representation.

\subsection{Intra-iteration dynamics and output convergence}\label{app-layer-axis}

While the geometric view maps hidden-state trajectories, analyzing the output space explains why internal layers fail as references and how models ultimately settle.

\begin{figure}
\centering
\includegraphics[width=\linewidth,alt={Predictions inside a recurrent iteration stay far from the final one, and Huginn's entropy settles by its sixteenth iteration.}]{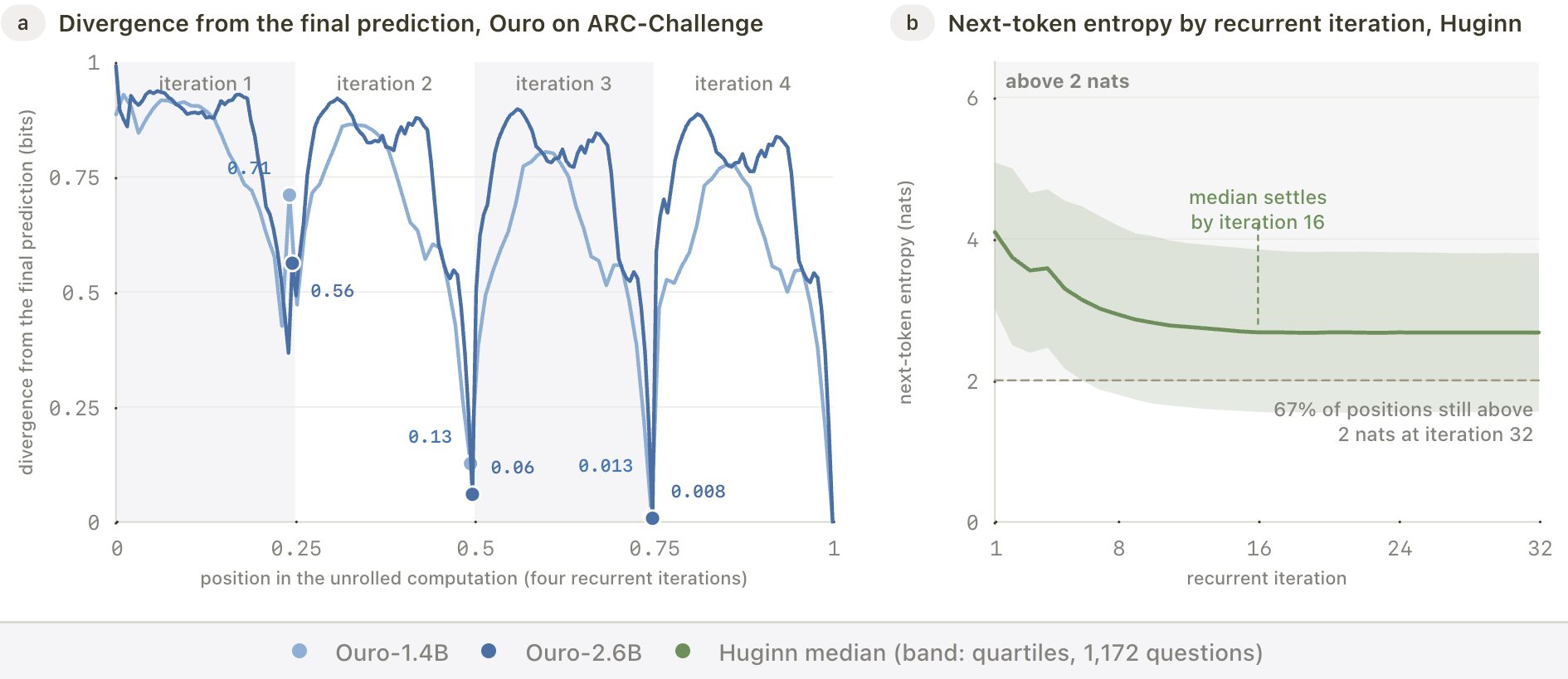}
\caption{\textbf{Predictions inside a recurrent iteration stay far from the final one, and Huginn's entropy settles by its sixteenth iteration.} \textbf{(a)} Mean Jensen--Shannon divergence (bits) between the prediction at each unrolled layer and the final prediction, first answer position, ARC-Challenge; shading marks Ouro's four recurrent iterations, dots their ends. \textbf{(b)} Next-token entropy at answer positions against the recurrent iteration, Huginn at \(R=32\): median and interquartile band, with a dashed line at 2 nats.}\label{fig-contrast}
\end{figure}

Figure~\hyperref[fig-contrast]{A7}(a) measures the output divergence between each unrolled intermediate prediction and the final prediction. This difference is quantified by the Jensen--Shannon divergence (JSD), a smoothed and symmetric version of the Kullback--Leibler (KL) divergence. For an intermediate output distribution \(P\) and the final distribution \(Q\), the JSD is defined as:

\begin{equation}
\text{JSD}(P \parallel Q) = \frac{1}{2} D_{\text{KL}}(P \parallel M) + \frac{1}{2} D_{\text{KL}}(Q \parallel M)
\end{equation}

where \(M = \frac{1}{2}(P + Q)\) is the mixture distribution.

Crucially, inside every pass, the prediction distribution \(P\) drifts wildly. For Ouro-1.4B, the JSD at the end of the first three passes sits tightly at 0.71, 0.13, and 0.013 bits (0.56, 0.06, and 0.008 for Ouro-2.6B). However, across the internal layers within those passes, the divergence spikes to an average of 0.57 to 0.84 bits, remaining just as erratic in the final pass as in the first.

This internal drift mathematically invalidates intra-iteration layers as guidance references. Our sweep of internal layers (layers 0, 8, 16, and 22/46 of the final pass at \(\omega \in \{0.25, 0.5, 1.0\}\)) revealed that extracting any internal state severely degrades ARC-Challenge performance, dropping by up to 20.6 points for Ouro-1.4B and 24.9 points for Ouro-2.6B below ordinary decoding. Conversely, the completed-iteration reference \(h_{1}\) at identical strengths raises accuracy by 2.1 and 2.6 points. Unlike standard dense networks---where methods like DoLa (\hyperlink{ref-chuang2023dola}{Chuang et al., 2024}) successfully contrast intermediate layers against the final layer---looped architectures do not form trained exit points inside a pass, making intra-iteration states unstable for contrastive guidance.

Finally, observing this output convergence also explains the robustness of reduced-depth settings. Figure~\hyperref[fig-contrast]{A7}(b) traces Huginn's recurrence explicitly at the token level. The median next-token entropy over answer positions drops rapidly from 4.1 nats after the first step and effectively settles at 2.7 nats by step 16. It remains flat through step 32. Because the recurrence physically settles by the middle of its loop, halving Huginn's iterations avoids accuracy penalties (Appendix~\hyperref[app-reduced-details]{D.2}). Importantly, even upon settling, 67\% of the token positions maintain an entropy above 2 nats (roughly an even choice among seven tokens). The recurrence converges on a distribution that still preserves meaningful ambiguity, which contrastive guidance resolves.

\section{Re-ranking Close Decisions}\label{app-rerank}

This appendix expands upon Section~\hyperref[sec-pair-analysis]{5.2}. We first detail how guidance gains heavily concentrate on the model's most uncertain decisions, including a step-by-step case study. We then geometrically decompose the logit update, demonstrating that LoopCD-Hidden fundamentally acts as an orthogonal re-ranking mechanism, while the remaining parallel component dictates temperature sharpening.

\subsection{Concentration of guidance gains on uncertain decisions}\label{app-close-decisions}

Guidance effectively functions as a tie-breaker rather than a blunt overriding mechanism. It alters a final decision only when the injected contrast exceeds the probability margin between the top two candidate options. Consequently, confident predictions remain unperturbed, while undecided predictions shift.

Table~\hyperref[tab-close-decisions]{A12} stratifies the evaluation questions into five equal-sized groups (fifths) based on the margin between the final prediction's top two options. Across every evaluated model, the least confident fifth commands the vast majority of the guidance gain, whereas the most confident fifth sees virtually zero impact. HellaSwag exhibits an even sharper concentration than ARC-Challenge. For instance, the least confident fifth yields +3.7 points for Ouro-1.4B and +7.3 for Ouro-2.6B, with no other fifth contributing more than +0.5. Ultimately, guidance flips very few answers overall (between 3.9\% and 14.9\%). The reported net gains represent a targeted excess of fixed answers over broken ones, concentrated on close decisions.

\begin{table}
\caption{\textbf{The gain comes from the least confident questions and fades toward the most confident.} LoopCD-Logits at \(\omega=0.5\), reference \(h_{1}\), questions in five equal groups (fifths) by the unguided prediction's confidence, least confident first; the last two columns give the share of answers changed and the net change in accuracy. ARC-Challenge and HellaSwag, shot counts as in Table \hyperref[tab-fewshot]{A2}.}\label{tab-close-decisions}

\begin{tabular}{lrrrrrrrr}
\toprule
\lxBeginTableHead
\lxTableColumnHead\textbf{} & \lxTableColumnHead\textbf{} & \multicolumn{5}{c}{\lxTableColumnHead\textbf{Gain on each fifth, closest to farthest (points)}} & \lxTableColumnHead\textbf{Answers\newline{}changed} & \lxTableColumnHead\textbf{Net\newline{}gain} \\
\lxTableColumnHead\textbf{Model} & \lxTableColumnHead\textbf{Benchmark} & \lxTableColumnHead\textbf{1} & \lxTableColumnHead\textbf{2} & \lxTableColumnHead\textbf{3} & \lxTableColumnHead\textbf{4} & \lxTableColumnHead\textbf{5} & \lxTableColumnHead\textbf{} & \lxTableColumnHead\textbf{} \\
\lxEndTableHead
\midrule
Ouro-1.4B & ARC-Challenge & +6.41 & +3.85 & +0.43 & 0.00 & 0.00 & 6.6\% & +2.1 \\
Ouro-2.6B & ARC-Challenge & +7.26 & +3.85 & +1.28 & +0.43 & 0.00 & 5.8\% & +2.6 \\
Huginn-0125 & ARC-Challenge & +7.26 & +4.70 & +2.13 & +0.85 & +0.43 & 14.9\% & +3.1 \\
Parcae-1.3B & ARC-Challenge & +13.25 & +5.56 & -0.43 & -0.85 & 0.00 & 8.8\% & +3.5 \\
Ouro-1.4B & HellaSwag & +3.69 & +0.10 & 0.00 & 0.00 & 0.00 & 3.9\% & +0.76 \\
Ouro-2.6B & HellaSwag & +7.27 & +0.45 & +0.10 & 0.00 & 0.00 & 4.2\% & +1.56 \\
\bottomrule
\end{tabular}

\end{table}

To physically illustrate how these borderline decisions flip, Table~\hyperref[tab-worked-example]{A13} traces a single ARC-Challenge question through Ouro-1.4B's recurrent passes. The question asks which natural disaster leaves a narrow path of destruction through a forest. Across the early passes, the score for "tornado" rises significantly faster than for "earthquake" (+0.90 versus +0.32). However, at the final pass, "earthquake" still narrowly leads by a margin of less than 0.01. By applying guidance, LoopCD extrapolates this existing momentum, lifting "tornado" into first place. The guidance mechanism introduces no external knowledge; it simply reads the directional momentum from the reference state and continues shifting the options along that axis.

\begin{table}
\caption{\textbf{Guidance flips a close decision on one ARC-Challenge question.} Ouro-1.4B, 25-shot: per-character log-likelihood of each option after the first pass, at the final pass, and under LoopCD-Logits (\(\omega=0.5\), reference \(h_{1}\)); bold marks the leading option.}\label{tab-worked-example}

\begin{tabular}{lrrr}
\toprule
\lxBeginTableHead
\lxTableColumnHead\textbf{Option} & \lxTableColumnHead\textbf{After the first pass, alone} & \lxTableColumnHead\textbf{Final prediction} & \lxTableColumnHead\textbf{Guided, \(\omega=0.5\)} \\
\lxEndTableHead
\midrule
a flood & -1.500 & -0.770 & -0.934 \\
a tornado (correct) & -1.113 & -0.209 & \textbf{-0.170} \\
a hurricane & -0.969 & -0.258 & -0.247 \\
an earthquake & -0.525 & \textbf{-0.201} & -0.199 \\
\bottomrule
\end{tabular}

\end{table}

\subsection{Geometric equivalence of LoopCD-Hidden and orthogonal re-ranking}\label{app-rerank-identity}

As outlined in Section~\hyperref[sec-pair-analysis]{5.2}, a logit-space contrast vector can be mathematically decomposed into two orthogonal components: a part parallel to the final logits (which rescales the distribution like a temperature change) and a part orthogonal to the final logits (which re-ranks the relative standing of tokens).

Crucially, in architectures where the entire stack loops, LoopCD-Hidden behaves almost identically to the orthogonal re-ranking component of LoopCD-Logits. Because any hidden-state update must ultimately pass through the final layer normalization before reaching the vocabulary head, its magnitude (length) is discarded. The head receives only a directional shift. Pure directional shifts inherently act as re-ranking mechanisms.

In Ouro, where only normalization separates the recurrent loop from the head, empirical measurements confirm this equivalence. We evaluate the final answers (verdicts) produced by three distinct guidance applications: LoopCD-Hidden, the full logit update (applying both orthogonal and parallel components), and the isolated logit re-ranking update (applying strictly the orthogonal component).

\begin{figure}
\centering
\includegraphics[width=\linewidth,alt={LoopCD-Hidden follows the logit re-ranking, in its answers and in its direction.}]{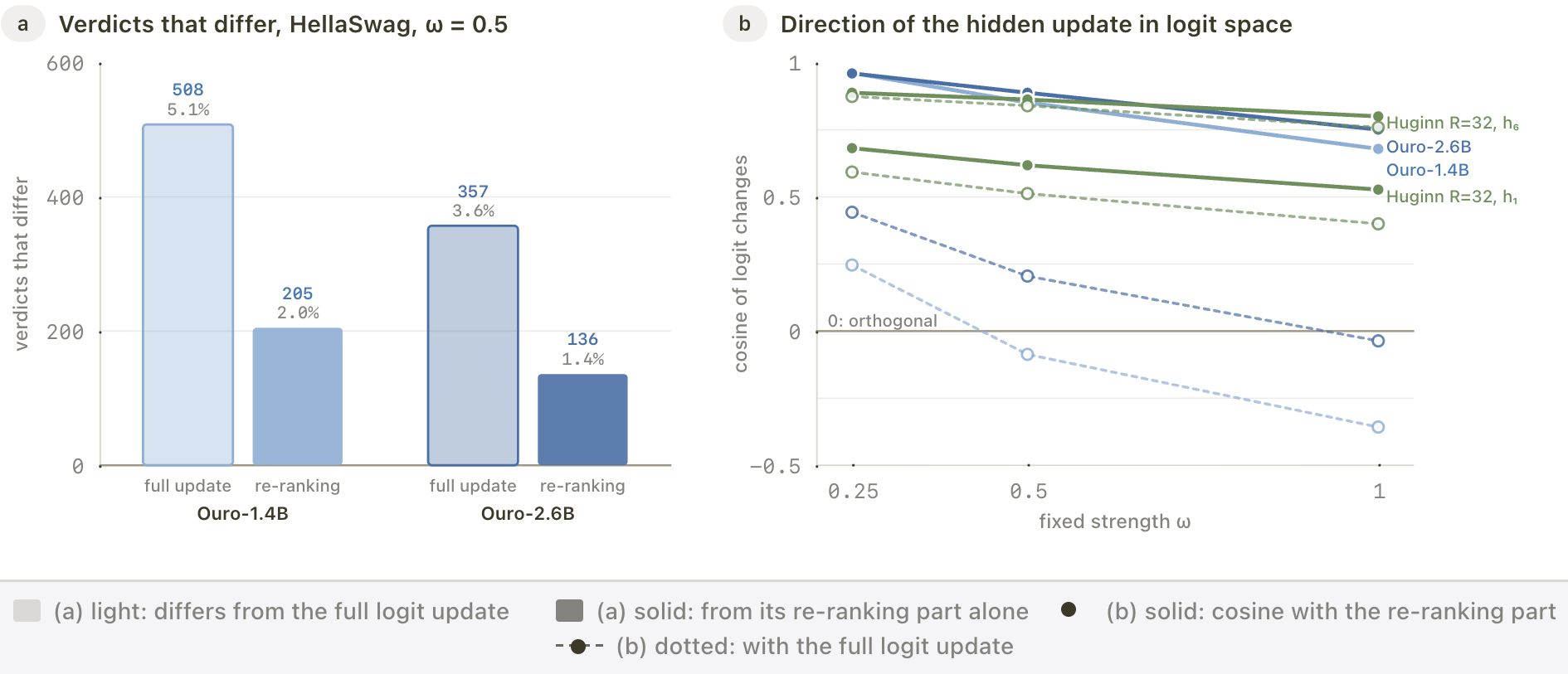}
\caption{\textbf{LoopCD-Hidden follows the logit re-ranking, in its answers and in its direction.} HellaSwag, \(\omega=0.5\), reference \(h_{1}\). \textbf{(a)} Questions, of 10,042, on which LoopCD-Hidden's answer differs from the full logit update (light) and from its re-ranking part alone (solid). \textbf{(b)} Cosine similarity in mean-centred logit space between the LoopCD-Hidden update and the re-ranking part (solid) or the full update (dashed), mean over 300 documents.}\label{fig-verdicts}
\end{figure}

On HellaSwag at \(\omega=0.5\), LoopCD-Hidden diverges from the pure logit re-ranking component on merely 205 of the 10,042 evaluation verdicts for Ouro-1.4B (136 for Ouro-2.6B). In stark contrast, it diverges from the full logit update on a much larger 508 verdicts (357 for Ouro-2.6B). Figure~\hyperref[fig-verdicts]{A8}(a) illustrates how effectively the hidden-state application replicates the isolated orthogonal behavior.

This equivalence is equally visible when measuring the update vectors directly in logit space (Figure~\hyperref[fig-verdicts]{A8}(b)). For both Ouro models at \(\omega=0.25\), the hidden-state update maintains a cosine similarity of 0.96 with the isolated re-ranking component's update, vastly exceeding its 0.25 similarity to the full logit update.

For Huginn, where a coda separates the loop from the head, the dynamics shift slightly. When using its optimal sixth state as a reference, the hidden-state update still closely tracks the re-ranking component (cosine similarity 0.89 at \(\omega=0.25\)). However, when forced to use the first state, alignment drops to 0.62 at \(\omega=0.5\). This drop occurs because Huginn's first hidden state is dominated by initialization noise (Appendix~\hyperref[app-reference-sweep]{E.1}), injecting random directional shifts rather than a coherent re-ranking signal.

\subsection{The double-edged effect of temperature sharpening}\label{sec-the-double-edged-effect-of-temperature-sharpening}

Because LoopCD-Hidden effectively mirrors orthogonal re-ranking, the functional difference between the hidden and logit forms is precisely the parallel component that the logit form retains: temperature sharpening.

The utility of this sharpening component is strictly benchmark-dependent. On HellaSwag, aggressive sharpening degrades performance. The full logit update at \(\omega=0.5\) yields +0.76 points for Ouro-1.4B, whereas isolating its re-ranking component yields +1.72. Because the hidden-state form naturally lacks this detrimental sharpening, it achieves double the gain (+1.73 points). Conversely, on ARC-Challenge, sharpening provides a distinct advantage. The pure re-ranking component requires a much larger strength (\(\omega=1.3\)) to reach its peak gain of +2.65, whereas the full logit update peaks efficiently at \(\omega=0.9\) with +2.47.

Consequently, for fully recurrent models lacking a coda, selecting between LoopCD-Logits and LoopCD-Hidden is effectively a dataset-dependent choice regarding whether temperature sharpening is beneficial, typically swaying net accuracy by one or two percentage points. Across our broader seven-benchmark suite, this dynamic holds consistently: LoopCD-Hidden excels on HellaSwag but trails LoopCD-Logits on ARC-Challenge, ARC-Easy, SciQ, and MMLU.

\section{Guidance Strength Dynamics}\label{app-strength}

This appendix expands upon Section~\hyperref[sec-strength-analysis]{5.3}, providing benchmark-level resolution for generation tasks and detailing the spatial distribution of updates across confidence fifths to explain the overshoot phenomenon at high guidance strengths.

\subsection{Strength boundaries: generation collapse and negative interpolation}\label{app-strength-sweeps}

\begin{figure}
\centering
\includegraphics[width=\linewidth,alt={Every generation benchmark gains only over a narrow range of strength.}]{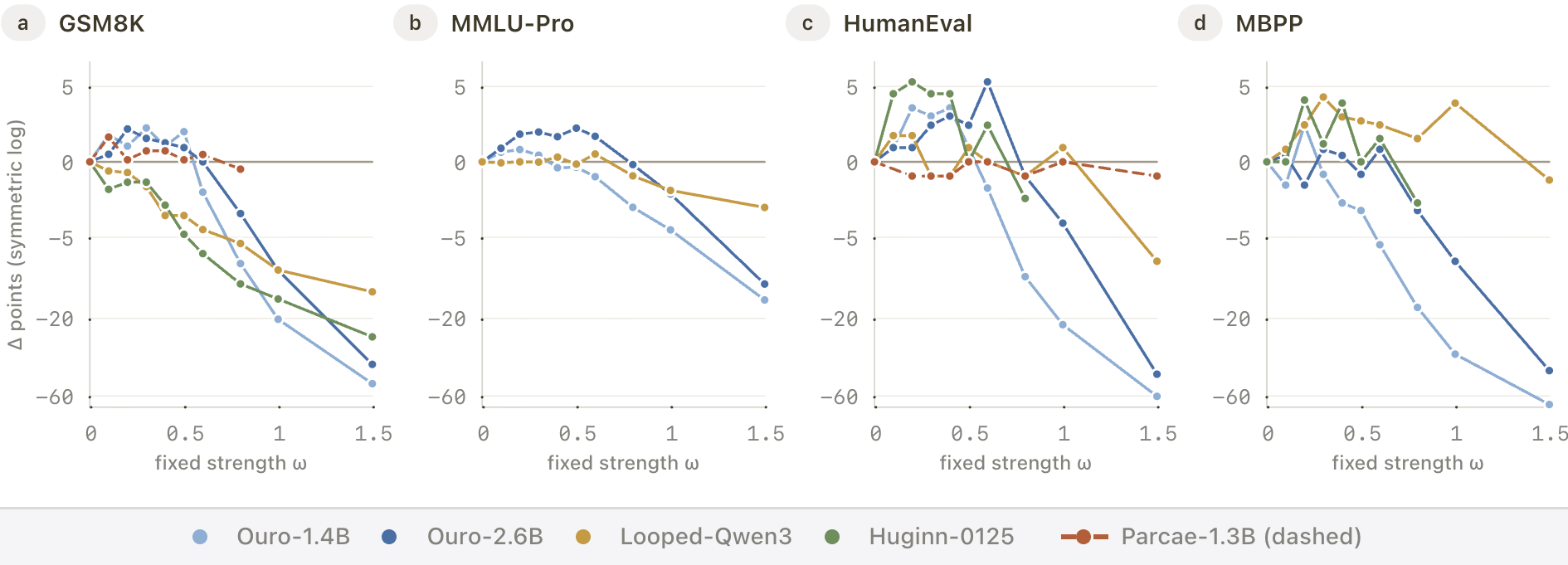}
\caption{\textbf{Every generation benchmark gains only over a narrow range of strength.} Change from the unguided model against the fixed logit strength, five models, on GSM8K, MMLU-Pro, HumanEval, and MBPP; a symmetric-log axis keeps the fall beyond \(\omega\approx0.6\) in frame. Huginn's GSM8K runs use chain of thought.}\label{fig-gen-sweeps}
\end{figure}

Figure~\hyperref[fig-gen-sweeps]{A9} disaggregates the generation sweeps from Figure~\hyperref[fig-strength-sweep]{9}(b) across individual benchmarks to illustrate the upper bounds of beneficial guidance. The operating window for autoregressive generation is uniformly narrow, and the degradation beyond this optimal positive range is catastrophic. For instance, on GSM8K, Ouro-1.4B loses 8 points at \(\omega=0.8\) and plummets by 20 points at \(\omega=1.0\). HumanEval exhibits similarly steep declines of 10 and 22 points at the same respective strengths.

While generation tasks expose the strict upper bounds of positive guidance, evaluating multiple-choice scoring at negative strengths exposes the vector's lower bounds. Negative strengths explicitly interpolate backward along the contrast vector toward the weaker reference state. As mathematically expected, applying a negative \(\omega\) uniformly degrades the seven-benchmark multiple-choice mean across all models at every value, predictably converging entirely to the standalone reference accuracy at \(\omega=-1.0\) (Figure~\hyperref[fig-weak]{5}(a)). This symmetric behavior validates that the isolated contrast vector strictly encodes the direction of model improvement.

\subsection{The trade-off between decision reach and overshoot}\label{app-quintile-reach}

As established in Section~\hyperref[sec-strength-analysis]{5.3}, applying a large fixed strength overshoots and corrupts previously settled decisions. Table~\hyperref[tab-quintile-reach-app]{A14} quantifies this phenomenon on ARC-Challenge, illustrating the inherent trade-off between extending reach into undecided questions and overshooting established ones as the fixed \(\omega\) increases from \(0.25\) to \(1.0\). Questions are stratified into fifths based on the probability margin between the final prediction's two leading candidates (from least to most confident). Table~\hyperref[tab-quintile-reach-app]{A14} summarizes the answer-change proportions and the accuracy delta within each fifth across model families.

\begin{table}
\caption{\textbf{A larger strength changes more answers, yet gains less on the least confident questions.} LoopCD-Logits on ARC-Challenge at three fixed strengths \(\omega\): the share of answers changed, and the accuracy change (points) in the least and the most confident of five equal groups (fifths) by the unguided prediction's confidence. A dash marks a strength not run for that model.}\label{tab-quintile-reach-app}

\begin{tabular}{lrrrrrr}
\toprule
\lxBeginTableHead
\lxTableColumnHead\textbf{} & \multicolumn{3}{c}{\lxTableColumnHead\textbf{Answers changed}} & \multicolumn{2}{c}{\lxTableColumnHead\textbf{Least confident fifth}} & \lxTableColumnHead\textbf{Most confident fifth} \\
\lxTableColumnHead\textbf{Model} & \lxTableColumnHead\textbf{\(\omega=0.25\)} & \lxTableColumnHead\textbf{\(0.5\)} & \lxTableColumnHead\textbf{\(1\)} & \lxTableColumnHead\textbf{\(\omega=0.5\)} & \lxTableColumnHead\textbf{\(1\)} & \lxTableColumnHead\textbf{\(\omega=0.5\)} \\
\lxEndTableHead
\midrule
Ouro-1.4B & 3.8\% & 6.6\% & 9.0\% & +6.4 & +3.9 & 0.0 \\
Huginn-0125 & 9.1\% & 14.9\% & 20.7\% & +7.3 & +5.1 & +0.4 \\
Parcae-1.3B & -- & 8.8\% & -- & +13.3 & -- & 0.0 \\
\bottomrule
\end{tabular}

\end{table}

As \(\omega\) grows, guidance forcefully modifies a progressively larger fraction of model verdicts. For Ouro-1.4B, the overall share of changed answers rises monotonically from 3.8\% at \(\omega=0.25\) to 6.6\% at \(\omega=0.5\) and 9.0\% at \(\omega=1.0\). Huginn-0125 exhibits a parallel progression across a wider dynamic range, changing 9.1\% of answers at \(\omega=0.25\), 14.9\% at \(\omega=0.5\), and 20.7\% at \(\omega=1.0\). Parcae-1.3B similarly modifies 8.8\% of verdicts at \(\omega=0.5\). Across all architectures, larger coefficients extend the update's reach into decisions that sat much farther from the decision boundary in the unguided baseline.

However, analyzing the spatial distribution of these accuracy gains reveals why net performance consistently peaks at moderate strengths. At \(\omega = 0.25\), accuracy gains are confined almost entirely to the least confident fifth, with negligible impact on remaining questions. At \(\omega = 0.5\), guidance optimally improves this least confident fifth (+6.4 points for Ouro-1.4B, +7.3 points for Huginn) while gradually spreading gains into the second and third fifths.

Crucially, at \(\omega = 1.0\), the accuracy within the least confident fifth actively recedes: Ouro-1.4B drops to +3.9 points, and Huginn drops to +5.1 points. Although \(\omega = 1.0\) successfully forces updates in more confident fifths, it severely overshoots the optimal decision boundary for the borderline alternatives in the least confident fifth, aggressively flipping previously corrected predictions back into errors. The aggregate score universally peaks between \(\omega = 0.25\) and \(\omega = 1.0\) because the marginal benefit of reaching more confident fifths is eventually eclipsed by the mounting overshoot penalties on close decisions. This strict spatial trade-off across question difficulty mathematically necessitates the adaptive gating mechanism of Eq.~\hyperref[eq-tokenwise]{(4)}, which dynamically applies maximum guidance where margins are narrow and withholds it where predictions are settled.

\section{Related Work}\label{app-related}

\textbf{Looped Transformer and recurrent computing.} Reusing Transformer layers was originally proposed to increase effective network depth without a proportional increase in parameter count (\hyperlink{ref-dehghani2019universal}{Dehghani et al., 2019}). Recent paradigms have shifted the primary focus of recurrence toward continuous latent reasoning and test-time computation scaling~(\hyperlink{ref-hao2025training}{Hao et al., 2025}; \hyperlink{ref-chen2025reasoning}{Chen et al., 2025}; \hyperlink{ref-ICLR2026_843e1d8a}{Kang et al., 2026}; \hyperlink{ref-ICLR2026_a4b1a25e}{Du et al., 2026}). By iteratively applying shared parameters, models can implicitly unroll reasoning steps in the latent space to solve complex logical tasks and generalize to longer sequences (\hyperlink{ref-giannou2023programmable}{Giannou et al., 2023}; \hyperlink{ref-yang2024learningalgorithms}{Yang et al., 2024}; \hyperlink{ref-fan2025length}{Fan et al., 2025}; \hyperlink{ref-saunshi2025latentthoughts}{Saunshi et al., 2025}; \hyperlink{ref-zhu2025virtualdepth}{Zhu et al., 2025b}; \hyperlink{ref-liu2026latent}{Liu et al., 2026}). Pre-trained architectures like Ouro (\hyperlink{ref-ouro2025}{Zhu et al., 2025a}), Huginn (\hyperlink{ref-geiping2025huginn}{Geiping et al., 2025}), and Parcae (\hyperlink{ref-parcae2026}{Prairie et al., 2026}) explicitly embed this recurrent trajectory into the language modeling objective. Extracting maximal reasoning performance from these trajectories often involves intervening during training, such as supervising latent steps with reasoning traces (\hyperlink{ref-fan2026lotus}{Fan et al., 2026}) and distributing reinforcement learning credit along the loop (\hyperlink{ref-williams2026rltt}{Williams et al., 2026}), or optimizing loop counts for parallel refinement during inference (\hyperlink{ref-yang2026loopcoderv2}{Yang et al., 2026}). LoopCD diverges from these structural modifications by capitalizing on the natural semantic divergence between early and final iterations of pre-trained loops, leveraging the trajectory to drive weak-to-strong extrapolation strictly at inference time.

\textbf{Weak-to-strong contrastive decoding.} Contrastive decoding (CD) improves generation by extrapolating an expert model's prediction away from a weaker reference, isolating and amplifying the expert's advanced capabilities (\hyperlink{ref-li2022contrastive}{Li et al., 2023}; \hyperlink{ref-obrien2023reasoning}{O'Brien \& Lewis, 2023}). The core bottleneck of CD remains the acquisition of an appropriate weak reference. While context-aware decoding generates this reference by perturbing the input prompt (\hyperlink{ref-shi2023trusting}{Shi et al., 2023}; \hyperlink{ref-tang2026subtraction}{Tang et al., 2026}), DoLa (\hyperlink{ref-chuang2023dola}{Chuang et al., 2024}) and Adaptive Layer-Wise CD (\hyperlink{ref-zhou2025alw}{Zhou et al., 2025}) bypass external modifications by extracting reference logits from intermediate spatial layers within the same dense model. However, spatial layer contrast fails in recurrent architectures. As demonstrated in our analysis, internal layers within an uncompleted recurrent loop drift drastically and do not constitute valid predictive states. LoopCD therefore redefines the contrastive axis from spatial depth to temporal iterations of the identical parameter block. This temporal contrast guarantees that the weak reference represents a structurally valid, fully projected prediction that perfectly shares the vocabulary and feature space of the final state, eliminating the architectural mismatch inherent in multi-model or spatial-layer CD.

\textbf{Guidance in diffusion models.} Guidance is a technique where a diffusion model's output is slightly adjusted at inference time, and has become a critical component of modern image generation models. Classifier-free guidance constructs a guidance vector using the difference between a conditional and an unconditional model pass~(\hyperlink{ref-ho2022cfg}{Ho \& Salimans, 2022}; \hyperlink{ref-chen2022analog}{Chen et al., 2022}). Similarly, autoguidance uses the difference between a weak and a strong model, where the weak and strong model must share similar dynamics (in the sense that they make the same sorts of errors)~(\hyperlink{ref-karras2024autoguidance}{Karras et al., 2024}). Although generally applied to diffusion models, both CFG and autoguidance can be applied to autoregressive models over tokens~(\hyperlink{ref-sanchez2023stay}{Sanchez et al., 2023}). LoopCD works similarly to autoguidance by using the difference between a strong (deep) and weak (shallow) model pass to guide the final output.

%% file: bookmarks.tex
%

\bookmark[page=1,level=0]{Decoding Looped Transformers Better for (Almost) Free}
\bookmark[page=2,level=1]{1 Introduction}
\bookmark[page=2,level=1]{2 Preliminaries: Looped Transformers}
\bookmark[page=3,level=1]{3 LoopCD: Contrastive Decoding from Recurrent States}
\bookmark[page=3,level=2]{3.1 Guidance Space}
\bookmark[page=4,level=2]{3.2 Guidance Strength}
\bookmark[page=5,level=1]{4 Experiments}
\bookmark[page=5,level=2]{4.1 Gains on Mathematical Reasoning and Code Generation}
\bookmark[page=6,level=2]{4.2 Consistent Improvements Across Architectures and Suites}
\bookmark[page=7,level=2]{4.3 Fewer Iterations, Matching or Superior Performance}
\bookmark[page=7,level=1]{5 How and Why LoopCD Works}
\bookmark[page=8,level=2]{5.1 Built-in Weak and Strong Predictions Across Recurrence}
\bookmark[page=10,level=2]{5.2 Re-ranking Close Decisions Under Uncertainty}
\bookmark[page=10,level=2]{5.3 Tuning Guidance Strength Across Scoring and Generation}
\bookmark[page=12,level=1]{6 Conclusion}
\bookmark[page=12,level=1]{References}
\bookmark[page=15,level=1]{A Looped Transformers and LoopCD}
\bookmark[page=15,level=2]{A.1 States and coda layers}
\bookmark[page=15,level=2]{A.2 Model architectures}
\bookmark[page=16,level=2]{A.3 The probability form of LoopCD-Logits}
\bookmark[page=16,level=1]{B Detailed Experimental Setup}
\bookmark[page=16,level=2]{B.1 Model configurations and paired comparisons}
\bookmark[page=16,level=2]{B.2 Benchmarks and table conventions}
\bookmark[page=17,level=2]{B.3 Prompts and sampling}
\bookmark[page=17,level=2]{B.4 Strengths and reference states}
\bookmark[page=18,level=1]{C Extended Results in Full Recurrent Depth}
\bookmark[page=18,level=2]{C.1 GSM8K and MMLU-Pro under LoopCD-Logits}
\bookmark[page=18,level=2]{C.2 Reasoning metrics: pass@1 and pass@10 across sampled trajectories}
\bookmark[page=19,level=1]{D Computation and Cost Analysis}
\bookmark[page=19,level=2]{D.1 FLOP accounting}
\bookmark[page=20,level=2]{D.2 Reduced-depth evaluation: performance recovery and computational tradeoffs}
\bookmark[page=21,level=2]{D.3 Representational dynamics: LoopCD-Logits versus LoopCD-Hidden}
\bookmark[page=24,level=1]{E The Reference Iteration}
\bookmark[page=24,level=2]{E.1 Reference selection across iterations}
\bookmark[page=25,level=2]{E.2 Geometric convergence in representation space}
\bookmark[page=27,level=2]{E.3 Intra-iteration dynamics and output convergence}
\bookmark[page=28,level=1]{F Re-ranking Close Decisions}
\bookmark[page=28,level=2]{F.1 Concentration of guidance gains on uncertain decisions}
\bookmark[page=28,level=2]{F.2 Geometric equivalence of LoopCD-Hidden and orthogonal re-ranking}
\bookmark[page=30,level=2]{F.3 The double-edged effect of temperature sharpening}
\bookmark[page=30,level=1]{G Guidance Strength Dynamics}
\bookmark[page=30,level=2]{G.1 Strength boundaries: generation collapse and negative interpolation}
\bookmark[page=31,level=2]{G.2 The trade-off between decision reach and overshoot}
\bookmark[page=32,level=1]{H Related Work}